\documentclass{article}

\usepackage[nonatbib,final]{neurips_2021_safe_rl}
\usepackage[square,numbers]{natbib}

\usepackage[seed=12210]{randomorder}  
\usepackage[utf8]{inputenc} 
\usepackage[T1]{fontenc}    
\usepackage{hyperref}       
\usepackage{url}            
\usepackage{booktabs}       
\usepackage{amsfonts}     
\usepackage{nicefrac}       
\usepackage{microtype}    
\usepackage[parfill]{parskip}
\usepackage{algorithm,algpseudocode}

\usepackage{float}
\usepackage{subcaption}
\usepackage{times}
\usepackage{graphicx}
\usepackage{caption}
\usepackage{epstopdf}
\usepackage{bbm}

\usepackage{color}
\usepackage{amsmath}
\usepackage{amsthm}
\usepackage{amssymb}
\usepackage[makeroom]{cancel}
\usepackage{mathtools}
\usepackage[justification=centering]{caption}
\usepackage[inline]{enumitem}
\usepackage{wrapfig}
\usepackage{bm}

\graphicspath {{figs/}}
\usepackage{defs}
      
\newcommand{\probf}{ProBF}

\title{ProBF: Learning Probabilistic Safety Certificates \\with Barrier Functions}

\author{%
  Athindran Ramesh Kumar\,\textcircled{r}\thanks{Equal contribution. The order of these two authors is random. See {\url{https://github.com/hips/author-roulette}}}\\
  Princeton University\\
  \texttt{arkumar@princeton.edu} 
   \And
   Sulin Liu\,\textcircled{r}$^*$ \\
   Princeton University \\
   \texttt{sulinl@princeton.edu} \\
   \AND
   Jaime F. Fisac \\
   Princeton University  \\
   \texttt{jfisac@princeton.edu} 
   \And
   Ryan P. Adams \\
   Princeton University  \\
   \texttt{rpa@princeton.edu} \\
   \And
   Peter J. Ramadge \\
   Princeton University \\
   \texttt{ramadge@princeton.edu} \\
}

\begin{document}
\maketitle
\begin{abstract}
Safety-critical applications require controllers/policies that can guarantee safety with high confidence. The control barrier function is a useful tool to guarantee safety if we have access to the ground-truth system dynamics.
In practice, we have inaccurate knowledge of the system dynamics, which can lead to unsafe behaviors due to unmodeled residual dynamics. Learning the residual dynamics with deterministic machine learning models can prevent the unsafe behavior but can fail when the predictions are imperfect. In this situation, a probabilistic learning method that reasons about the uncertainty of its predictions can help provide robust safety margins. In this work, we use a Gaussian process to model the projection of the residual dynamics onto a control barrier function. We propose a novel optimization procedure to generate safe controls that can guarantee safety with high probability. The safety filter is provided with the ability to reason about the uncertainty of the predictions from the GP. We show the efficacy of this method through experiments on Segway and Quadrotor simulations. Our proposed probabilistic approach is able to reduce the number of safety violations significantly as compared to the deterministic approach with a neural network.
\end{abstract}

\section{Introduction}

Reasoning about safety is critical when it comes to controlling a system 
in an unsafe or unpredictable environment. Traditional control design
identifies a parametric first-principles model based on prior knowledge, then uses this model as the basis of controller design. However, such a model has inherent inaccuracies due to parameter estimation errors and errors arising from unmodelled dynamics. 
These issues can make it difficult for the controller to achieve its goals while avoiding unsafe behaviors. 
Learning can be used to reduce model uncertainty (i.e., residuals between the actual and the nominal model)~\citep{berkenkamp2015safe, berkenkamp2017safe, mckinnon2019learning, kumar2021learning}. This can avoid the use of controllers that assume a worst-case bounded model  error~\citep{zhou1996robust, mitchell2005toolbox}, which does not accurately reflect the true model uncertainty. 
\par
In this work, we focus on safety for a system whose exact dynamics are not available. Given an accurate model, one can utilize Control Barrier Functions (CBFs)~\citep{ames2016control,ames2019control} to provide a safety filter for the desired controller by solving a quadratic program (CBF-QP) to project the control to the space of safe controls. For example, \citet{taylor2020learning} directly learns the residuals dynamics of the CBF with a neural network and then synthesizes a safe controller by filtering the goal-reaching controller with the learned dynamics. This learning-based approach (LCBF-QP) results in better modeling of the dynamics for applying the CBF to ensure safety. However, it has a strong requirement that the learned dynamics is sufficiently accurate. This is needed to ensure that the CBF is successful in guaranteeing safety. In practice, this requirement is hard to satisfy with neural nets which are prone to errors as a result of limited training data or out-of-distribution test data. Further, the predictions are deterministic, so even predictions with large potential errors are used with full confidence.
\par
Here, 
we take a probabilistic view of learning model uncertainty. We use Gaussian processes (GPs)~\citep{williams2006gaussian}---a highly flexible Bayesian nonparametric model---to directly model the uncertain dynamics of how the CBF evolves over time. The GP model not only allows us to incorporate prior knowledge about the dynamics via specifying its kernel function, 
but also provides closed-form posterior distributions for evaluation. Our approach is in similar spirit to~\citep{berkenkamp2017safe,fisac2018general, wang2018safe} where probabilistic modeling is used to better capture the uncertainty in the residual dynamics. Specifically, we use probabilistic modeling for safety-critical control with CBFs. (See Section~\ref{sec:related_work} for a more detailed discussion of the differences.) Since we are using CBF to do safety filtering, we use GPs to directly model the residual dynamics of the CBF, which results in a lower output dimension for the GPs. Additionally, with the posterior distribution over the CBF dynamics, we are able to construct a novel CBF safety filtering optimization procedure that provides probabilistic safety guarantee with high confidence under the GP models. However, incorporating variance estimates into the CBF safety constraint gives rise to a new type of non-convex quadratically constrained quadratic program (QCQP), which is challenging to solve. We show how this non-convex optimization problem can be transformed to a convex one. This result is potentially of independent interest for CBF safety filtering with probabilistic constraints under Gaussian dynamics. We call our approach \textit{Probabilistic Safety Certificates with Control Barrier Functions} ({ProBF}).
The effectiveness of this probabilistic safety approach is demonstrated through experiments on a Segway and Quadrotor simulation.

\section{Background}
\subsection{Learning Residual Dynamics for Control Barrier Functions}
First we introduce an important class of functions.
A function $\alpha \colon \reals \rightarrow \reals$ belongs to the set of 
extended class-$\mathcal{K}$ functions $\mathcal{K}_{\infty, e}$~\citep{ames2016control}
if it satisfies: (1) $\alpha$
is strictly increasing and $\alpha(0) = 0$, (2) $\lim_{r \to \infty}\alpha(r) = \infty$ and $\lim_{r \to -\infty}\alpha(r) = -\infty$.

Now consider a non-linear control-affine system with state $\boldx \in \reals^s$ and control $\boldu \in \reals^m$:
\begin{equation}
\dot{\boldx}=\boldf(\boldx)+\boldg(\boldx) \boldu \quad \xrightarrow{\boldu=\mathbf{k}(\boldx)} \quad \dot{\boldx}=\boldf(\boldx)+\boldg(\boldx) \mathbf{k}(\boldx).\label{eq:system_dynamics}
\end{equation}

A control barrier function (CBF) can be used to synthesize a safe controller for the system  \eqref{eq:system_dynamics} that is forward invariant with regards to safety. 
 
\begin{definition}
\textbf{Control Barrier Function (CBF)}\citep{ames2016control}: Let
$h: \reals^{s} \rightarrow \reals$ be a continuously differentiable function with $0$ being a regular value and let $\mathcal{S}:= \{\boldx \in \mathcal{S} \,|\, h(\boldx) \geq 0\} \subset \reals^{s}.$ Then the function $h$ is a Control Barrier Function $(\mathrm{CBF})$ for~\eqref{eq:system_dynamics} on $\mathcal{S}$ if there exists $\alpha \in \mathcal{K}_{\infty, e}$ such that for all $\boldx \in \mathcal{S}:$
\begin{equation}
   \sup _{\boldu \in \reals^{m}} \dot{h}(\boldx, \boldu)=\sup _{\boldu \in \reals^{m}}\left[\frac{\partial h}{\partial \boldx}(\boldx)(\boldf(\boldx)+\boldg(\boldx) \boldu)\right] 
\geq
-\alpha(h(\boldx)). \label{eq: cbf}
\end{equation}
\end{definition}
\begin{theorem} \textbf{Safety Invariance}\citep{ames2016control}:
~Let $h$ be a CBF such that $\mathcal{S}$ is the set of safe states. Then
the system \eqref{eq:system_dynamics} is safe with any controller that satisfies \eqref{eq: cbf}:
i.e. $
K_{\mathrm{cbf}}(\boldx)=\left\{\boldu \in \mathbb{R}^{m} \mid \frac{\partial h}{\partial \boldx}(\boldx)(\boldf(\boldx)+\boldg(\boldx) \boldu) \geq-\alpha(h(\boldx))\right\}
$. \label{thm: cbf}
\end{theorem}
By Theorem~\ref{thm: cbf}, the existence of a CBF enables us to provide a safety filter for the desired controller $\boldu_{d}$ by solving the following quadratic program (CBF-QP):
\begin{align*}
    \min_{\boldu}\;\|\boldu-\boldu_{d}\|^{2}
    \quad \text{s.t.}\;\frac{\partial h}{\partial \boldx}(\boldx)\left(\boldf(\boldx)+\boldg(\boldx) \boldu\right)+\alpha(h(\boldx))\geq 0.
\end{align*}

In practice, the true system model is never known precisely. Instead, 
we have a nominal model 
$
   \widehat{\dot{\boldx}}=\widehat{\boldf}(\boldx)+\widehat{\boldg}(\boldx) \boldu
$
that approximates the true system.
Using the same approach as \citet{taylor2019episodic,taylor2020learning}, we decompose the true system into a nominal model and the residual dynamics: 
\begin{equation}
    \dot{\boldx}=\widehat{\boldf}(\boldx)+\widehat{\boldg}(\boldx) \boldu+\left(\boldf(\boldx)-\widehat{\boldf}(\boldx)\right)+\Big(\boldg(\boldx)-\widehat{\boldg}(\boldx)\Big) \boldu
\end{equation}
The residual dynamics also affects how the time derivative of the CBF is calculated:
\begin{equation}
    \dot{h}(\boldx, \boldu)=\underbrace{\frac{\partial h}{\partial \boldx}(\boldx)\left(\widehat{\boldf}(\boldx)+\widehat{\boldg}(\boldx) \boldu\right)}_{\hat{h}(\boldx, \boldu)}
    +\underbrace{\frac{\partial h}{\partial \boldx}(\boldx) \left(\boldf(\boldx) - \widehat{\boldf}(\boldx) \right)}_{b(\boldx)}
    +\underbrace{\frac{\partial h}{\partial \boldx}(\boldx) \Big(\boldg(\boldx) - \widehat{\boldg}(\boldx)\Big) }_{\mathbf{a}(\boldx)^{\top}} \boldu.
\end{equation}
Here $b(\boldx)$ and $\mathbf{a}(\boldx)$ are the residual dynamics in the CBF time derivative. 

To apply CBF on the true system, we make an assumption on the residual dynamics. Intuitively, we assume that we have the same capability of maintaining safety with the true system as with the system nominal model. This holds when the true system has the same actuation capability as the nominal model~\citep{sastry1999nonlinear}.
\begin{assumption}
If a function $h$ is a valid CBF for the nominal model of the system, then it is $a$ valid CBF for the uncertain system. 
\end{assumption}

In~\citet{taylor2020learning}, trajectory data are collected in discrete time steps following the episodic learning framework. Let
$
\mathcal{D}=\left\{\left(\left(\boldx_{i}, \boldu_{i}\right), \dot{h}_{i}\right)\right\}_{i=1}^{N}
$ denote the collected data from all trajectories,
where $\dot{h}$ is approximated by taking the discrete time approximation of the true CBF values.
The authors then formulate the learning of residual dynamics as an empirical risk minimization (ERM) problem:
\begin{equation*}
    \min _{{\mathbf{a}} \in \mathcal{H}_{a}, {b} \in \mathcal{H}_{b}} \frac{1}{N} \sum_{i=1}^{N} \mathcal{L}\left(\widehat{\dot{h}}(\boldx_i, \boldu_i)+\mathbf{a}(\boldx_i)^{\top} \boldu_i+{b}(\boldx_i), \; \dot{h}_{i}\right).
\end{equation*}
After learning the unmodeled dynamics, a new controller can be synthesized by making use of the updated estimate of the CBF time derivatives using CBF-QP.

\subsection{Gaussian Processes}
In this section, we provide a review of the basic background concepts of Gaussian processes necessary for the discussion of our proposed ProBF approach described in Section~\ref{sec:method}

Given a mean function~$\mu(\boldx)$ and a positive-definite covariance function~$k(\boldx, \boldx^\prime)$ with $\boldx, \boldx^\prime \in \mathcal{X}$, a Gaussian processes defines a distribution over functions~${f: \mathcal{X} \to \reals}$:
\begin{align*}
f(\boldx) &\sim \mathcal{GP}\left(\mu(\cdot), k\left(\cdot, \cdot\right)\right), &
\mu(\boldx) &=\mathbb{E}[f(\boldx)], &
k\left(\boldx, \boldx^{\prime}\right) &=\cov\left(f(\boldx), f(\boldx^{\prime})\right)
\end{align*}
For any finite set of points in~$\mathcal{X}$,~${\boldX := \{ \boldx_1, \ldots, \boldx_N \}}$, the corresponding function values~${\boldf} := ( f(\boldx_1), \cdots, f(\boldx_N) )^{\top}$ follow a multivariate Gaussian distribution:~${{\boldf} \sim \mathcal{N}(\bmu, \boldK_{\boldX\boldX})}$, where~$[\bmu]_i = \mu(\boldx_i)$ and $[\boldK_{\boldX\boldX}]_{ij} = k(\boldx_i,\boldx_j)$.
For a training dataset~${\mathcal{D} = \{(\boldx_i,y_i)\}_{i=1}^N}$, each~$y_i$ is commonly assumed to be observed with an additional i.i.d.\ zero-mean Gaussian noise to~$f(\boldx_i)$, i.e.,~${y_i = f(\boldx_i) + \epsilon_i}$, where~${\epsilon_i \sim \mathcal{N}(0, \sigma_\epsilon^2)}$. Denote $\boldy := [y_1, \cdots, y_N]^\top \in \reals^{N \times 1}$. For new data input $\widetilde{\boldX} := \{ \widetilde{\boldx}_{1}, \ldots, \widetilde{\boldx}_{{N}^\prime}\}$ of size ${N}^\prime$, the Gaussianity of the prior and likelihoods make it possible to compute the predictive distribution in closed form:
\begin{align*}
\widetilde{\boldf} | \widetilde{\boldX}, \mathcal{D} &\sim \mathcal{N}\left(\widetilde{\bmu}, \boldK_{\widetilde{\boldf}}\right),\\
\widetilde{\bmu}&=\boldK_{\widetilde{\boldX}\boldX}\!\!\left(\boldK_{\boldX \boldX}+\sigma_\epsilon^2 \boldI\right)^{-1}\!\!\! \boldy, \\
\boldK_{\widetilde{\boldf}}&=\boldK_{\widetilde{\boldX} \widetilde{\boldX}}\!-\!\boldK_{\widetilde{\boldX} \boldX}\!\!\left(\boldK_{\boldX \boldX}\!+\!\sigma_\epsilon^2 \boldI\right)^{-1}\! \boldK_{\boldX \widetilde{\boldX}},
\end{align*}
where $\boldK_{\boldX \widetilde{\boldX}} \in \reals^{N \times N^\prime}$ with ${[\boldK_{\boldX \widetilde{\boldX}}]}_{ij} = k(\boldx_i,\widetilde{\boldx}_j)$.

The kernel function is chosen to reflect various model assumptions, e.g., smoothness, periodicity, etc. (See Chapter 4 of \citet{williams2006gaussian} for an extensive discussion.) It is also common for the kernel function to have so-called hyperparameters~$\theta$ that govern its specific structure, and the parameterized kernel function is written as~$k_{\theta}(\cdot,\cdot)$. These hyperparameters are most commonly determined via maximizing the log marginal likelihood (evidence) with respect to the hyperparameter~$\theta$ using empirical Bayes~\citep{berger2013statistical,mackay1992evidence}. The log MLL for observed data $\{\boldX, \boldy\}$ is given by:
\begin{equation}\label{eq:mll}    
\log p(\boldy|\boldX,\theta) =-\frac{1}{2} \boldy^{\top} \left(\boldK_{\boldX\boldX}(\theta) + \sigma_\epsilon^2 \boldI \right)^{-1} \boldy-\frac{1}{2} \log \left| \boldK_{\boldX\boldX}(\theta) + \sigma_\epsilon^2 \boldI \right|-\frac{N}{2} \log 2 \pi\,,
\end{equation}
where we write~$\boldK_{\boldX\boldX}(\theta)$ to indicate the dependence of $\boldK_{\boldX\boldX}$ on the hyperparameters.

\section{Probabilistic Safety Certificate with Control Barrier Function}\label{sec:method}
\subsection{Probabilistic Modeling of Residual Dynamics}
Instead of modeling the residual dynamics as deterministic functions, we place Gaussian process~\citep{rasmussen2003gaussian} priors on ${\mathbf{a}}(\boldx)$ and ${b}(\boldx)$. 
 Using the distribution over the residual dynamics, we use the confidence estimate to provide robust safety on regions of the state space where the estimates are not accurate.
In particular, ${b}(\boldx) \sim \mathcal{GP}(0,k_b(\cdot,\cdot))$ is modeled as a single-output GP and ${\mathbf{a}}(\boldx)$ is modeled as a multi-output GP with ${\mathbf{a}_j}(\boldx) \sim \mathcal{GP}(0,k_{\bolda,j}(\cdot,\cdot))$, where $k_b$ and $k_{\bolda,j}$ are the covariance functions. ${\bolda_j(\boldx)}$'s are independent of each other if we assume all entries of $\boldA(\boldx)$ are independent. 
As a result, $d(\boldx,\boldu):=h(\boldx,\boldu)-\hat{\dot{h}}(\boldx, \boldu) = \bolda(\boldx)^\top\boldu + b(\boldx)$ will be a GP with zero mean and covariance function $k_d$ given in closed form:
\begin{align*}
d(\boldx,\boldu) &\sim\mathcal{GP}(\mu(\cdot),k_{d}(\cdot,\cdot)),\\
\mu_d(\boldx,\boldu) &=0,\\
k_d((\boldx,\boldu), (\boldx^{\prime},\boldu^\prime)) &=\sum_{j=1}^{m}k_{\bolda,j}(\boldx,\boldx^\prime)\boldu_{j}\boldu_{j}^\prime+k_{b}(\boldx,\boldx^\prime).
\end{align*}
Assume we are given the training dataset $\mathcal{D}_{\text{train}}$ in the form of collected trajectories $\mathcal{X}_{\text{train}}:=((\boldx_{1},\boldu_{1}),\dots,(\boldx_{N},\boldu_{N}))$ and $\boldY_{\text{train}}:=[d_{1},\dots,d_{N}]^{\top} \in \reals^{N}$. We denote $\boldX_{\text{train}}:=\left[\boldx_{1},\dots,\boldx_{N}\right] \in \reals^{s\times N}$ and $\boldU_{\text{train}}:=\left[\boldu_{1},\dots,\boldu_{N}\right] \in \reals^{m\times N}$. Let $K_{\text{train}}\in\mathbb{R}^{N\times N}$ be the covariance matrix on $\mathcal{X}_{\text{train}}$ evaluated using $k_{d}(\cdot,\cdot)$.

The posterior distribution of a new test point $d(\boldx^*,
\boldu^*)$ is given by:
\begin{align}
    d(\boldx^{*},\boldu^{*})|\mathcal{D}_\text{train}&=\mathcal{N}\left(\mu_d(\boldx^{*},\boldu^{*}),\;\sigma_d^2(\boldx^{*},\boldu^{*})
    \right), \label{eq: posterior}\\
    \mu_d(\boldx^{*},\boldu^{*}) &= \bar{\bolda}(\boldx^*)^{\top}\boldu^*+\bar{b}(\boldx^*),\\
    \sigma_d^2(\boldx^{*},\boldu^{*}) &= {\boldu^*}^\top\bSigma_\bolda(\boldx^{*})\boldu^{*}+2\bSigma_{\bolda,b}(\boldx^{*})^\top\boldu^{*}+\sigma_b(\boldx^*)^2,
\end{align}
where $\bar{\bolda}(\boldx),\bar{b}(\boldx)$ represents the posterior mean of ${\bolda(\boldx)},{b(\boldx)}$ and $\bSigma_\bolda(\boldx),\sigma_b^2(\boldx),\bSigma_{\bolda,b}(\boldx)$ represents the posterior covariance of ${\bolda}(\boldx),{b}(\boldx)$. It is worth noting that although 
$\bolda(\boldx)$ is independent of $b(\boldx)$, the posterior variance does not simply decompose into the variances induced by $\bolda(\boldx)^{\top} \boldu$ and $b(\boldx)$.
For the closed forms of posterior mean and covariance and their derivations, we refer readers to Appendix \ref{sec:posterior}.
\subsection{Projection to Probabilistic Safety}
The posterior distribution provided by the GP characterizes the uncertainty of the CBF dynamics. The GP posterior is expected to be more confident (i.e. with smaller variance) about its predictions in regions with more data available, and less confident (i.e. with larger variance) in regions with less data.
Given the posterior distribution from the GP, we find actions that guarantee safety with high probability by solving the following probabilistic-safety projection problem:
\begin{align}
    &\min_{\boldu}\;  \|\boldu-\boldu_{d}\|^{2} \label{eq: prob_qp_obj}
    \\
    &\text{s.t.}\; \Pr \Big(\frac{\partial h}{\partial \boldx}(\boldx)\left(\mathbf{\hat{f}}(\boldx)+\mathbf{\hat{g}}(\boldx) \mathbf{\boldu}\right)+\alpha(h(\boldx))+d(\boldx,\boldu)>0 \Big) \geq 1-\epsilon. \label{eq: prob_qp}
\end{align}
Since $d(\boldx,\boldu)$ is Gaussian from \eqref{eq: posterior}, the probabilistic safety constraint in~\eqref{eq: prob_qp} is equivalent to: 
\begin{align}
   \quad & \left(\bar{\bolda}(\boldx)+\boldc_{\bolda}(\boldx)\right)^{\top}\boldu+\bar{b}(\boldx)+c_b(\boldx)   -\delta\sqrt{\mathbf{\boldu}^{\top}\bSigma_{\bolda}(\boldx)\mathbf{\boldu}+2\bSigma_{\bolda,b}(\boldx)^{\top}\mathbf{\boldu}+\sigma_b(x)^2} \geq 0, \label{eq: prob_cbf_qp}
\end{align}
for a corresponding $\delta$ calculated from $\epsilon$, where $\boldc_{\bolda} = \left( \frac{\partial h}{\partial \boldx}(\boldx)\mathbf{\hat{g}}(\boldx)\right)^\top $ and $\boldc_{\boldb} = \frac{\partial h}{\partial \boldx}(\boldx)\mathbf{\hat{f}}(\boldx) + \alpha(h(\boldx))$ are the constant terms obtained from the known nominal dynamics. Note that if we set $\delta = 0$, this reduces back to the LCBF-QP~\citep{taylor2020learning} formulation as a special case of our method by only using the mean predictions from the GP.

The constraint in \eqref{eq: prob_cbf_qp} is non-convex, but it is possible to transform it to a convex one.

\begin{theorem}\label{thm: convex}
\textbf{Convex Projection to Probabilistic Safety:}
The non-convex probabilistic safety projection optimization problem in~\eqref{eq: prob_qp_obj} can be transformed to a convex optimization problem. 
\end{theorem}
Here we give a brief outline of the proof. For details, please refer to Appendix \ref{cvxappendix}.

\textit{Proof Sketch.}
We can first transform this problem into a quadratically constrained quadratic program (QCQP) by introducing an auxiliary variable $s$:
\begin{align}
    \min_{\boldu,s}\quad&\|\boldu-\boldu_{d}\|^{2} \label{eq: QCQP}\\
   \text{s.t.}\quad&\left(\bar{\bolda}(\boldx)+\boldc_a(\boldx)\right)^{\top}\boldu+\bar{b}(\boldx)+c_b(\boldx) -\delta s \geq 0 \label{eq: constraint_1}\\ &\boldu^{\top}\bSigma_{\bolda}(\boldx)\boldu+2\bSigma_{\bolda,b}(\boldx)^{\top}\boldu+\sigma_b(x)^2 \leq s^{2} \label{eq: constraint_2}\\
   &s\geq 0 \label{eq: constraint_3}
\end{align}
Then, by defining $\bar{\boldu}=\begin{bmatrix}\boldu & t &s\end{bmatrix}^{\top} \in \mathbb{R}^{m+2}$, we can rewrite the non-convex quadratic constraint in \eqref{eq: constraint_2} as $\Vert \overline{\bSigma} \mathbf{\bar{\boldu}}\Vert \leq \boldc^{\top} \mathbf{\bar{\boldu}}$
with $t=1$, $\overline{\bSigma}=\sqrt{\begin{bmatrix} \bSigma_{\bolda}(\boldx) & \bSigma_{\bolda,b}(\boldx)&\mathbf{0} \\
\bSigma_{\bolda,b}(\boldx)&\sigma_b(\boldx)^2&0\\
0&0&0\end{bmatrix}}$ and $\boldc=\begin{bmatrix}\mathbf{0}\\0\\1\end{bmatrix}$. Now, we can rewrite the non-convex QCQP as a convex program:
\begin{align*}
\min_{\mathbf{\bar{\boldu}}}\quad&\text{ }\mathbf{\bar{\boldu}^{\top}}\boldQ\mathbf{\bar{\boldu}}
\\
\text{s.t.}\quad&\Vert \overline{\bSigma} \mathbf{\bar{\boldu}}\Vert \leq \boldc^{\top} \mathbf{\bar{\boldu}}, \quad
\boldC^{\top}\mathbf{\bar{\boldu}}+\boldd\leq 0, \quad \boldD^{\top}\mathbf{\bar{\boldu}}+\boldf= 0
\end{align*}
with the matrices $\boldQ \succcurlyeq 0$, $\boldC$, $\boldD$, $\boldd$, $\boldf$ defined properly to enforce constraints \eqref{eq: QCQP}, \eqref{eq: constraint_1} and \eqref{eq: constraint_3}. The equality constraint $t=1$ is enforced by:
\begin{align}
\boldD&=\begin{bmatrix}
\mathbf{0}\\1\\0
\end{bmatrix},  \boldf=\begin{bmatrix}\mathbf{0}\\-1\\0\end{bmatrix}.
\end{align}
The remaining inequality constraints can be enforced with:
\begin{align}
    \boldC=\begin{bmatrix}
    -\left(\bar{\bolda}(\boldx)+\boldc_a(\boldx)\right)^{\top}& -(\bar{b}(\boldx)+\boldc_b(\boldx))& \delta\\
    0 & 0 & -1\\
    \end{bmatrix}, \boldd=\mathbf{0}. 
\end{align}
Finally, the matrix in the objective function $\boldQ$ can be defined as:
\begin{align}
    \boldQ = \begin{bmatrix}
    \boldI&-\boldu_{d}&\bold0\\-\boldu_{d}^{\top}&\boldu_{d}^{\top}\boldu_{d}&\bold0\\\bold0&\bold0&\bold0
    \end{bmatrix}.
\end{align} We can then solve this convex program using standard convex optimization solver such as CVXPY~\citep{diamond2016cvxpy}. 

\section{Numerical Results}
\subsection{Segway}
We test our approach on a simulated Segway platform~\citep{taylor2020learning}. The control barrier function is designed to keep the Segway from flipping over while not exceeding an angular velocity threshold:
$
    h(\theta,\dot{\theta})=\theta_{m}^{2}-\dot{\theta}^{2}-\left(\theta-\theta_{e}\right)^{2}
$,
with $\theta_{m}=0.2617$, $\theta_{e}=0.1383$. The stabilizing controller is a proportional-derivative (PD) controller designed to reach the goal state. The model uncertainty comes from
the nominal system parameters being inaccurate. Due to the residual dynamics, the PD controller with the CBF-QP escapes out of the safe region. We learn the residuals with a GP using an episodic learning framework similar to~\citet{taylor2020learning}. We use Matérn $\nicefrac{5}{2}$ kernels for $k_{\bolda,j}$ and $k_{b}$ with individual lengthscales for each dimension. The Matérn $\nicefrac{5}{2}$ kernel is chosen because it better reflects the smoothness assumption of the residuals to be learned.
The lengthscales are set by maximizing marginal likelihood on the training data.
The \probf~safety projection is solved with an initially chosen $\delta=1.0$ until any time step when the convex program becomes infeasible. After this point, $\delta$ is reduced to ensure feasibility and used for the rest of the episode.
\par
\begin{figure}[ht!]
\centering
\begin{subfigure}{.45\textwidth}
	   \centering
	   \includegraphics[width=\textwidth]{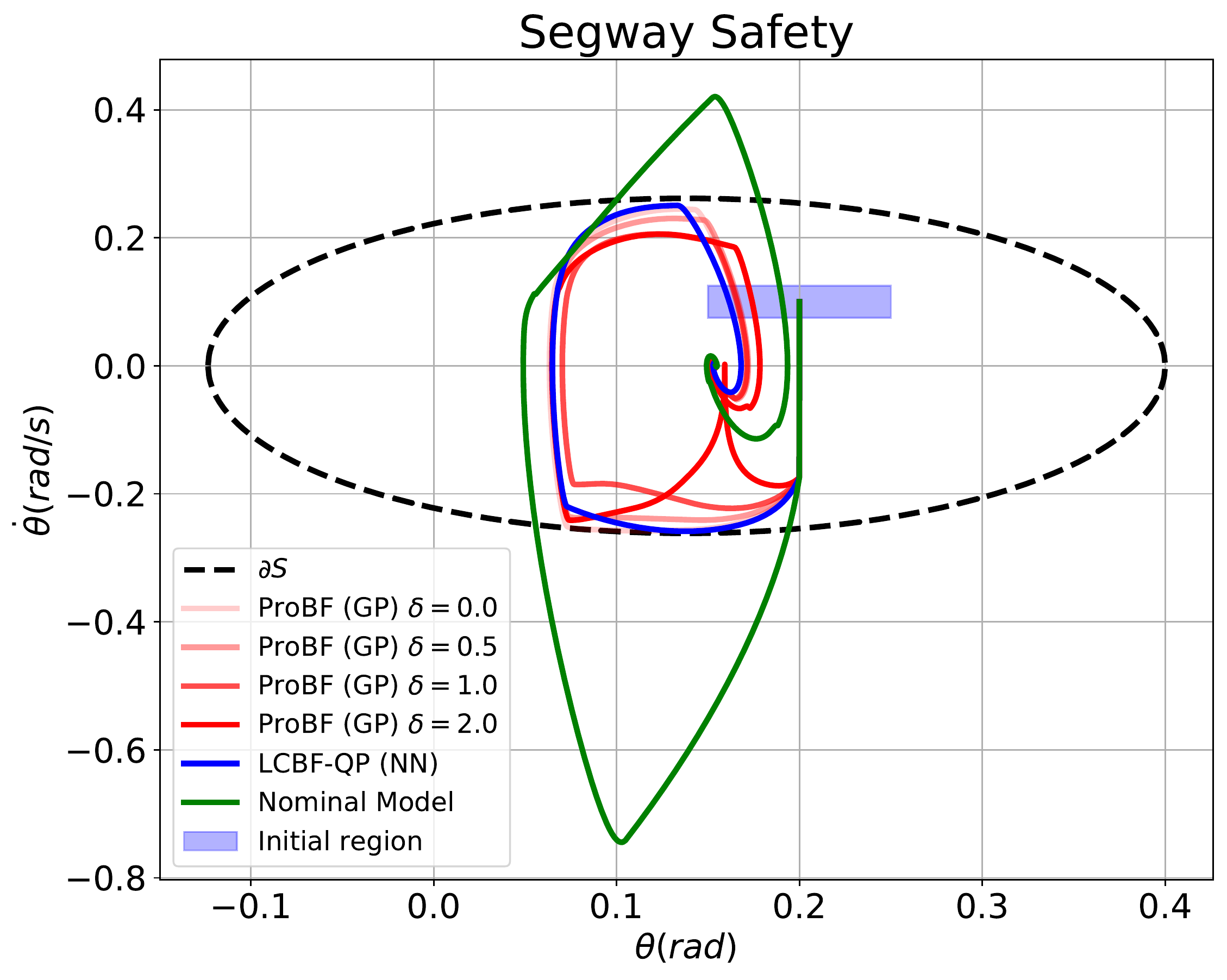}
	   \caption{Starting from the center of the region used for training. }\label{fig:center}	   
\end{subfigure}
\begin{subfigure}{.45\textwidth}
	   \centering
	   \includegraphics[width=\textwidth]{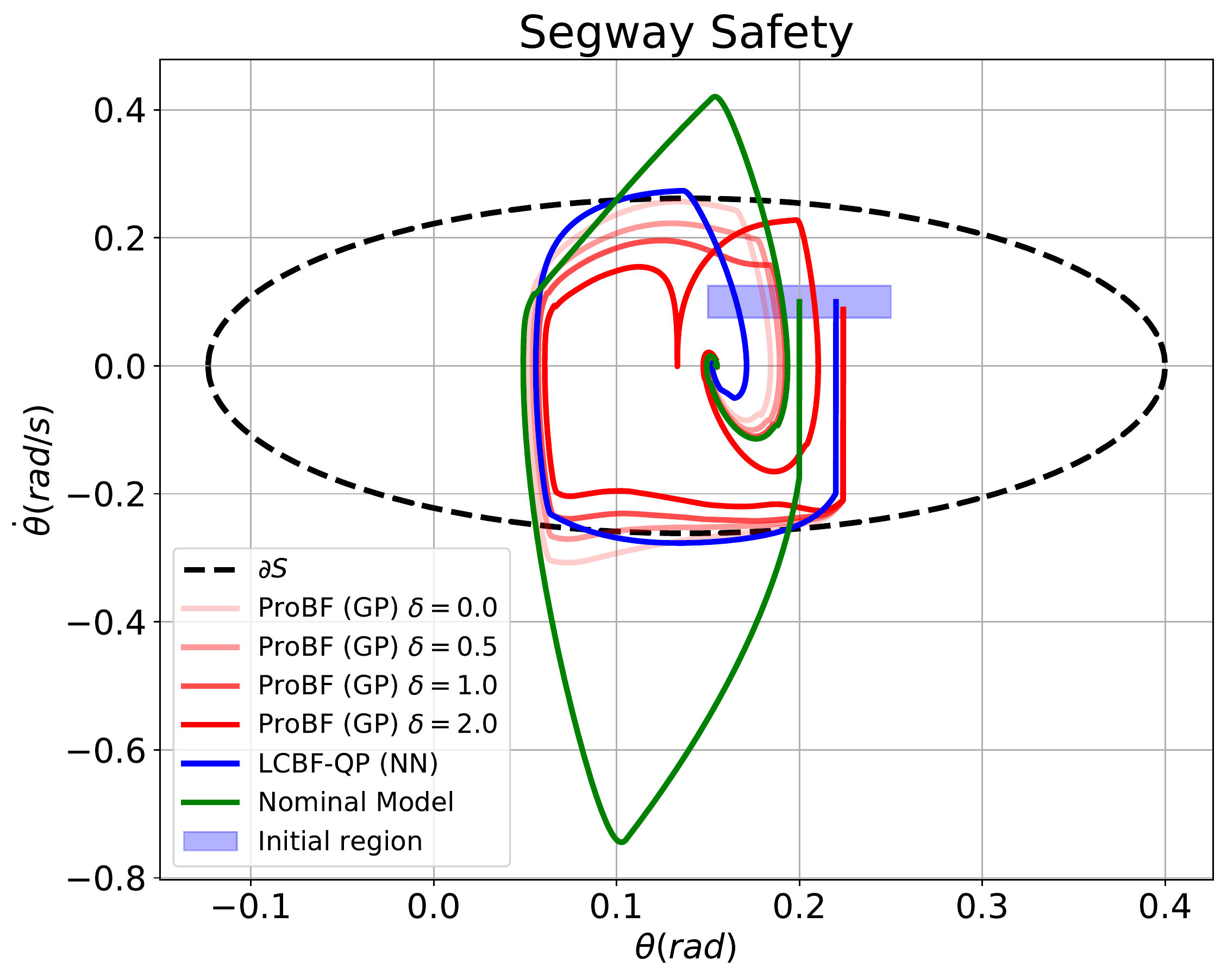}
	   \caption{Starting from closer to the boundary of the region used for training.}\label{fig:bound}
\end{subfigure}	 
\caption{Segway: Comparing the learned controller with \probf~and LCBF-QP. For \probf, we use different $\delta$'s that correspond to different level of probabilistic safety. $\delta \mathcal{S}$ is the boundary of safety set.}\label{fig:compare}
\end{figure}

In Figure~\ref{fig:compare}, we depict the advantage of using our approach (\probf) compared to the learned CBF with a neural network (LCBF-QP)~\citep{taylor2020learning}. We train both the GP and NN using the same set of initial points randomly drawn from a fixed region of the state space for $5$ episodes. In Figure \ref{fig:center}, we test the trained models from the center of the region. Both the GP and NN can give rise to a safe controller but the safety margin can be controlled using $\delta$. In Figure \ref{fig:bound}, we test from a point closer to the boundary. In this case, the NN and GP without variance ($\delta=0$) marginally violate safety. However, solving the \probf~safety filtering optimization with the variance can ensure safety. 
\par
We repeat the same episodic learning procedure $10$ times for both methods with the different sets of initial points and compare the number of safety violations by testing on $10$ new random initial points. In each run, both methods are trained for $5$ episodes. Across the $10$ runs, we observe that ProBF violates safety on $24 \pm 22.7\%$ of the testing trials, while LCBF-QP violates $47 \pm 30.6\%$ of the time. Further, most of the safety violations do not result in total catastrophe and are minor deviations out of the safe set.
With the help of uncertainty estimates, ProBF is able to reduce the average violations by $50\%$ and has a lower variance. (See Appendix~\ref{sec:appendix_experiment} for plots of trajectories.)
Further, \probf~also gives early warning of safety violations when the uncertainty is too big such that there is no feasible solution to \probf.
\subsection{Planar Quadrotor with Gravity}
We also test our probabilistic approach on a 2D quadrotor example with gravity in one of the directions. An extension technique~\citep{quadrotorex} is used to design a feedback-linearization stabilizing controller with 2 controls for the quadrotor. The two control inputs to the drone are the thrust force and the torque about the center of mass.
The stabilizing controller will steer the quadrotor to the origin from the region surrounding $[2.0,2.0]$. We define an unsafe region as a circle around the point $(1.85,0.6)$ in the 2D space. Hence, we define the control barrier function as $h(x,y)=(x-1.85)^{2}+(y-0.6)^{2}-0.28$. The true parameters of the quadrotor are: $m=1.8, J=1.1$, while the nominal estimates are $m=1.5, J=1.3$. An higher order CBF-QP (fourth order) is solved using a technique similar to~\cite{wang2020learning}. This requires us to compute fourth derivatives of the barrier function leading to spiky residuals. With just the nominal model, the stabilizing controller with the CBF-QP cannot prevent entry into the unsafe region. This shows a clear need to learn the residual terms to improve safety. We count a safety violation whenever $h<-0.05$ anywhere in the trajectory.
\par
\begin{figure}[ht!]
\centering
\begin{subfigure}{.45\textwidth}
	   \centering
	   \includegraphics[width=\textwidth]{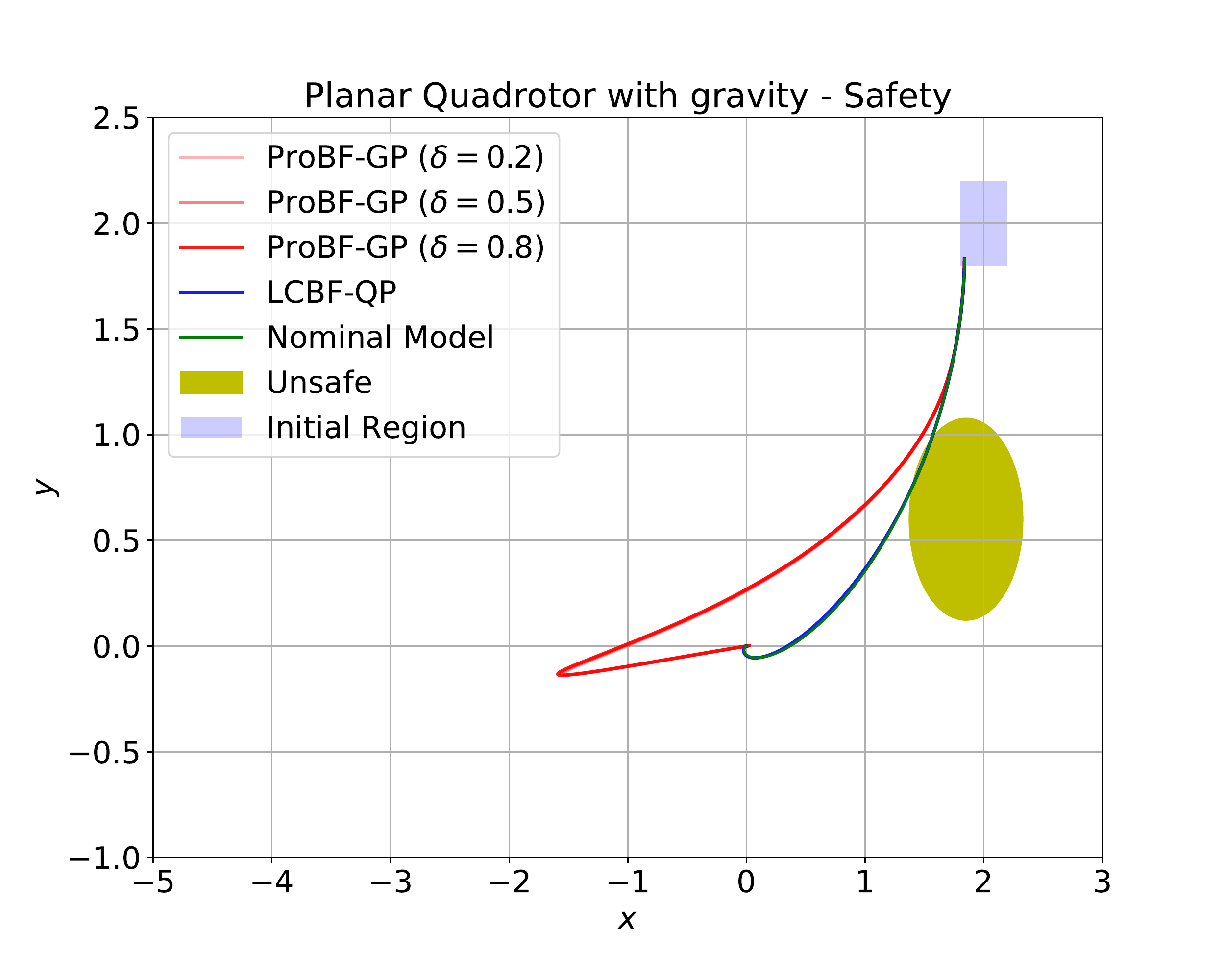}
	   \caption{Starting from the left-bottom of the region used for training. }\label{fig:qlb}	   
\end{subfigure}
\begin{subfigure}{.45\textwidth}
	   \centering
	   \includegraphics[width=\textwidth]{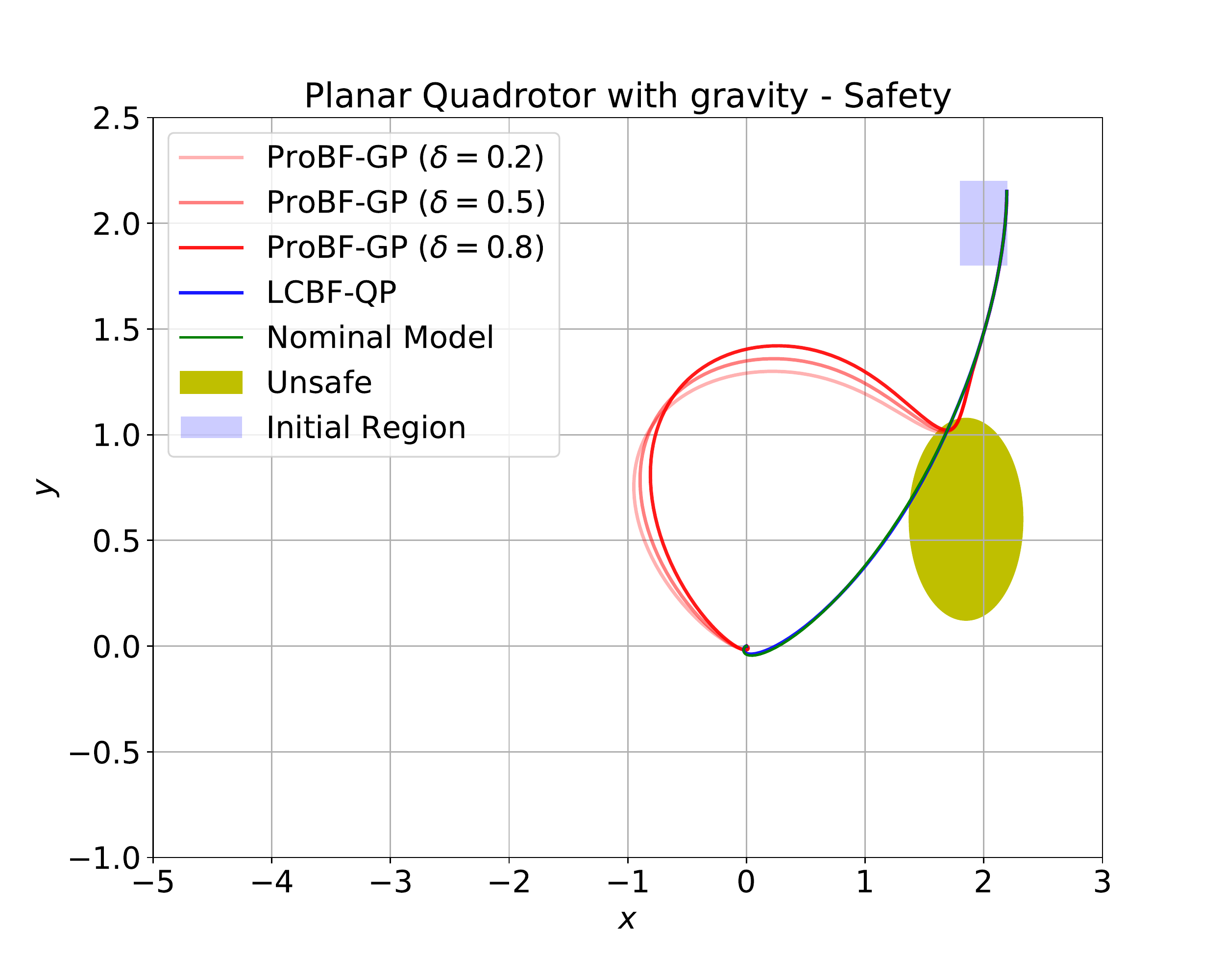}
	   \caption{Starting from the top-right of the region used for training.}\label{fig:qtr}
\end{subfigure}	 
\caption{Quadrotor: Comparing the learned controller with \probf~and LCBF-QP. For \probf, we use different $\delta$'s that correspond to different level of probabilistic safety.}\label{fig:quadcompare}
\end{figure}

In Figure~\ref{fig:compare}, we compare ProBF with LCBF-QP ~\citep{taylor2020learning} for quadrotor safety. We train both the GP and NN using the same set of initial points randomly drawn from a fixed region of the state space for $10$ episodes. In Figure \ref{fig:qlb}, we test the trained models from the left-bottom of the initial region. Both the GP and NN can give rise to a safe controller but the safety margin can be controlled using $\delta$. In Figure \ref{fig:qtr}, we test from a point from the top-right of the initial region. In this case, the NN performs poorly and the quadrotor passes right into the unsafe region. However, solving the \probf~safety filtering optimization with the variance can ensure safety. 
We repeat the same episodic learning procedure $10$ times for both methods with the different sets of initial points and compare the number of safety violations by testing on $10$ new random initial points. In each run, both methods are trained for $10$ episodes. For both LCBF-NN and ProBF, the training diverged for one seed without any further tuning of the hyperparameters.  Across the $9$ remaining runs, we observe that ProBF violates safety on $11.1 \pm 11.9\%$ of the testing trials, while LCBF-QP violates $38.8 \pm 21.3\%$ of the time. Again, ProBF is able to significantly reduce the average violations (by more than $70\%$) and has a lower variance. (See Appendix~\ref{sec:appendix_experiment} for plots of trajectories.)

\section{Related Work}\label{sec:related_work}
\subsection{Control Barrier Functions}
Control barrier function \citep{ames2016control,ames2019control} is a tool to provide safety filters with guarantees.  
Robust CBFs \citep{nguyen2016optimal,gurriet2018towards} can be used to provide robust safety but the uncertainty of the given system is assumed to be worst-case from a bounded set leading to unnecessary conservatism. The safety filtering optimization using robust CBFs can only be solved approximately by making polytopic over-approximations of the safety constraint set.
Similar idea is exploited in~\citet{luo2020multi} for multi-robot collision avoidance under uncertainty by assuming bounded uniform random noise in the system dynamics.
The controls are solved via approximate projection onto the probabilistic collision free sufficiency conditions.
Instead of assuming worst case uncertainty, \citet{taylor2020learning} learns a safe controller by episodically learning the residuals in the barrier certificate dynamics with a neural network. 

In \citet{choi2020reinforcement}, a reinforcement learning approach is taken to learn the residuals terms in CLF and CBF. However, the function approximator is still parametric. \citet{cheng2019end} use CBF's to incorporate safety constraints into model-free reinforcement learning.
In contrast, our work is directed towards the model-based setting where an approximate model is given and the unmodeled dynamics are learned with a Bayesian model.

\subsection{Gaussian Processes for Learning Dynamics}
Several previous works have proposed to use GP for modelling dynamics~\citep{berkenkamp2017safe,fisac2018general,wang2018safe}. In \citet{berkenkamp2017safe}, the authors propose to use GP for modelling the transition dynamics in model-based reinforcement learning. In \citet{fisac2018general}, GP is used to model the disturbance term in the dynamical system for safe learning-based control.

The most relevant works to ours are~\citet{wang2018safe} and ~\citet{khojasteh2020probabilistic}. ~\citet{wang2018safe} uses GP to model the residual control-independent dynamics for safe learning of quadrotors with CBFs. Using a similar idea,~\citet{cheng2020safe} models the residual control-independent dynamics using GP and constructs an uncertainty polytope given a desired confidence level. Robust CBF~\citep{nguyen2016optimal,gurriet2018towards} is then applied on the uncertainty polytope for multi-agent collision avoidance.
In~\citet{khojasteh2020probabilistic}, system dynamics is modeled with GPs and a probabilistic safety constrained non-convex QCQP is solved. In comparison, we model residual dynamics directly on the time derivative of CBF unlike~\citet{khojasteh2020probabilistic}. In contrast to~\citet{wang2018safe}, our approach places probabilistic models on both the control-independent dynamics and the control-dependent dynamics. We additionally derive a novel probabilistic safety-filtering optimization formulation by reformulating the non-convex optimization problem as a convex one. This result can be of independent interest for solving any probabilistic safety projection problems where dynamics are Gaussian-distributed. 

While preparing the camera ready manuscript, we discovered a line of recent work~\citep{castaneda2021gaussian, castaneda2021pointwise} that studies modeling residual dynamics of control Lyapunov functions (CLFs) and control Barrier functions (CBFs) with GPs. This line of work solves the chance-constrained optimization by transforming it to a convex one with a similar technique~\citep{lobo1998applications} to ours. While our work shares the same spirit of using probabilistic CLF/CBFs and applies similar techniques to transform the non-convex QCQP into a convex one, we focus on achieving safety with a given stabilizing controller and extend it to higher-order CBF constraints for the planar quadrotor. 
Further, we validate the effectiveness of probabilistic modeling over deterministic modeling by comparing statistics of safety violations over multiple random seeds for both training and testing on two different examples.
\section{Conclusions and Future Work}\label{sec:conclusions}
Probabilistic modeling is crucial for incorporating learning methods in safety-critical control. Learning residual models that can provide epistemic uncertainty can help the controller take conservative actions for safety in regions where its estimates are bad. In this work, we learn the residuals of barrier dynamics using a GP. We incorporate the uncertainty estimates from the GP into a probabilistic constraint. This non-convex program is transformed into a convex program that can optimally project the control actions to the safe set of actions. We provide numerical experiments on a Segway and Quadrotor to validate the efficacy of our method. As future work, extending this approach for safe exploration and providing robust mechanisms to prevent infeasibility of the convex program would be interesting.

\section{Acknowledgement}
We would like to thank Andrew J. Taylor for his help on kindly providing and explaining the code of running experiments with~\citep{quadrotorex}. PJR acknowledges support from NSF MRI Award: 1919452. 
\bibliography{mybib}
\bibliographystyle{abbrvnat}
\appendix
\newpage
\section{Posterior Mean and Variance Decomposition}\label{sec:posterior}
The posterior mean at a given state-control pair $(\boldx^*,\boldu^*)$ can be further reduced to a linear function of $\boldu^*$:
\begin{align}
    \mathbb{E}\left[d(\boldx^{*},\boldu^{*})|\mathcal{D}_\text{train}\right] &=\boldK_{*,\mcX_\text{train}}\left(\boldK_{\mcX_\text{train}, \mcX_\text{train}}+\sigma_\epsilon^2 \boldI\right)^{-1} \boldy_\text{train}
    \\&=\sum_{i=1}^s \balpha_i k_d((\boldx^*,\boldu^*),(\boldx_i,\boldu_i))
    \\&= \sum_{i=1}^s \balpha_i \sum_{j=1}^m k_{a,j}(\boldx^*,\boldx_i)\boldu_{i,j}\boldu_{*,j} + \sum_{i=1}^s \balpha_i k_b(\boldx^*,\boldx_i)
    \\&=\bar{\bolda}(\boldx^*)^{\top}\boldu^*+\bar{b}(\boldx^*),
\end{align}
where $\balpha = \left(\boldK_{\mcX_\text{train}, \mcX_\text{train}}+\sigma_\epsilon^2 \boldI\right)^{-1} \boldy_\text{train}$.

Similarly, we derive posterior variance. Since control $\boldu$ is deterministic, we will be directly calculate the posterior covariance of $\bolda(\boldx) \in \reals^m$ and $b(\boldx) \in \reals$:
\begin{align}
  &\cov\left(\begin{bmatrix} 
  \bolda(\boldx^{*})\\b(\boldx^{*})\end{bmatrix}\Big| \boldY_\text{train}\right)=
  \\&\cov\left(\begin{bmatrix} \bolda(\boldx^{*})\\b(\boldx^{*})\end{bmatrix}\right) - 
  \cov\left(\begin{bmatrix} 
  \bolda(\boldx^{*})\\b(\boldx^{*})\end{bmatrix},\boldY_\text{train}\right)
  \left(\boldK_{\mcX_\text{train}, \mcX_\text{train}}+\sigma_\epsilon^2 \boldI\right)^{-1}
  \cov\left(\boldY_\text{train},\begin{bmatrix} 
  \bolda(\boldx^{*})\\b(\boldx^{*})\end{bmatrix}\right)
  \label{eq:covariance}
\end{align}

\begin{align}
    \cov\left(\begin{bmatrix} \bolda(\boldx^{*})\\b(\boldx^{*})\end{bmatrix}\right) = \begin{bmatrix}
  \diag(\var(\bolda(\boldx^*)_1,\cdots,\var(\bolda(\boldx^*)_m)))&0\\0&\var(b(\boldx^{*}))
  \end{bmatrix}
\end{align}

\begin{align}
    \cov\left(\begin{bmatrix} 
    \bolda(\boldx^{*})\\b(\boldx^{*})\end{bmatrix},\boldY_\text{train}\right) &= \cov\left(\begin{bmatrix} 
    \bolda(\boldx^{*})\\b(\boldx^{*})\end{bmatrix},\left[\bolda(\boldx_1)^\top\boldu_1+b(\boldx_1),\cdots,\bolda(\boldx_s)^\top\boldu_s+b(\boldx_s)\right]^\top\right)\\
    &=\begin{bmatrix} \widetilde{\boldK}_{a}(\boldx^*,\boldx_\text{train}) \\
  \widetilde{\boldK}_{b}(\boldx^{*},\boldx_\text{train})^\top 
  \end{bmatrix},
\end{align}
where $\widetilde{\boldK}_{\bolda}(\boldx^*,\boldX_\text{train})_{j,i} = k_{a,j}(\boldx^*,\boldx_i)\boldu_{i,j}$ and $\widetilde{\boldk}_b(\boldx^*,\boldX_\text{train})_i = k_b(\boldx^*,\boldx_i)$.

Now we can give the posterior covariance in closed form:
\begin{align*}
\sigma_{b}(\boldx^{*})^2 = \var[{b}(\boldx^{*})|\boldY_\text{train}] 
&=k_{b}(\boldx^*,\boldx^*)-\widetilde{\boldk}_b(\boldx^*,\boldX_\text{train})^{\top}  \left(\boldK_{\mcX_\text{train}, \mcX_\text{train}}+\sigma_\epsilon^2 \boldI\right)^{-1}
\widetilde{\boldk}_b(\boldx^*,\boldX_\text{train})
\\
\bSigma_\bolda(\boldx^*)=\var[\bolda(\boldx^{*})_j|\boldY_\text{train}]
&=k_{a,j}(\boldx^*,\boldx^*)-
\widetilde{\boldK}_a(\boldx^{*},\boldX
_\text{train})_{j,*}
\left(\boldK_{\mcX_\text{train}, \mcX_\text{train}}+\sigma_\epsilon^2 \boldI\right)^{-1}
\widetilde{\boldK}_a(\boldx^{*},\boldX
_\text{train})_{j,*}^\top
\\
\cov\left[
\bolda(\boldx^{*})_j,
{b}(\boldx^{*})
|\boldY_\text{train}\right]
&=- \widetilde{\boldK}_a(\boldx^{*},\boldX
_\text{train})_{j,*}
\left(\boldK_{\mcX_\text{train}, \mcX_\text{train}}+\sigma_\epsilon^2 \boldI\right)^{-1}
\widetilde{\boldk}_b(\boldx^*,\boldX_\text{train})
\\
\cov\left[
\bolda(\boldx^{*})_j,
\bolda(\boldx^{*})_k
|\boldY_\text{train}\right]
&=- \widetilde{\boldK}_a(\boldx^{*},\boldX
_\text{train})_{j,*}
\left(\boldK_{\mcX_\text{train}, \mcX_\text{train}}+\sigma_\epsilon^2 \boldI\right)^{-1}
\widetilde{\boldK}_a(\boldx^{*},\boldX
_\text{train})_{k,*}^\top
\end{align*}

Let's define:
\begin{align}
\cov\left(\begin{bmatrix} 
\bolda(\boldx^{*})\\b(\boldx^{*})\end{bmatrix}\Big| \boldY_\text{train}\right) 
=\begin{bmatrix} 
\bSigma_a(\boldx^*) &\bSigma_{a,b}(\boldx^*)\\\bSigma_{a,b}(\boldx^*)^\top &\sigma_b(\boldx^*)^2\end{bmatrix}.
\end{align}
The posterior distribution of $d(\boldx^*,
\boldu^*)$ is given by:
\begin{align*}
    d(\boldx^{*},\boldu^{*})|\boldY_\text{train}&=\mathcal{N}\left(\bar{\bolda}(\boldx^*)^{\top}\boldu^*+\bar{b}(\boldx^*),\quad
    {\boldu^*}^\top\bSigma_a(\boldx^{*})\boldu^{*}+2\bSigma_{a,b}(\boldx^{*})^\top\boldu^{*}+\sigma_b(\boldx^*)^2\right).
\end{align*}
\section{Proof of Theorem \ref{thm: convex}: Probabilistic Safety Projection Optimization}\label{cvxappendix}
The probabilistic safety projection problem of interest is:
\begin{align}
\centering
    \min_{\mathbf{\boldu}}\quad & \|\mathbf{\boldu}-\mathbf{\boldu}_{d}\|^{2} \label{eq: aprob_cbf_qp}\\
   \text{s.t.}\quad &\left(\bar{\bolda}(\boldx)+\boldc_a(\boldx)\right)^{\top}\boldu+\bar{b}(\boldx)+\boldc_b(\boldx)   -\delta\sqrt{\mathbf{\boldu}^{\top}\bSigma_{\bolda}(\boldx)\mathbf{\boldu}+2\bSigma_{\bolda,b}(\boldx)^{\top}\mathbf{\boldu}+\sigma_b(\boldx)^2} \geq 0. \nonumber
\end{align}
We can first transform this problem into a quadratically constrained quadratic program (QCQP) by introducing an auxiliary variable $s$:
\begin{align}
    \min_{\boldu,s}\quad&\|\boldu-\boldu_{d}\|^{2} \label{eq:aQCQP}\\
   \text{s.t.}\quad&\left(\bar{\bolda}(\boldx)+\boldc_a(\boldx)\right)^{\top}\boldu+\bar{b}(\boldx)+c_b(\boldx) -\delta s \geq 0\\ &\boldu^{\top}\bSigma_{\bolda}(\boldx)\boldu+2\bSigma_{\bolda,b}(\boldx)^{\top}\boldu+\sigma_b(\boldx)^2 \leq s^{2} \label{eq: aconstraint_2}\\
   &s\geq 0
\end{align}
The above program is a non-convex QCQP. 
 \par
Then, by defining $\bar{\boldu}=\begin{bmatrix}\boldu & t &s\end{bmatrix}^{\top} \in \mathbb{R}^{m+2}$, we can write the non-convex quadratic constraint in \eqref{eq: aconstraint_2} as $\Vert \overline{\bSigma} \mathbf{\bar{\boldu}}\Vert \leq \boldc^{\top} \mathbf{\bar{\boldu}}$
with $t=1$, $\overline{\bSigma}=\sqrt{\begin{bmatrix} \bSigma_{\bolda}(\boldx) & \bSigma_{\bolda,b}(\boldx)&0 \\
\bSigma_{\bolda,b}(\boldx)&\sigma_b(\boldx)^2&0\\
0&0&0\end{bmatrix}}$ and $\boldc=\begin{bmatrix}\mathbf{0}\\0\\1\end{bmatrix}$.
We pose the QCQP as a convex program:
\begin{align*}
\min_{\mathbf{\bar{u}}}\quad&\text{ }\mathbf{\bar{u}^{\top}}\boldQ\mathbf{\bar{u}}
\\
\text{s.t.}\quad&\Vert \overline{\bSigma} \mathbf{\bar{u}}\Vert \leq \boldc^{\top} \mathbf{\bar{u}},\quad
\boldC^{\top}\mathbf{\bar{u}}+\boldd\leq 0,\quad
\boldD^{\top}\mathbf{\bar{u}}+\boldf= 0
\end{align*}
Note that the constraint:
\begin{align*}
    \quad&\Vert \overline{\bSigma} \mathbf{\bar{u}}\Vert \leq \boldc^{\top} \mathbf{\bar{u}}
\end{align*}
is equivalent to:
\begin{align*}
   \quad&\Vert \overline{\bSigma} \mathbf{\bar{u}}\Vert^{2} \leq s^{2} 
\end{align*}
as $s\geq0$. Further:
\begin{align}
    \Vert \overline{\bSigma} \mathbf{\bar{u}}\Vert^{2}&= \boldu^{\top}\bSigma_{a}(\boldx)\boldu+2t\bSigma_{a,b}(\boldx)^{\top}\boldu+t^{2}\sigma_b(\boldx)^2
\end{align}
with the additional constraint $t=1$. This equality constraint can be enforced by choosing 
\begin{align}
\boldD&=\begin{bmatrix}
\mathbf{0}\\1\\0
\end{bmatrix},  \boldf=\begin{bmatrix}\mathbf{0}\\-1\\0\end{bmatrix}.
\end{align}
The remaining inequality constraints can be enforced by choosing $\boldC$, $\boldd$ as follows:
\begin{align}
    \boldC=\begin{bmatrix}
    -\left(\bar{\bolda}(\boldx)+\boldc_a(\boldx)\right)^{\top}& -(\bar{b}(\boldx)+\boldc_b(\boldx))& \delta\\
    0 & 0 & -1\\
    \end{bmatrix}, \boldd=\mathbf{0}. 
\end{align}
Finally, $\boldQ$ can be defined  to recover the original objective function as:
\begin{align}
    \boldQ = \begin{bmatrix}
    \boldI&-\boldu_{d}&\bold0\\-\boldu_{d}^{\top}&\boldu_{d}^{\top}\boldu_{d}&\bold0\\\bold0&\bold0&\bold0
    \end{bmatrix}.
\end{align}
\section{Additional Experiment Results}\label{sec:appendix_experiment}
\begin{figure}[ht!]
	\centering
	\includegraphics[ width=0.32\columnwidth]{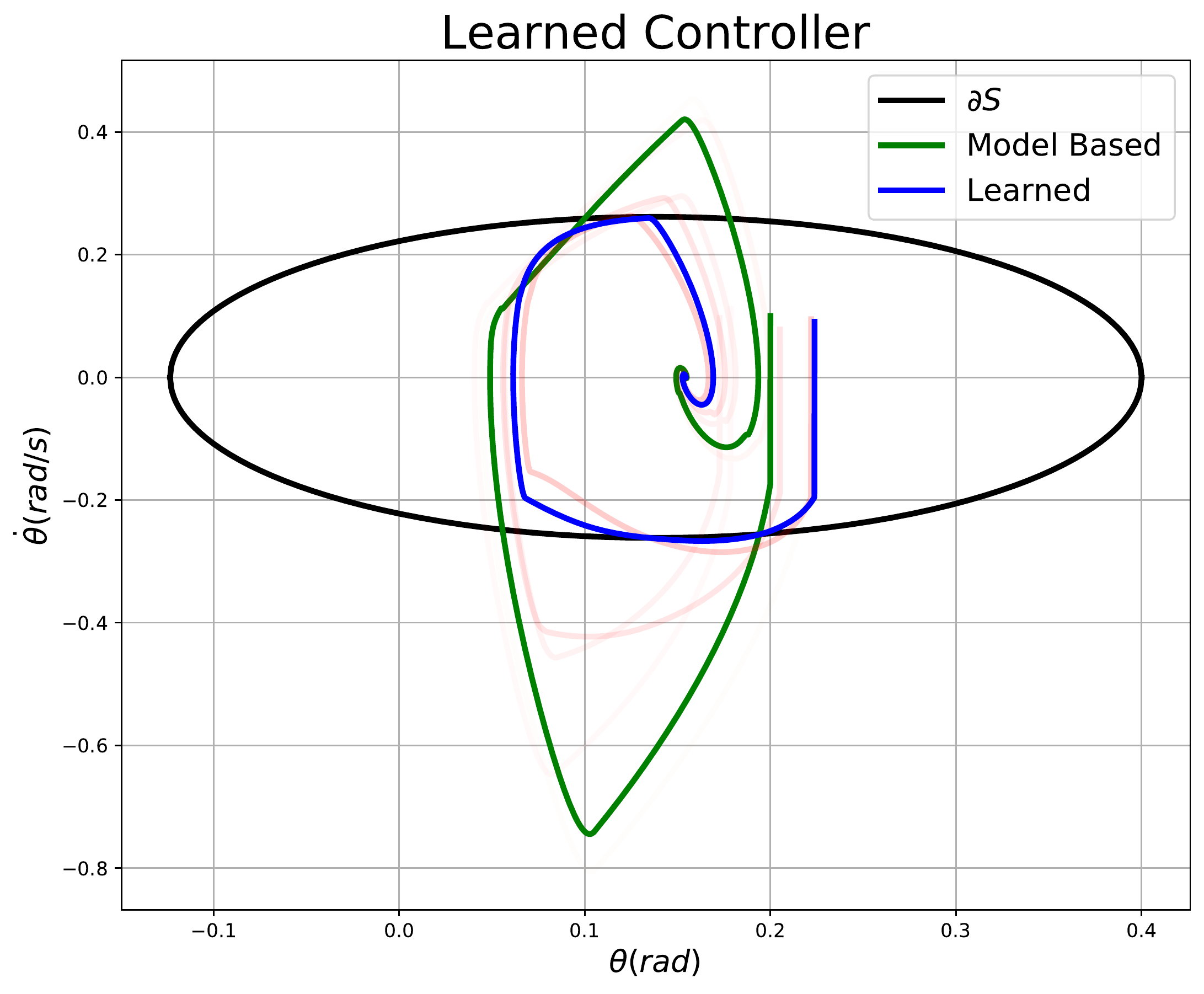}
	\includegraphics[ width=0.32\columnwidth]{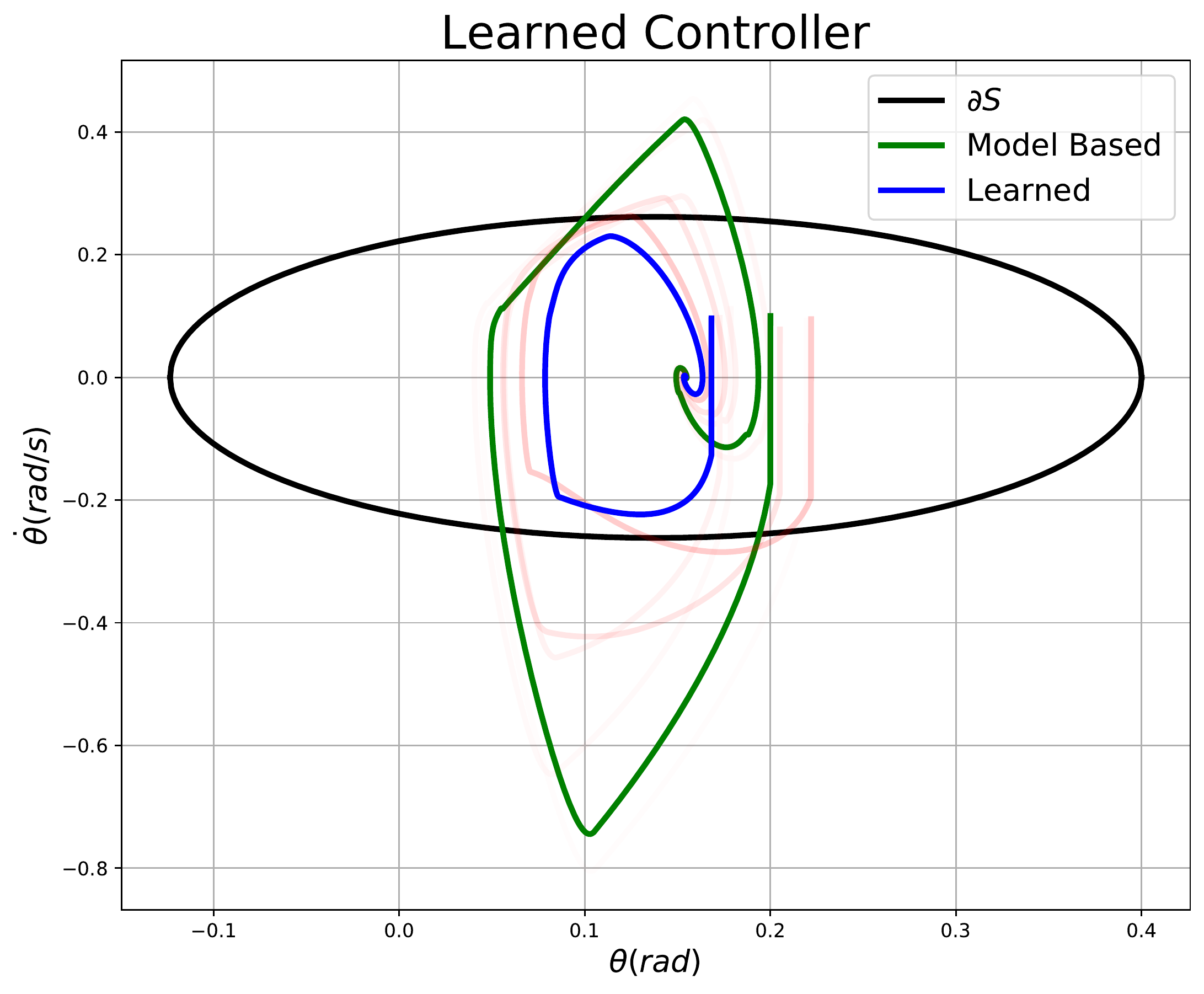}
	\includegraphics[
	width=0.32\columnwidth]{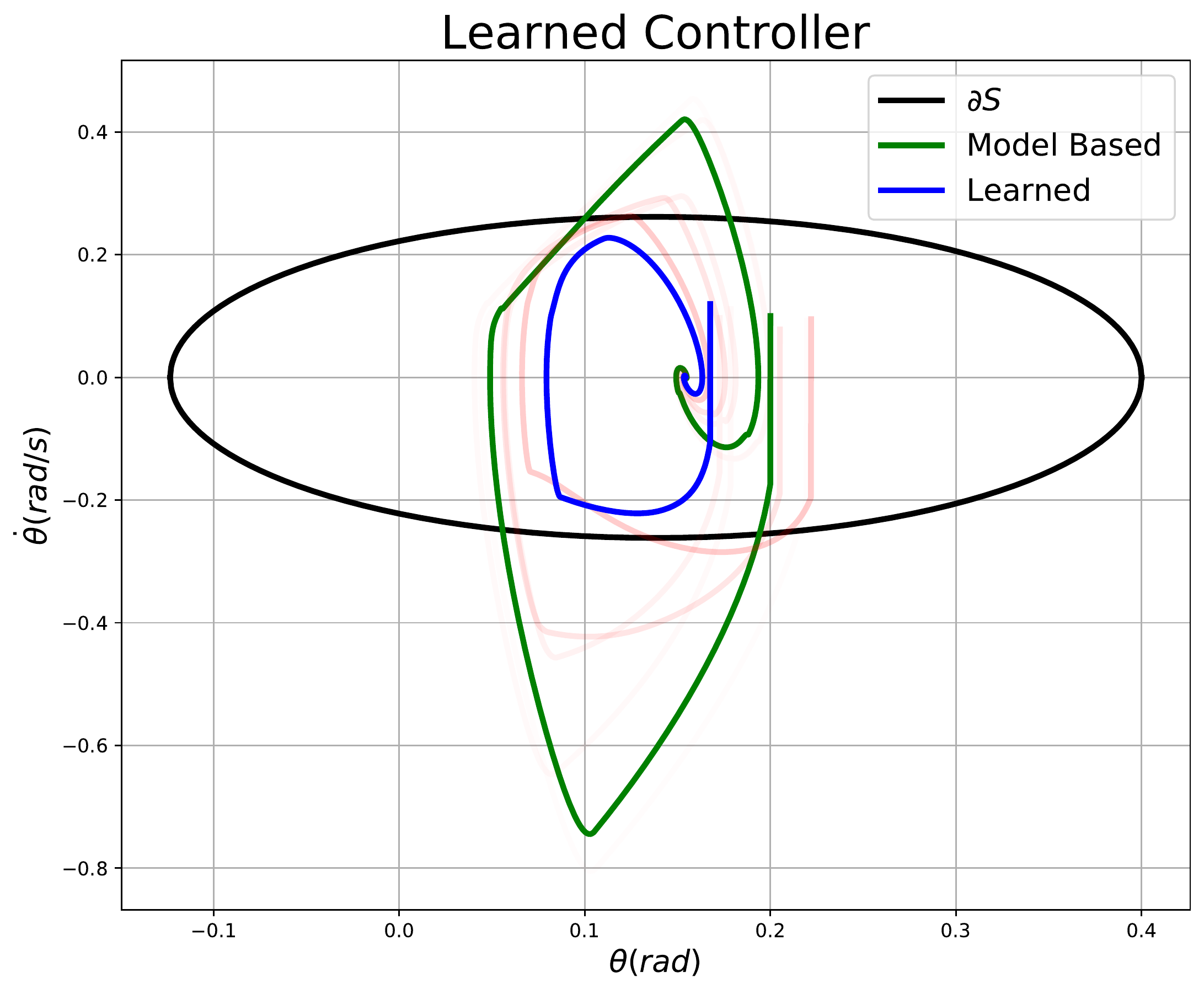}
	\includegraphics[ width=0.32\columnwidth]{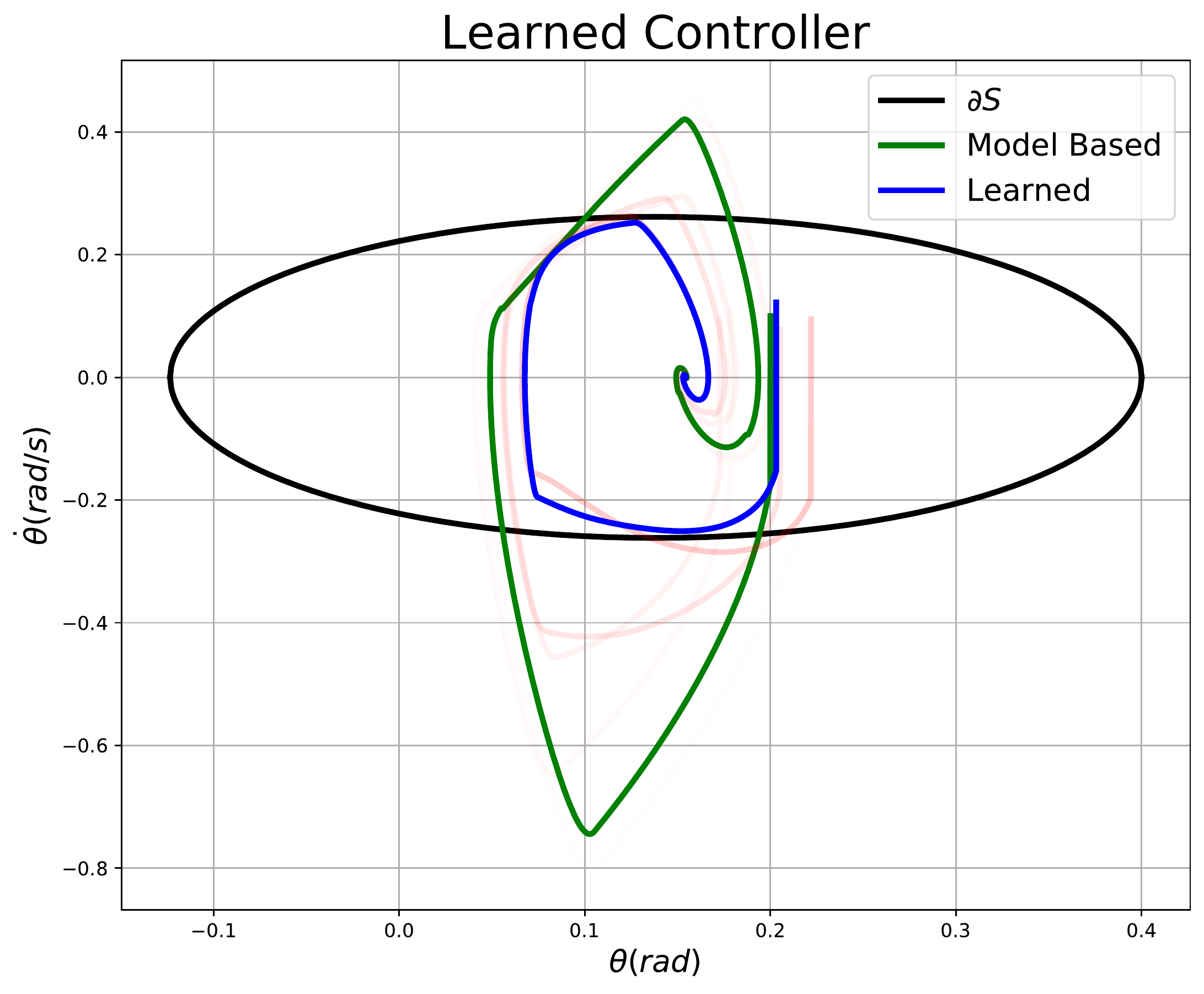}
	\includegraphics[ width=0.32\columnwidth]{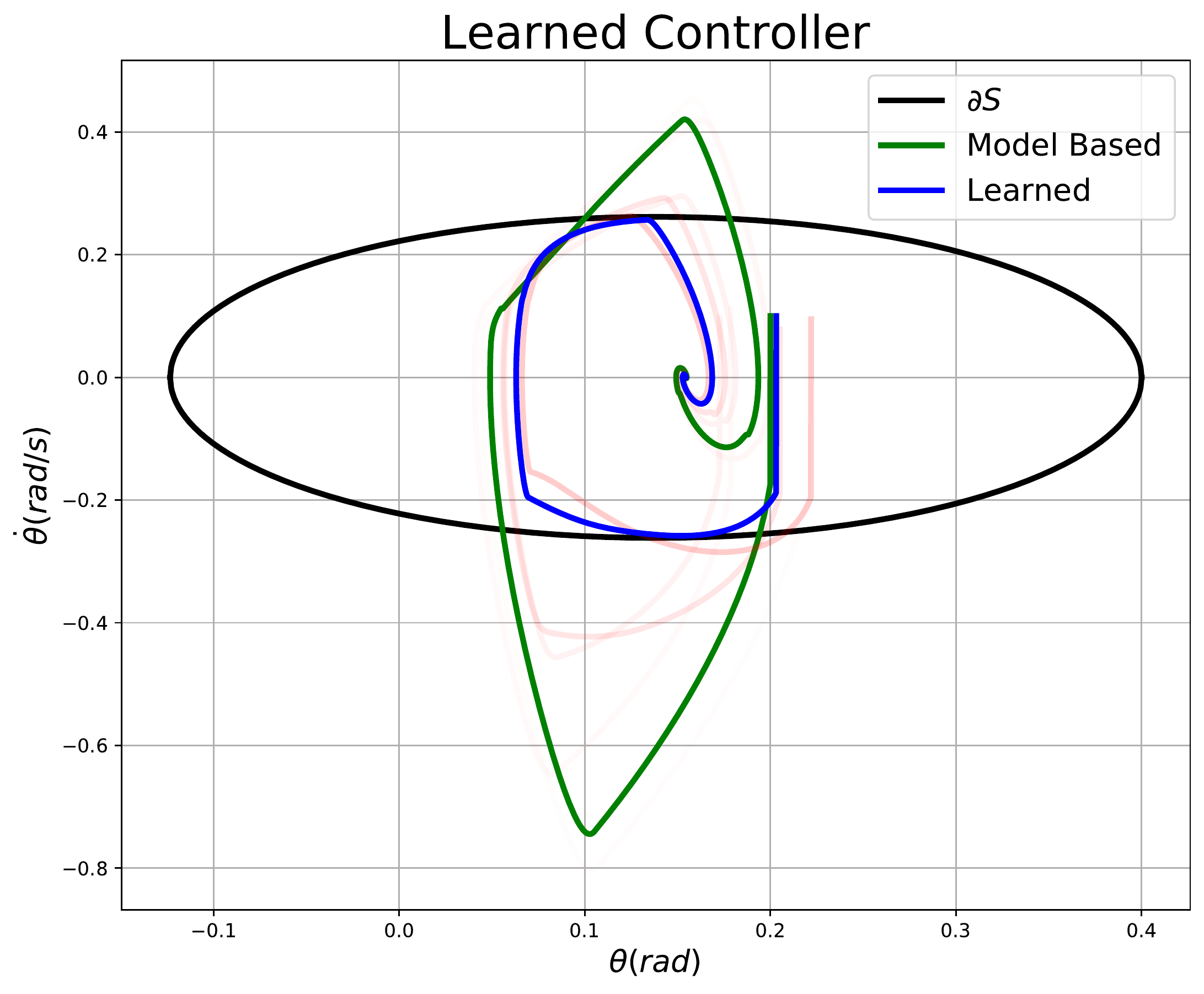}
	\includegraphics[ width=0.32\columnwidth]{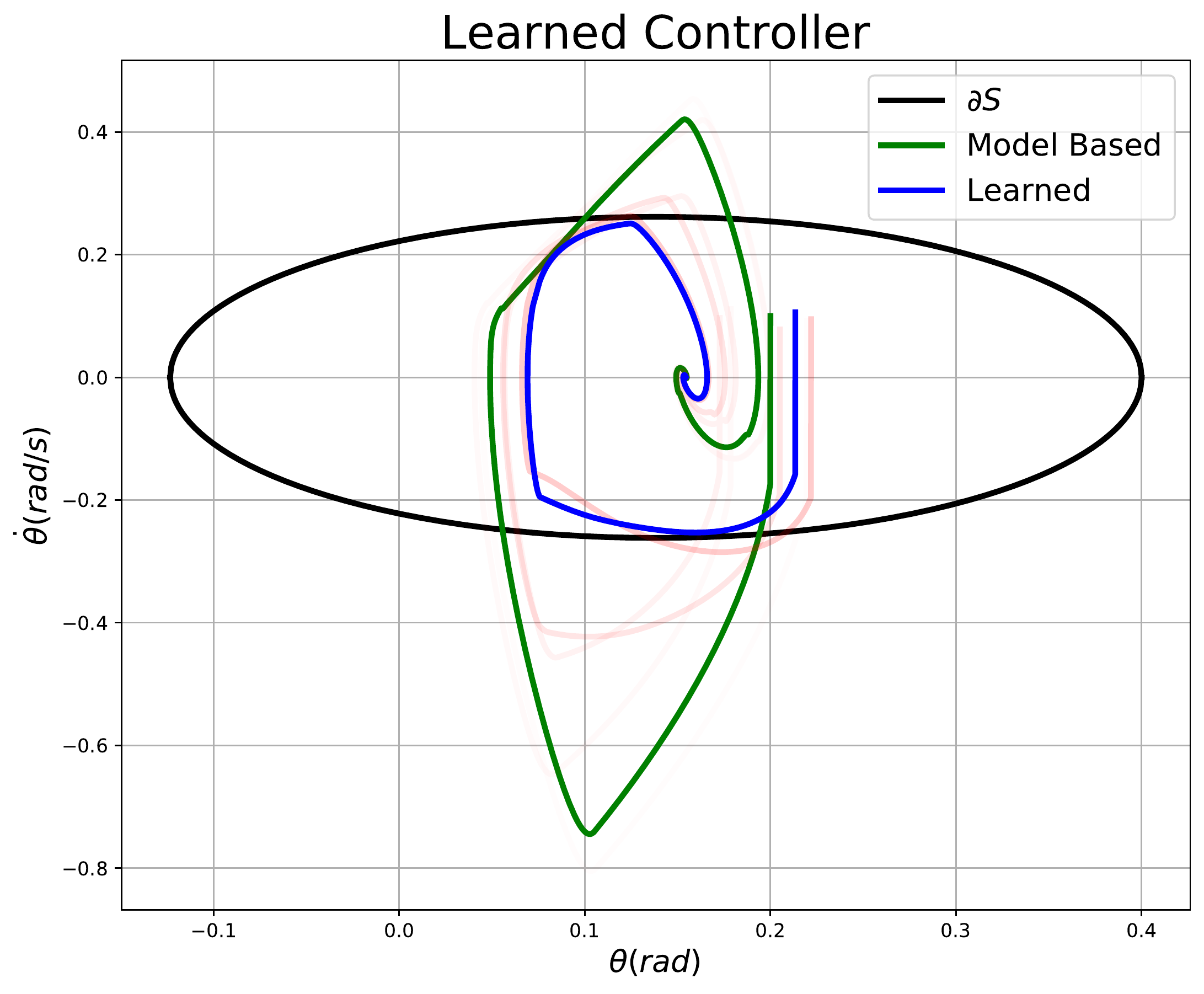}		
	\includegraphics[ width=0.32\columnwidth]{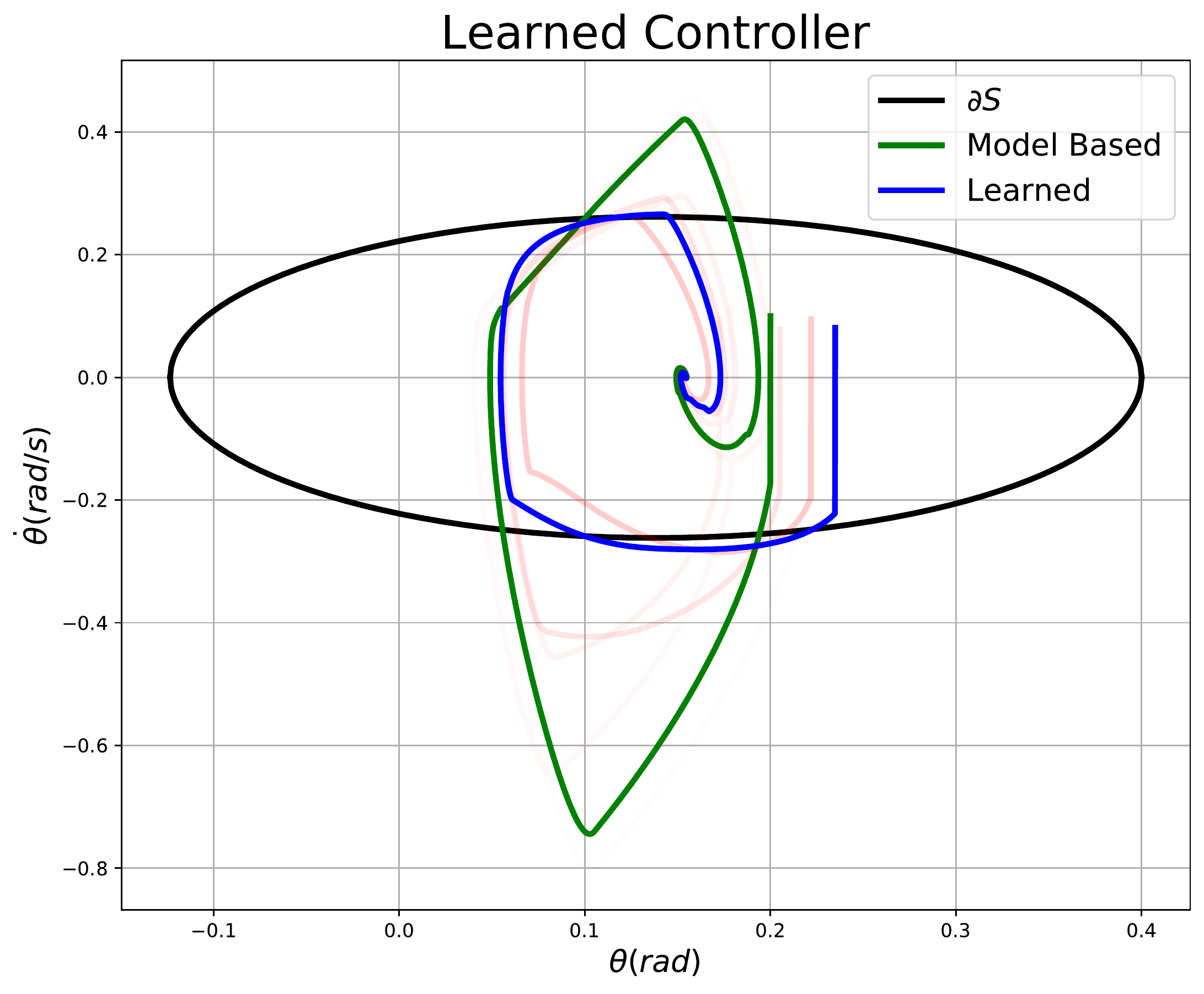}
	\includegraphics[ width=0.32\columnwidth]{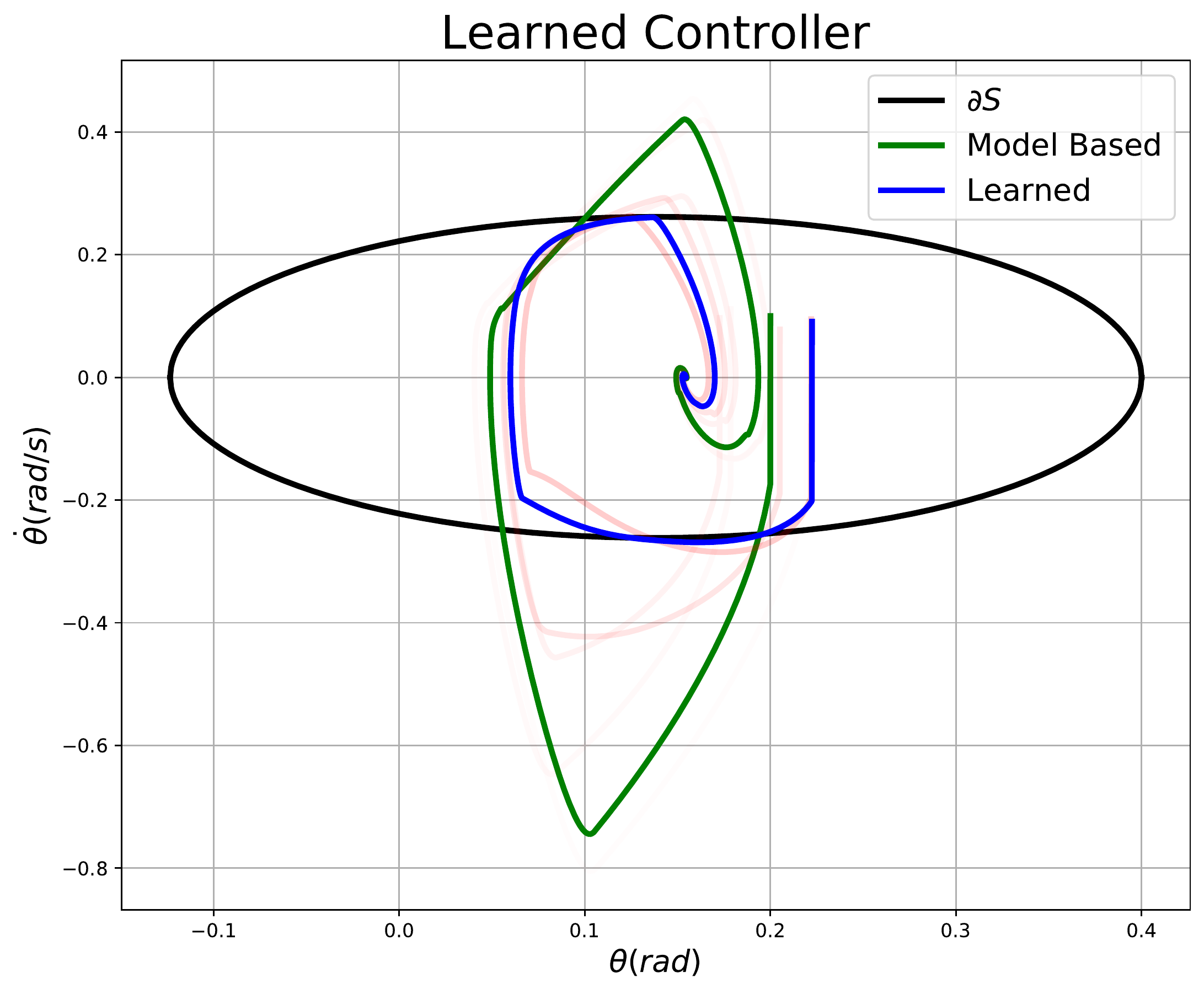}
	\includegraphics[ width=0.32\columnwidth]{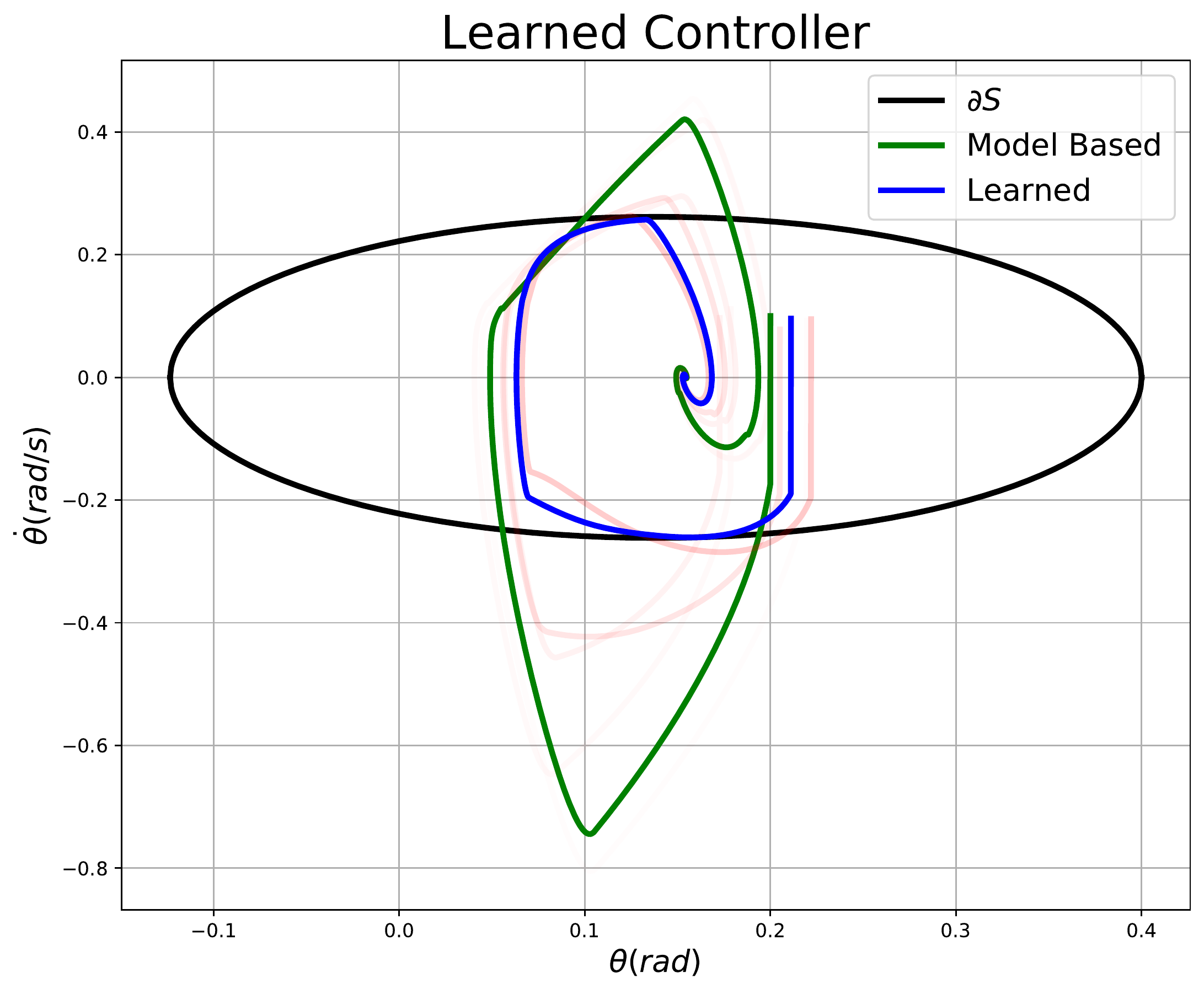}	
	\includegraphics[ width=0.32\columnwidth]{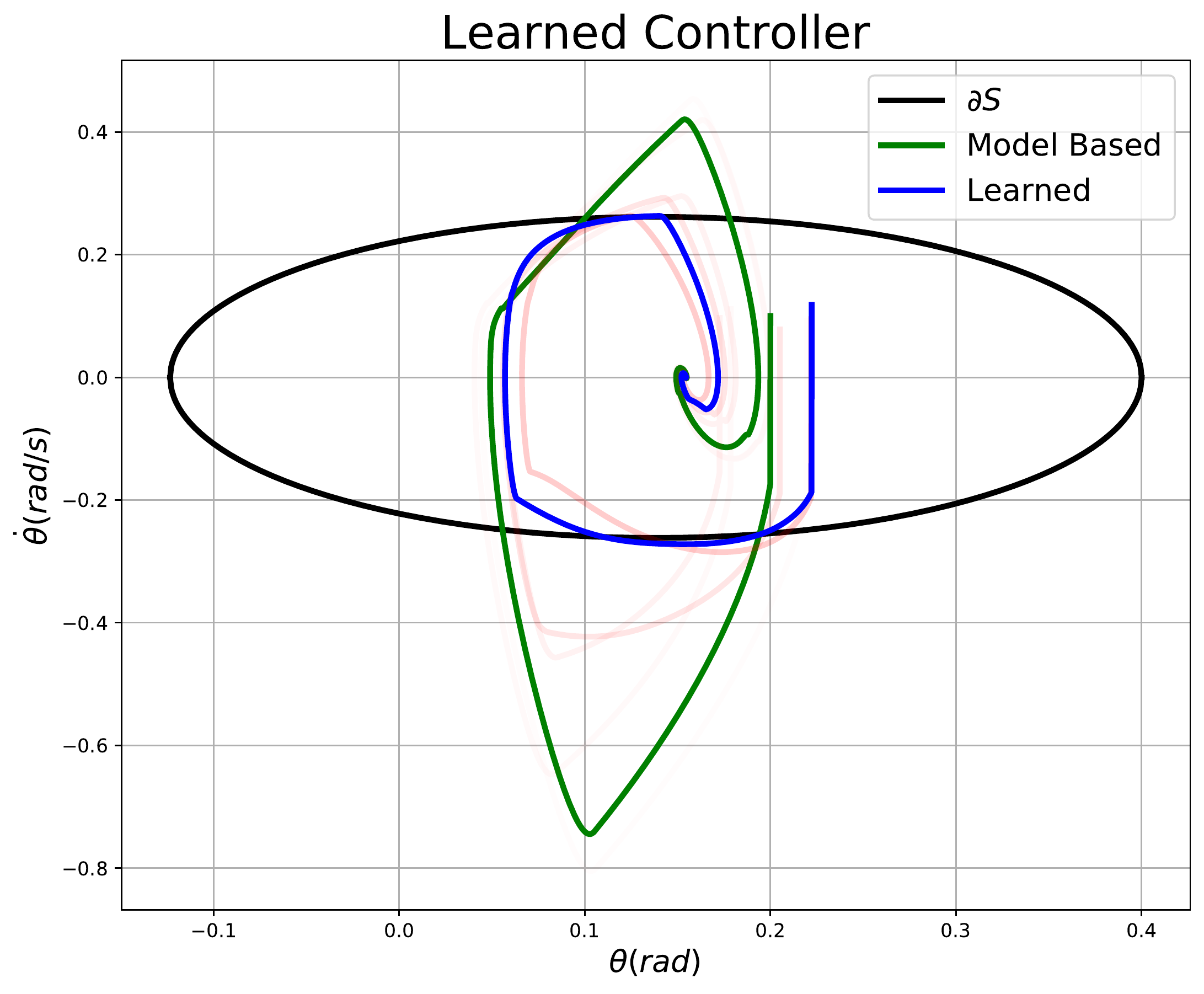}			
	\vspace{-5pt}
	\caption{Segway-LCBF-QP with neural network: one random training run, tested on 10 random initial points}\label{fig:NN_test_samples}
	\vspace{3pt}
	\vspace{-0.4cm}
	\label{fig:snn_seed123}
\end{figure}

\begin{figure}[ht!]
	\centering
	\includegraphics[ width=0.32\columnwidth]{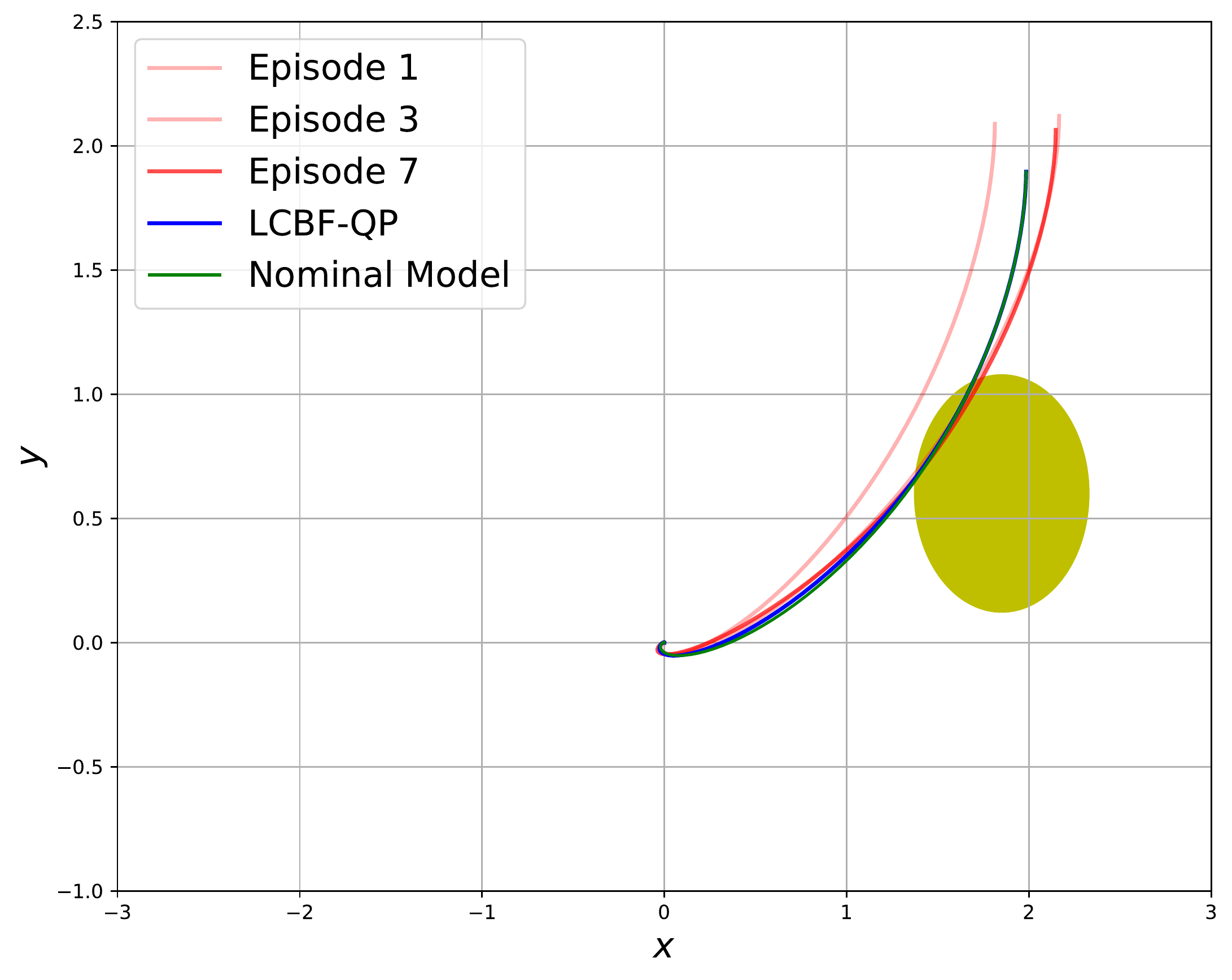}
	\includegraphics[ width=0.32\columnwidth]{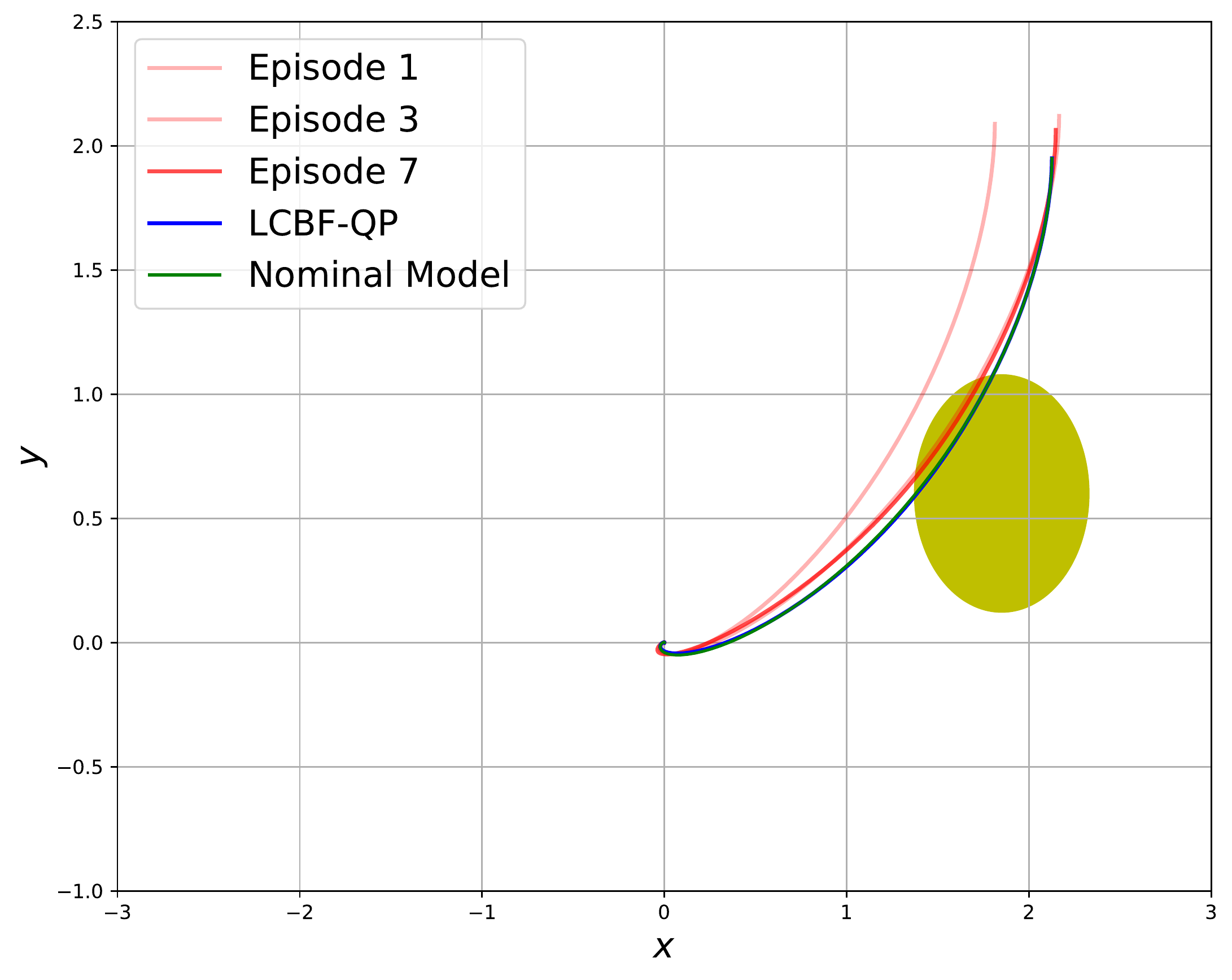}
	\includegraphics[ width=0.32\columnwidth]{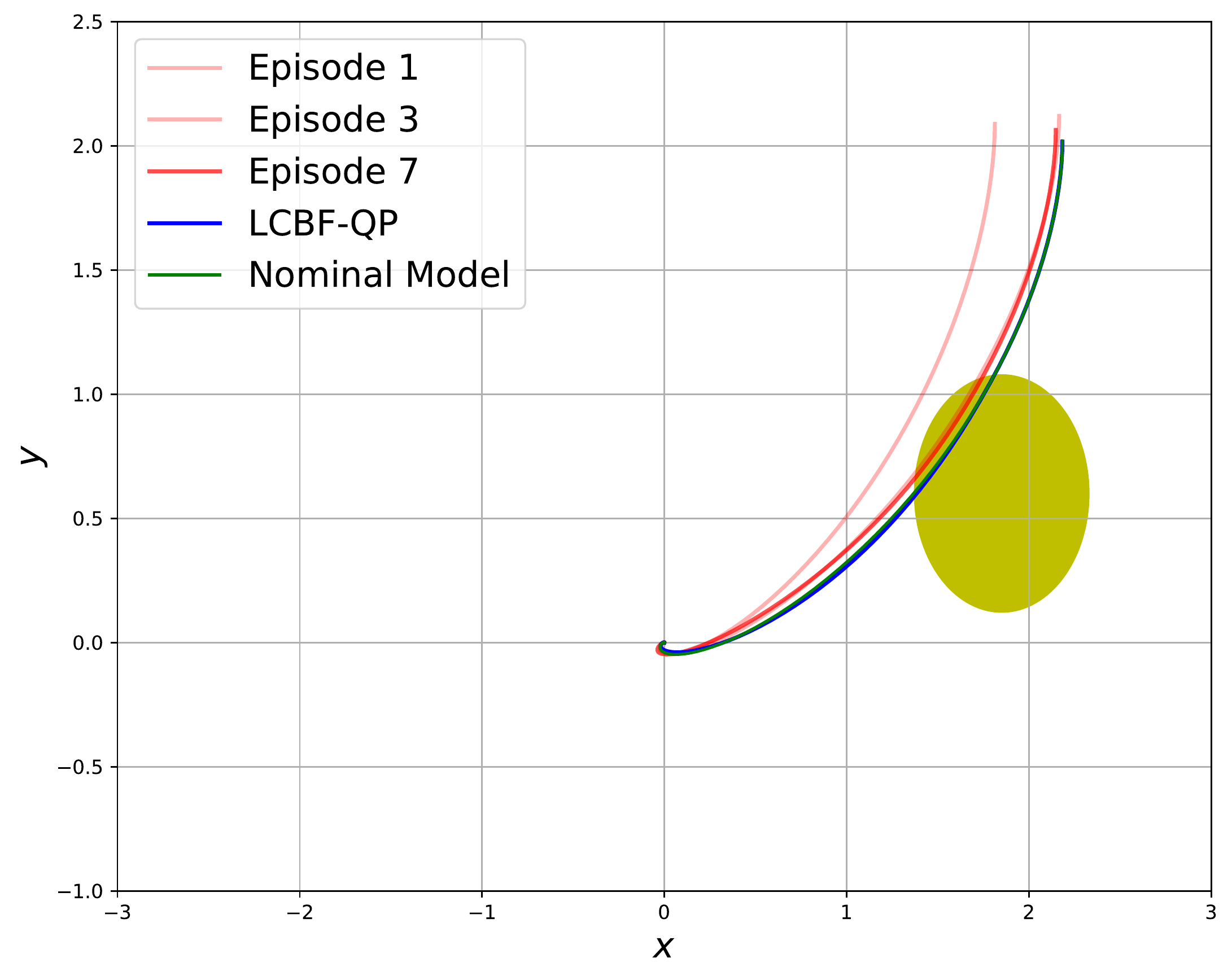}
	\includegraphics[ width=0.32\columnwidth]{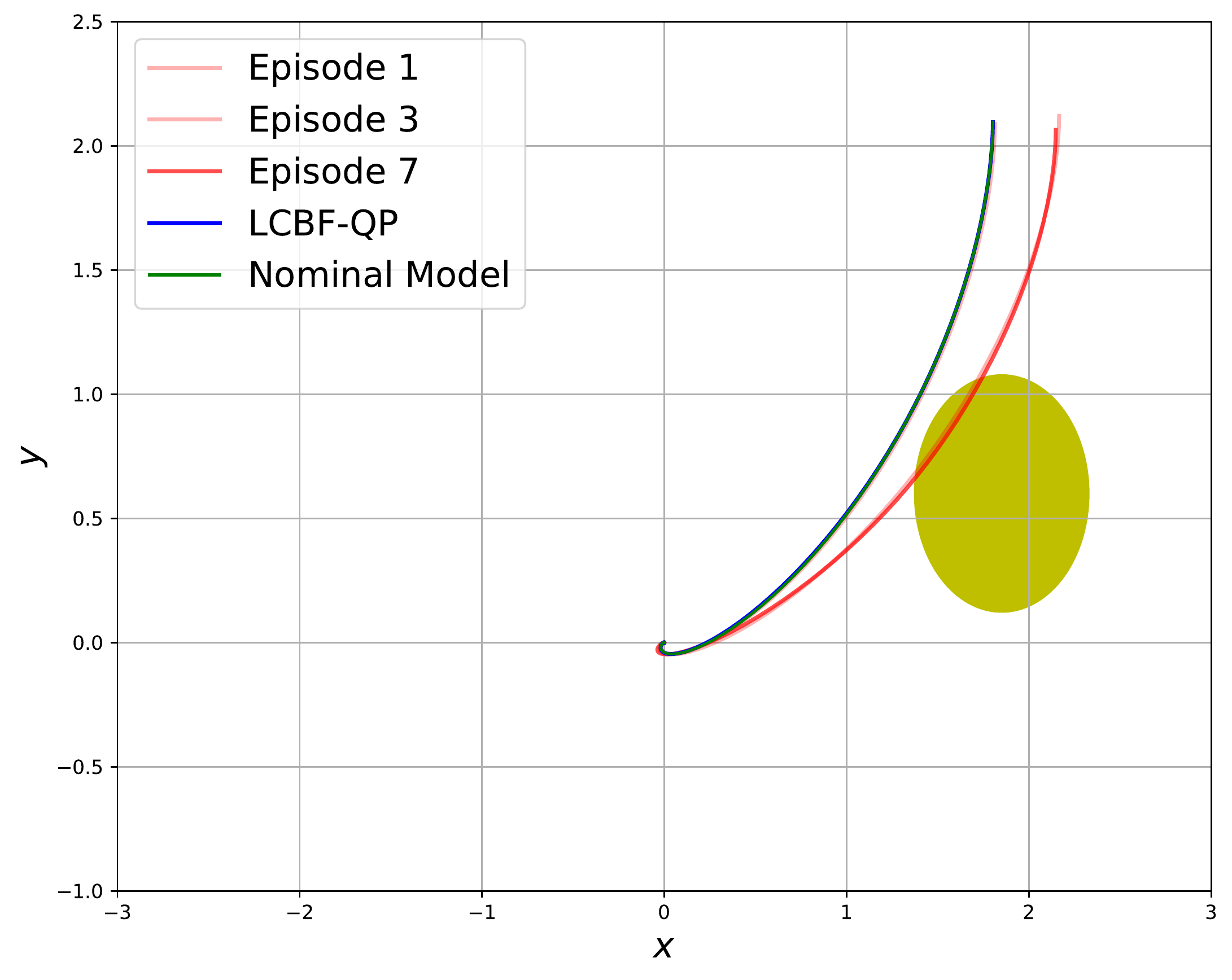}
	\includegraphics[ width=0.32\columnwidth]{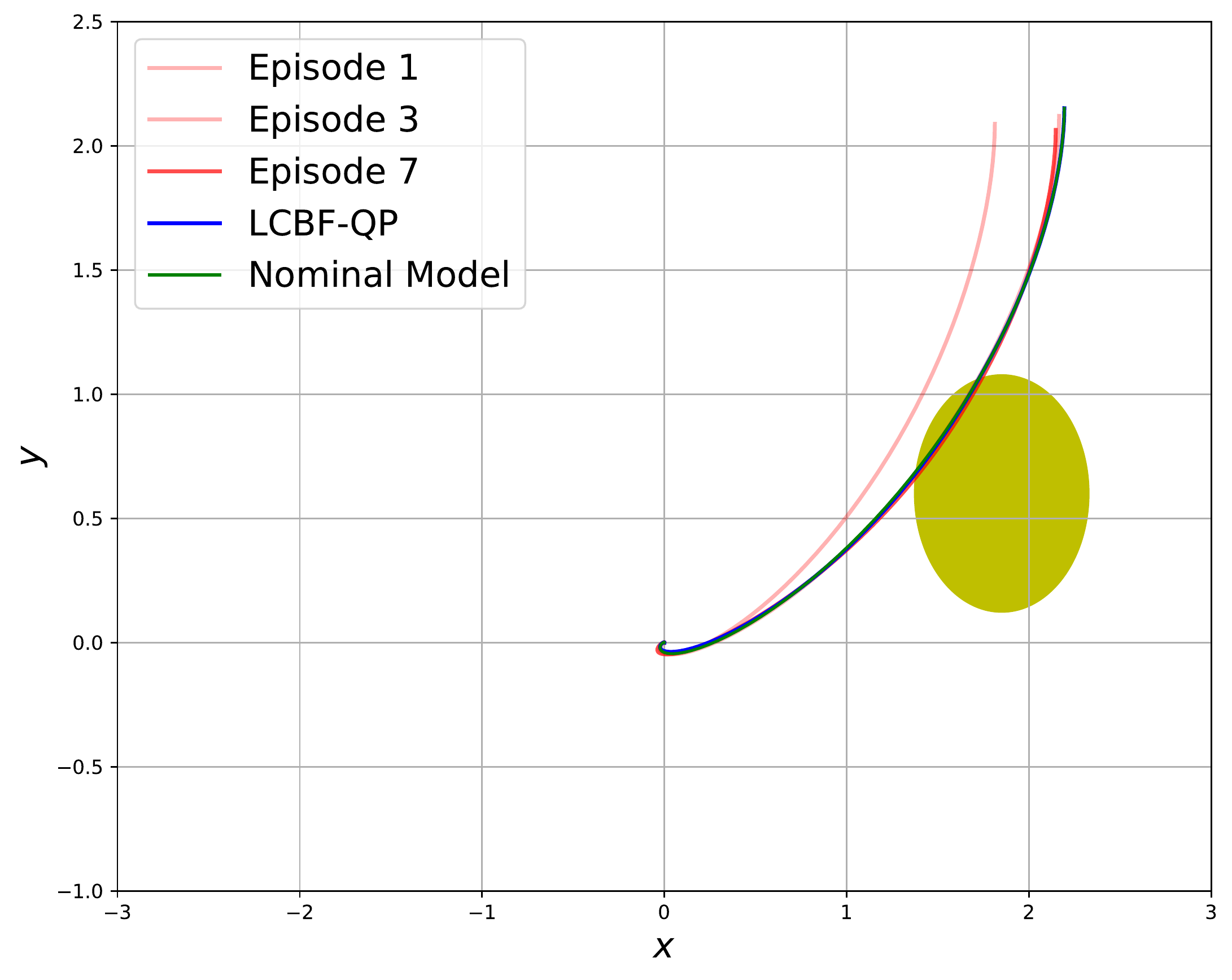}
	\includegraphics[ width=0.32\columnwidth]{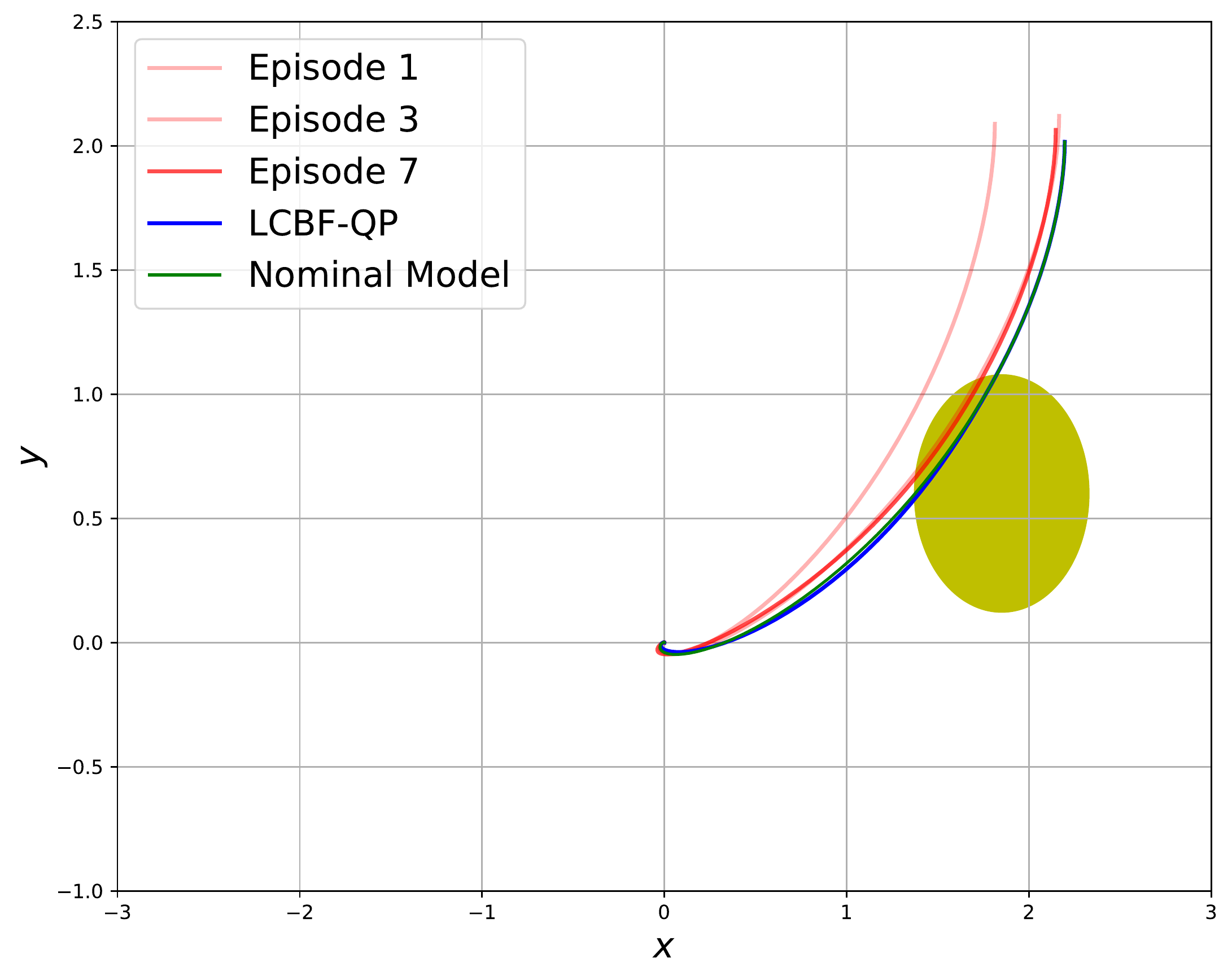}		
	\includegraphics[ width=0.32\columnwidth]{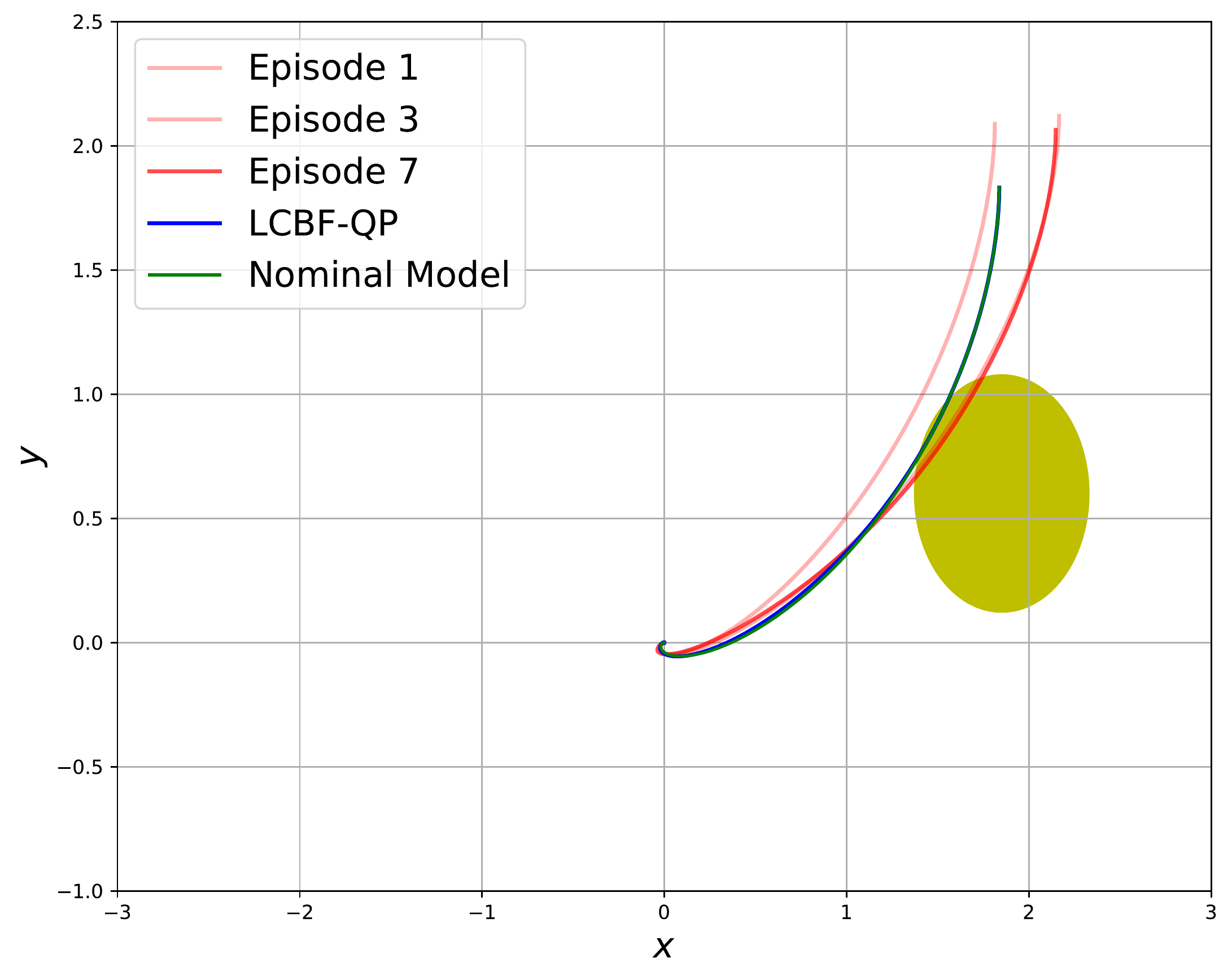}
	\includegraphics[ width=0.32\columnwidth]{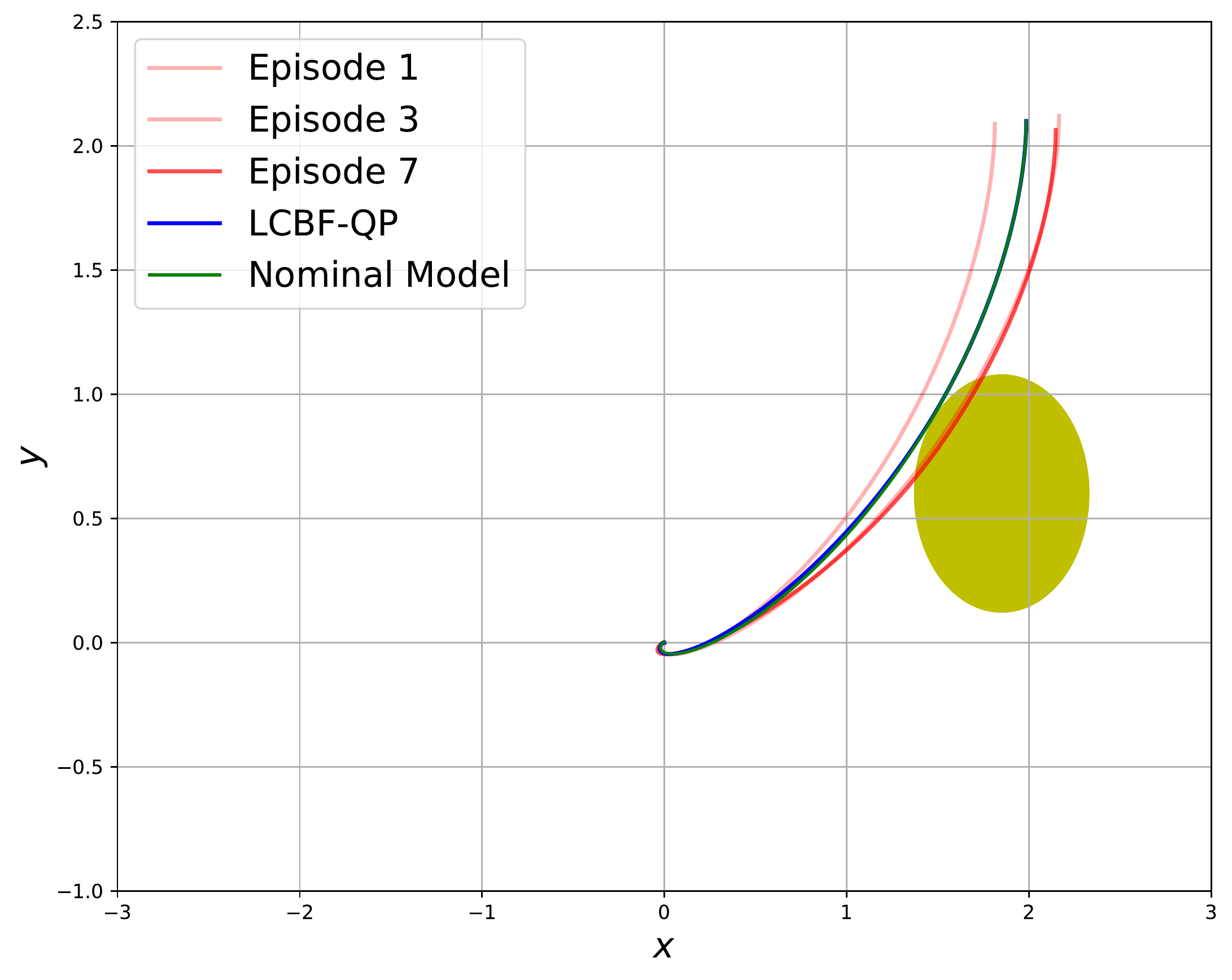}
	\includegraphics[ width=0.32\columnwidth]{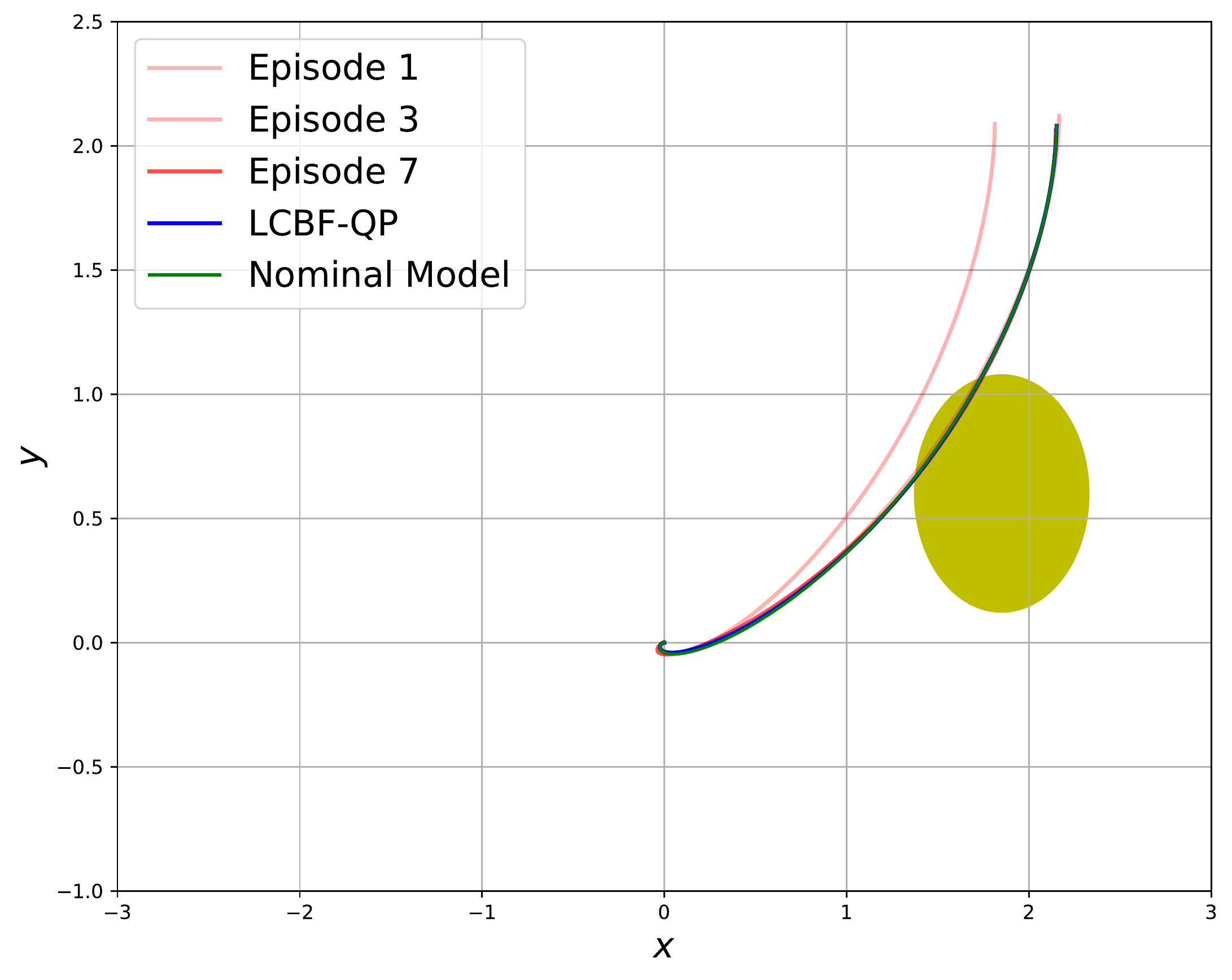}	
	\includegraphics[ width=0.32\columnwidth]{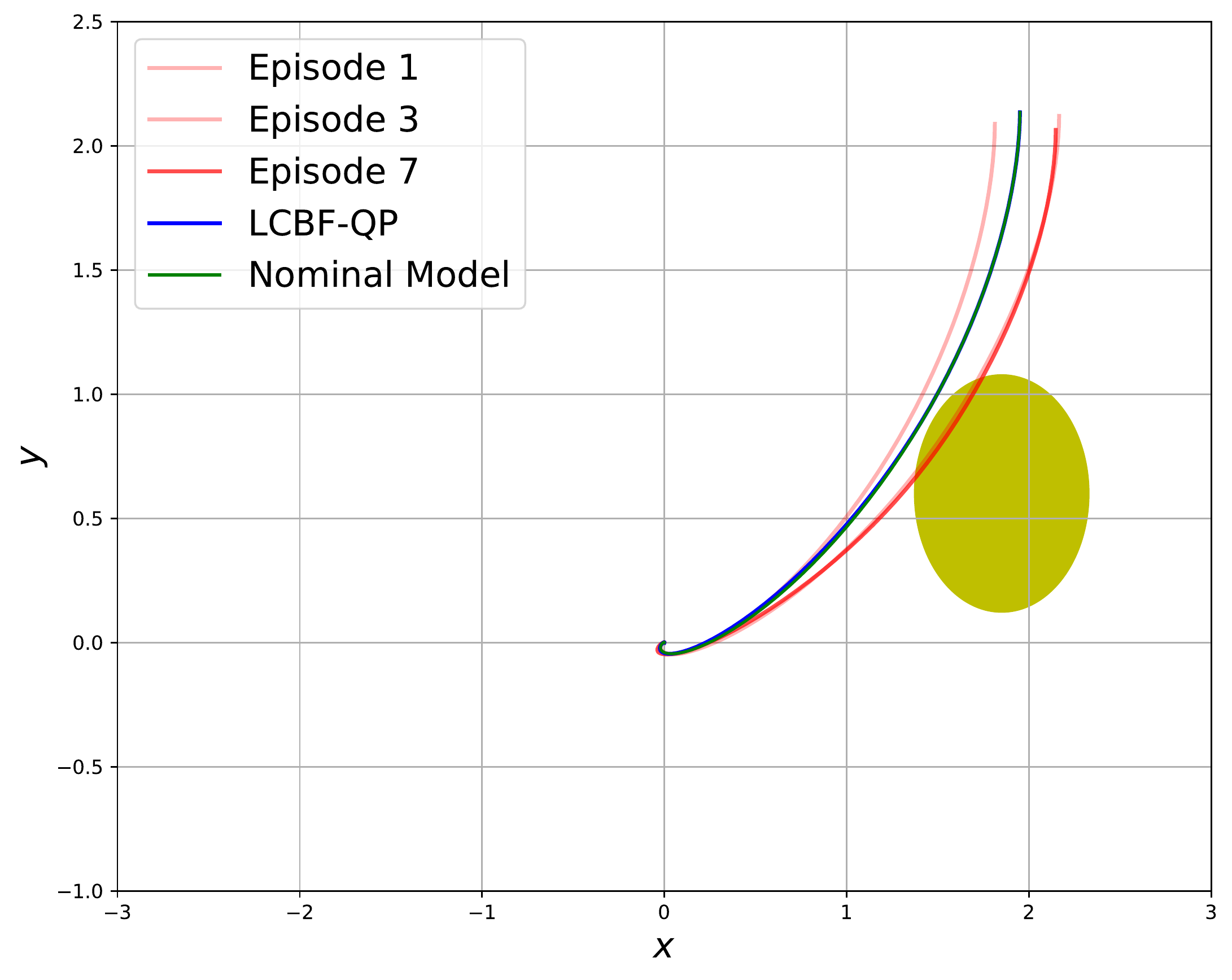}			
	\vspace{-5pt}
	\caption{Quadrotor-LCBF-QP with neural network: one random training run, tested on 10 random initial points}\label{fig:NN_quad_samples}
	\vspace{3pt}
	\vspace{-0.4cm}
	\label{fig:qnn_seed123}
\end{figure}

\begin{figure}[ht!]
	\centering
	\includegraphics[ width=0.32\columnwidth]{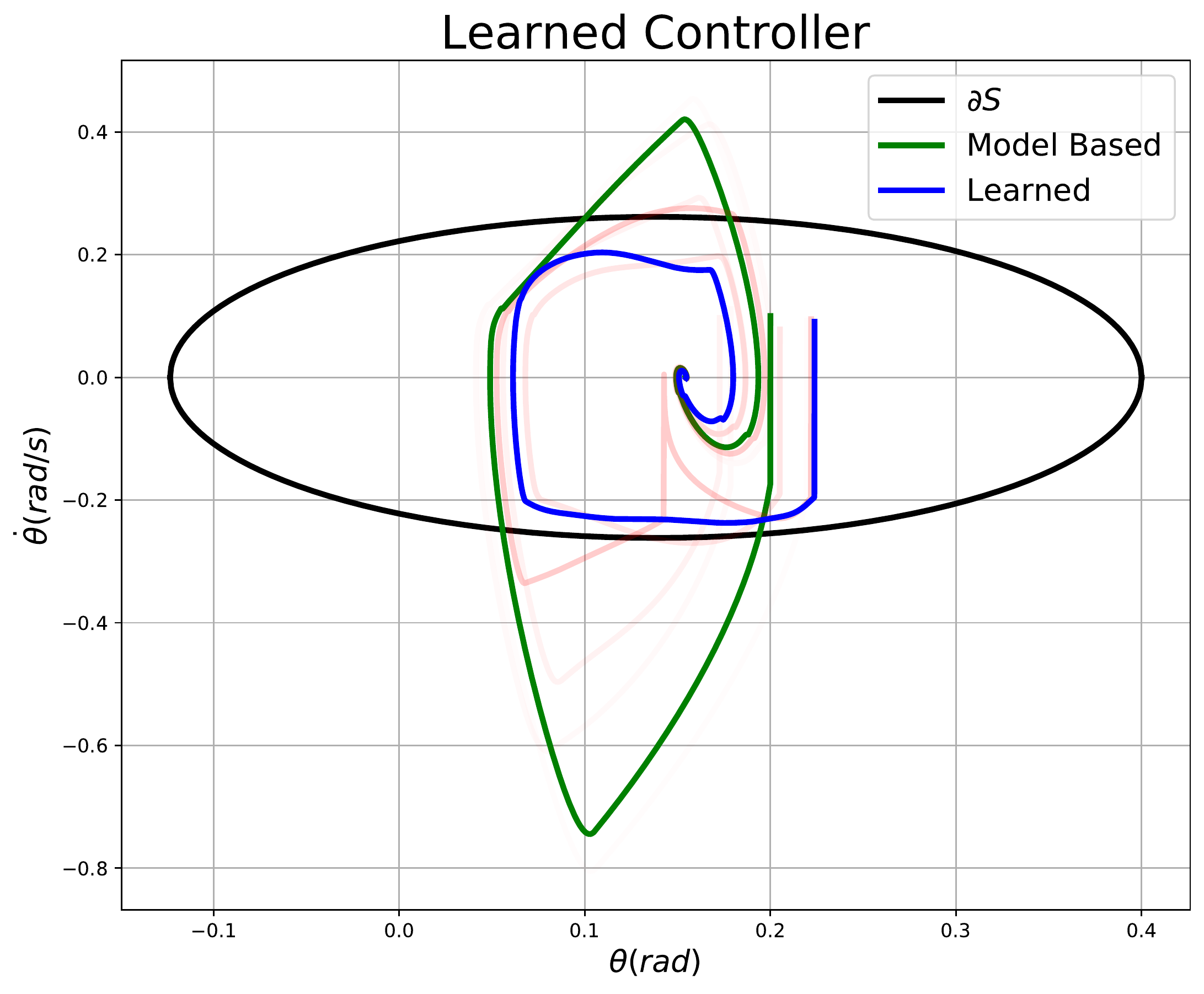}
	\includegraphics[ width=0.32\columnwidth]{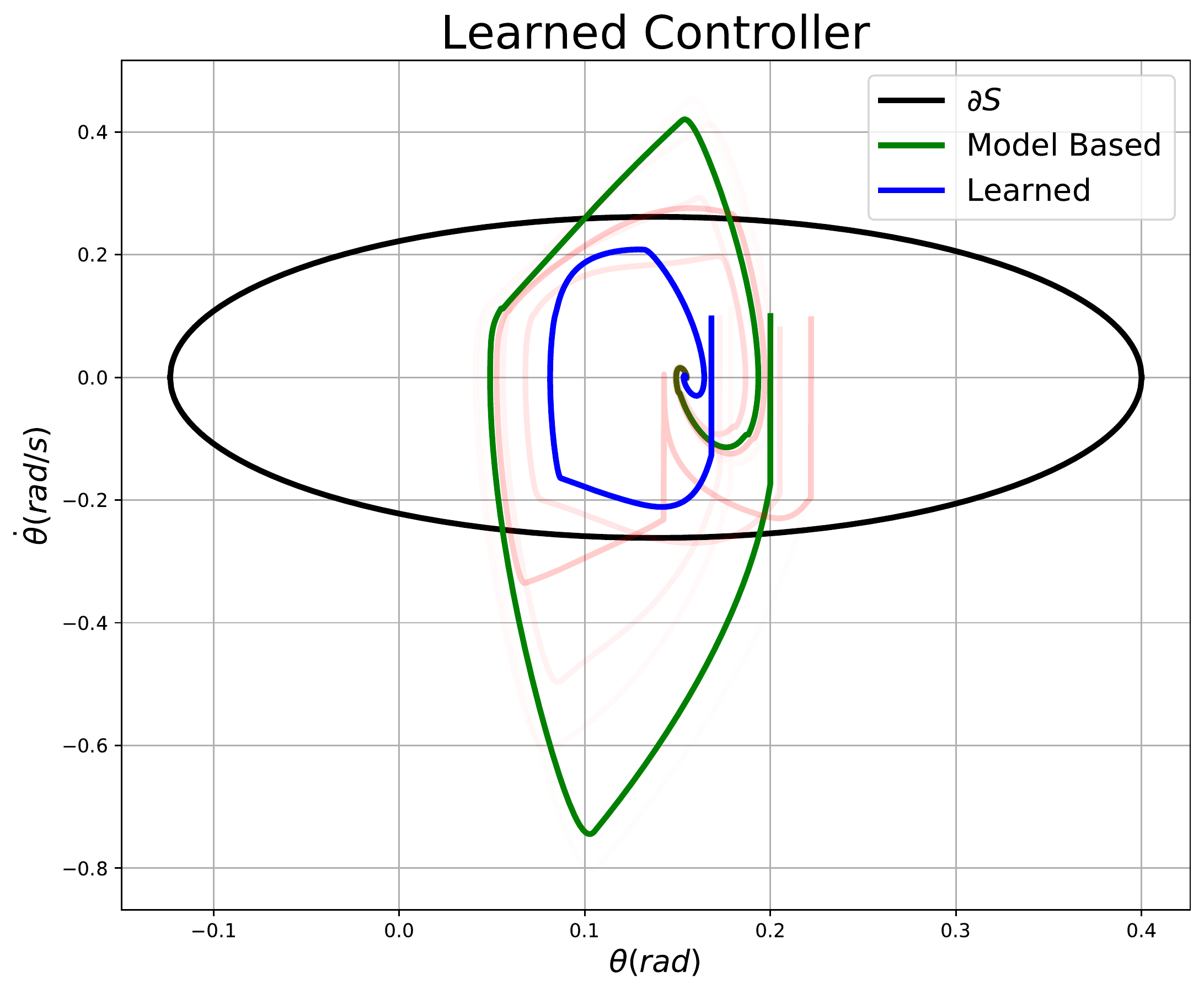}
	\includegraphics[
	width=0.32\columnwidth]{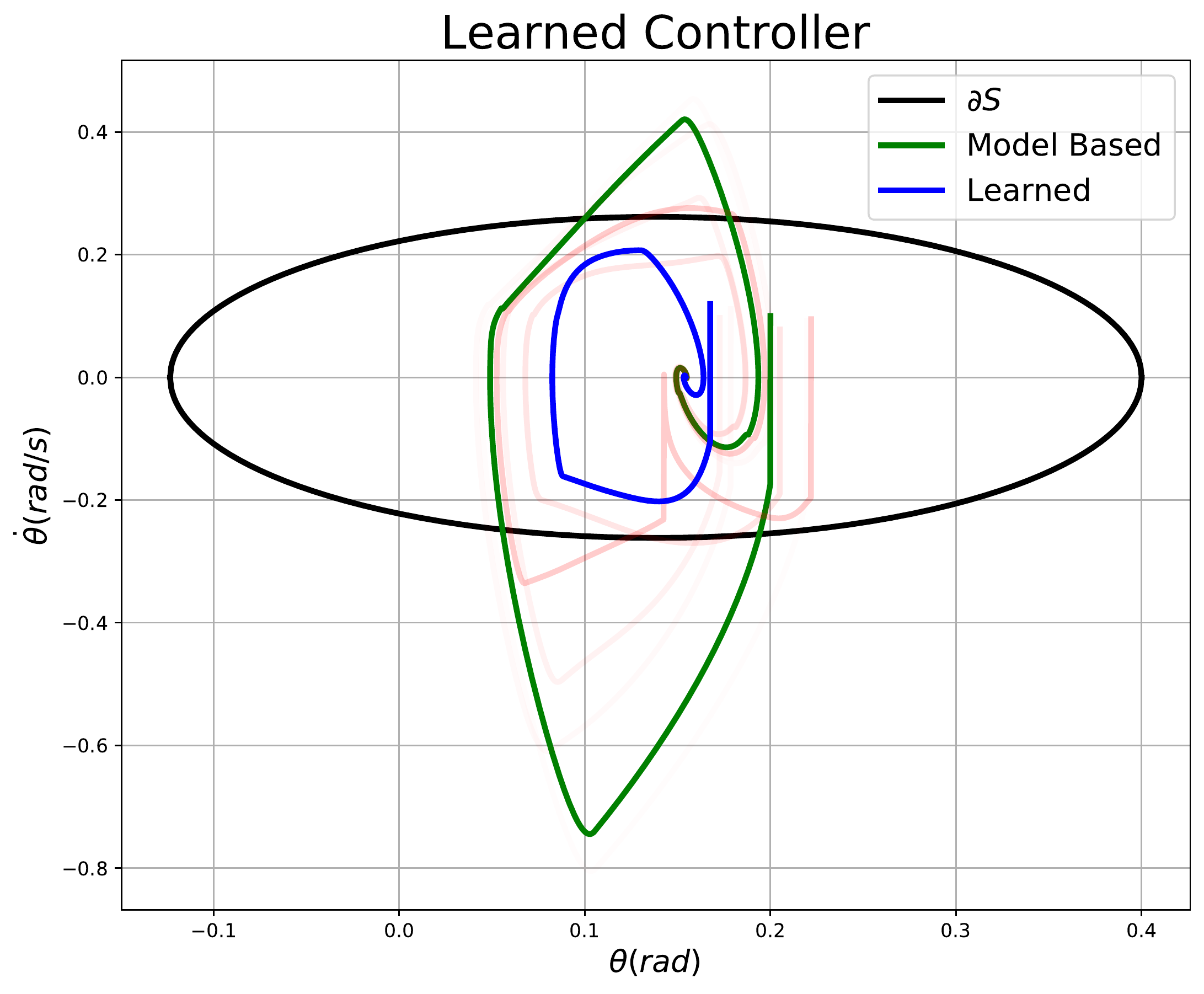}
	\includegraphics[ width=0.32\columnwidth]{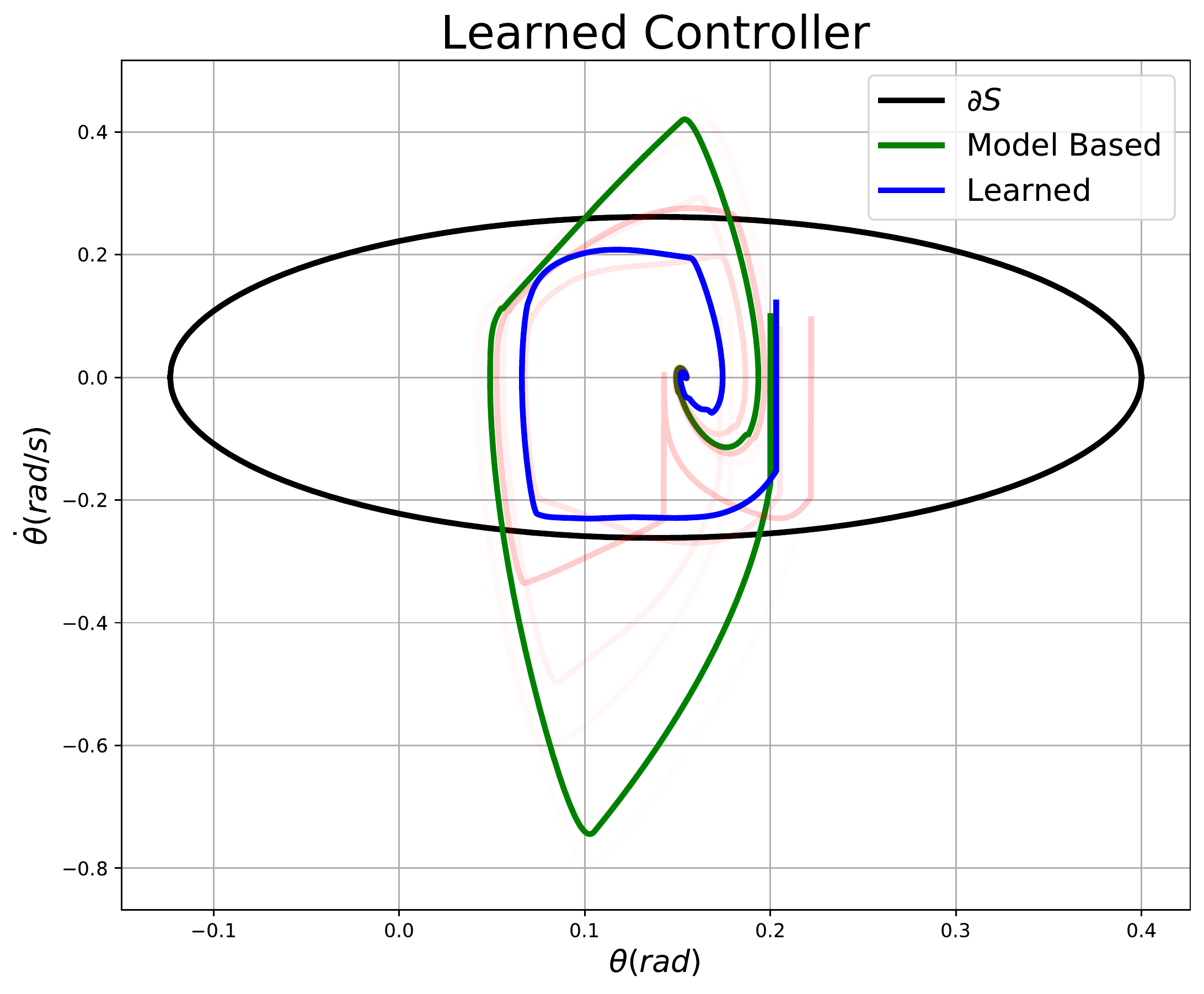}
	\includegraphics[ width=0.32\columnwidth]{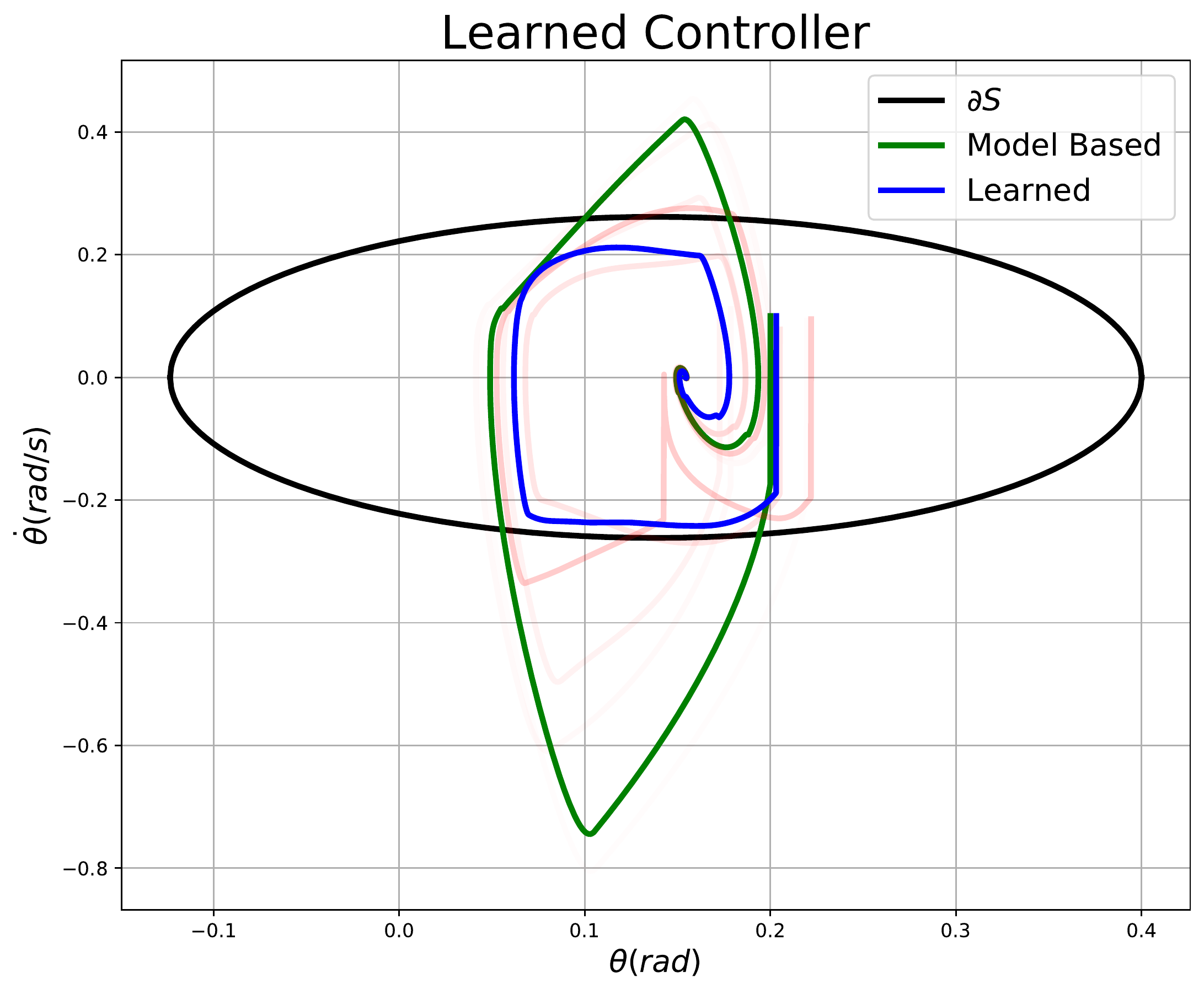}
	\includegraphics[ width=0.32\columnwidth]{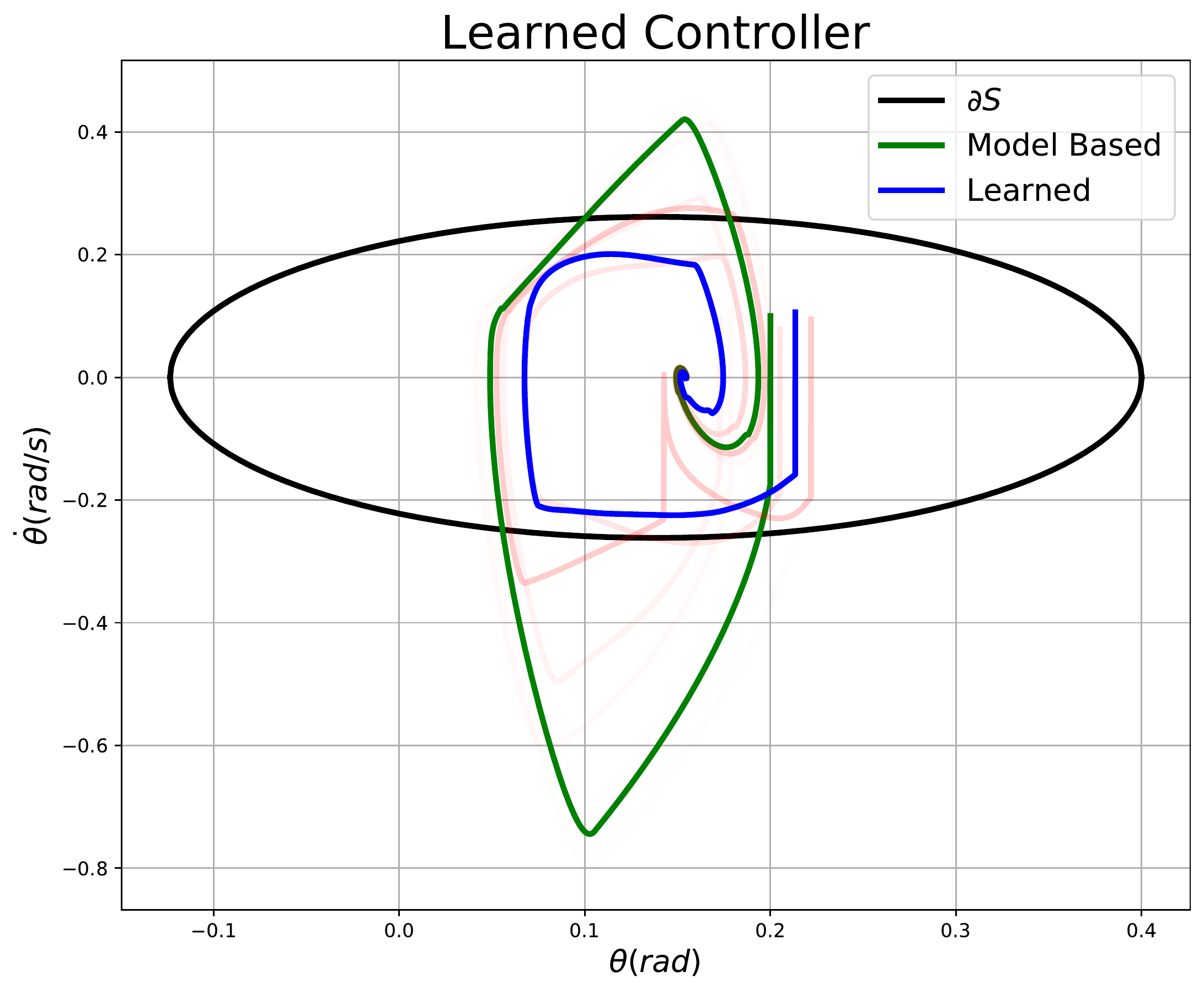}		
	\includegraphics[ width=0.32\columnwidth]{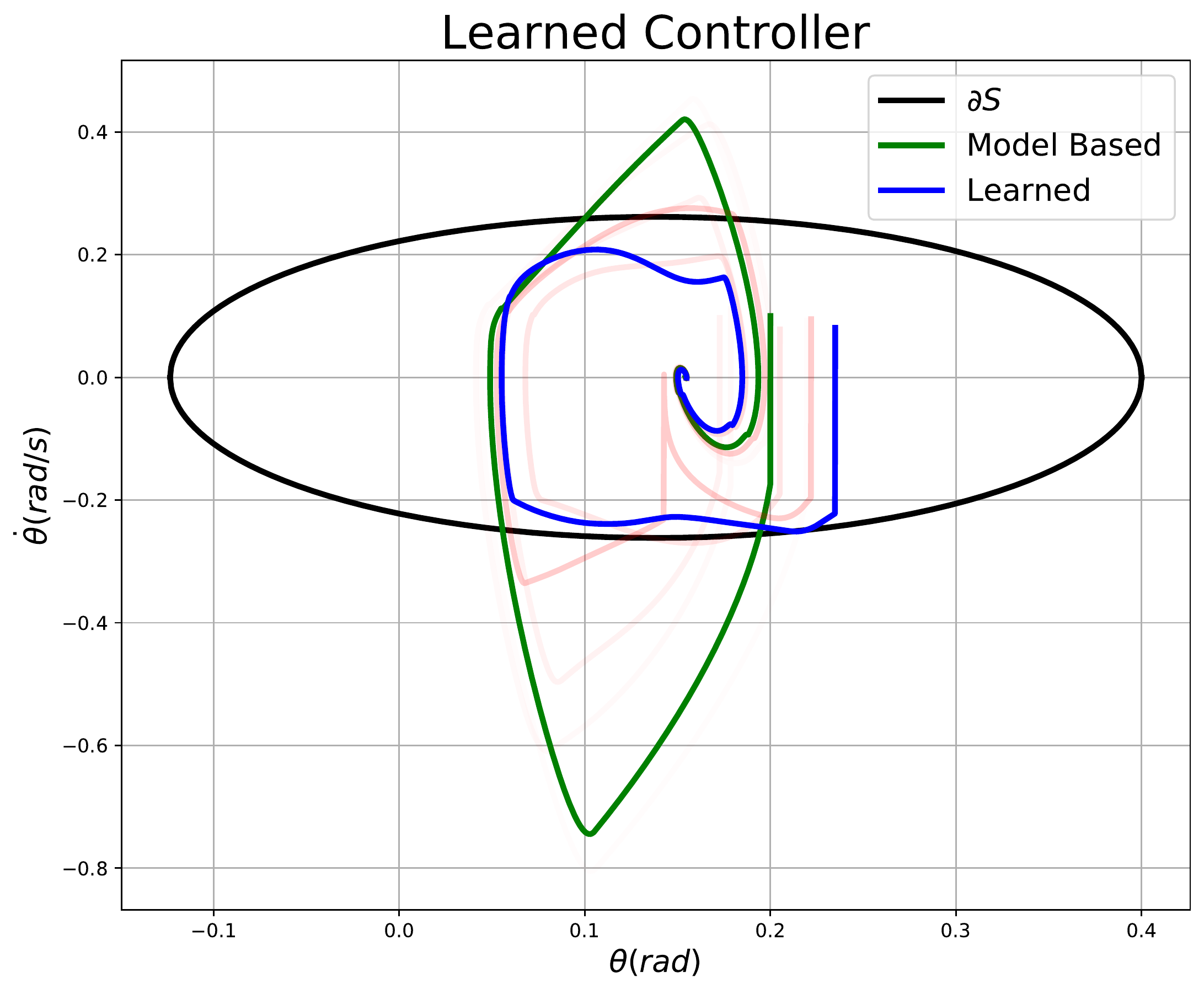}
	\includegraphics[ width=0.32\columnwidth]{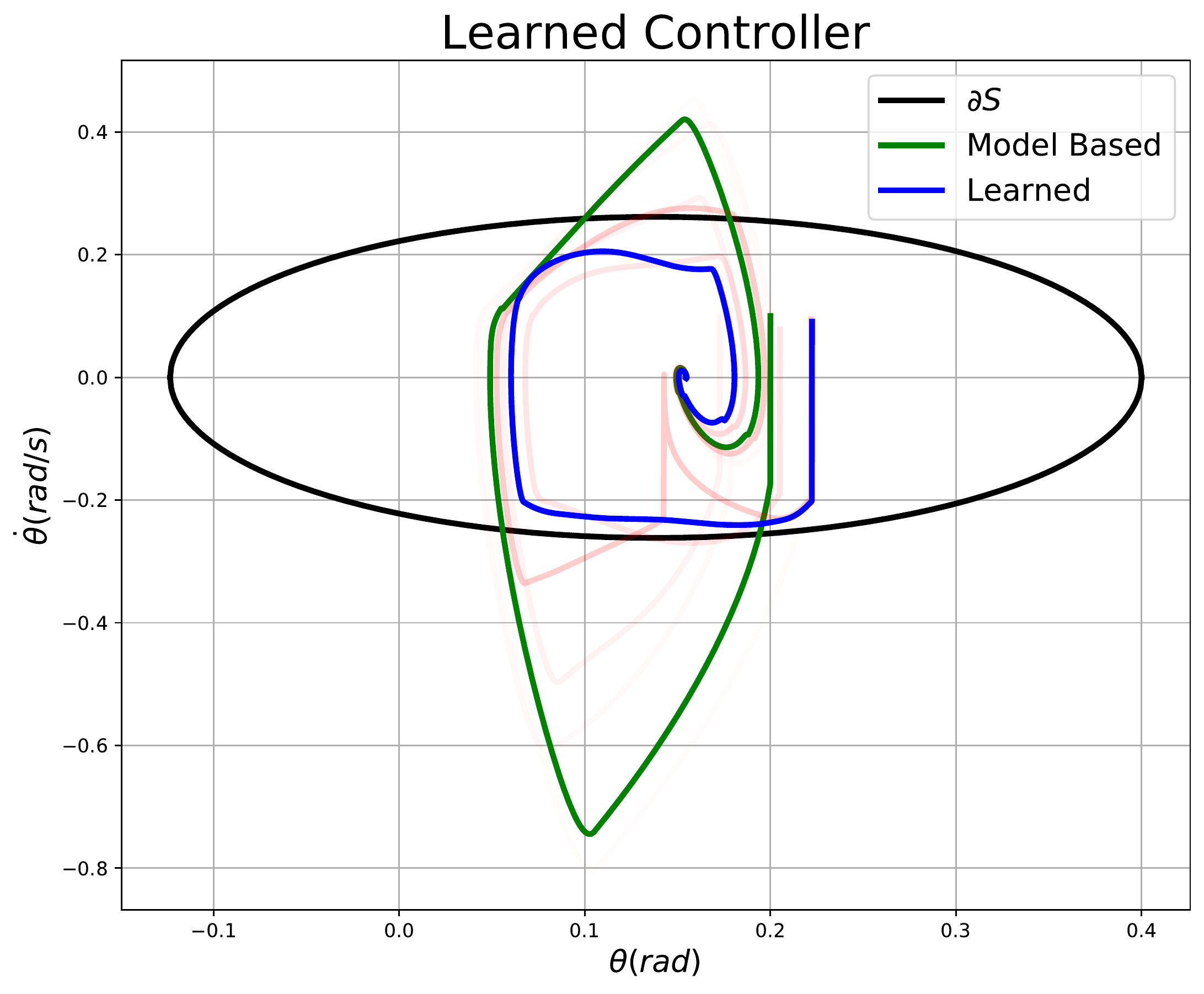}
	\includegraphics[ width=0.32\columnwidth]{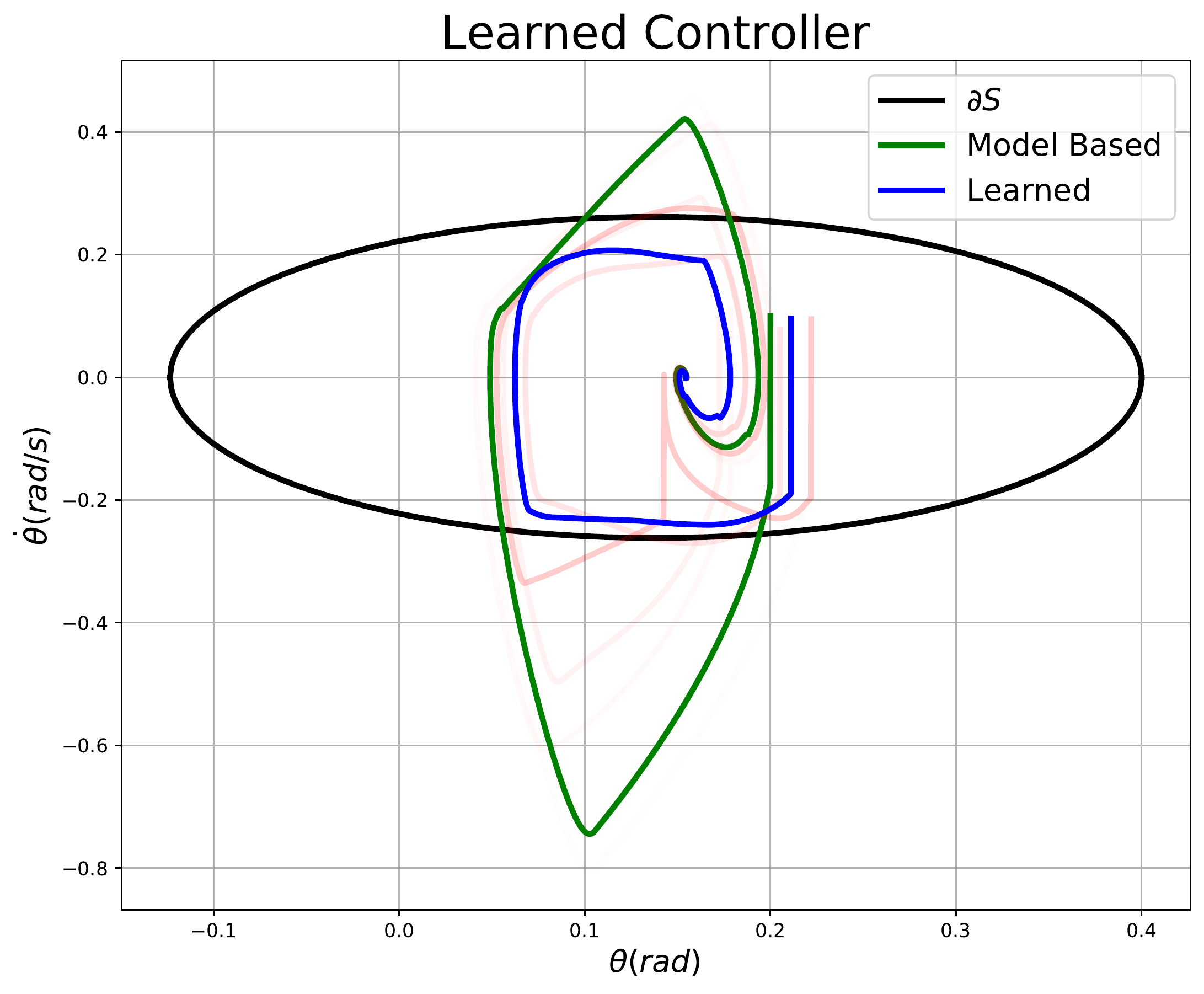}	
	\includegraphics[ width=0.32\columnwidth]{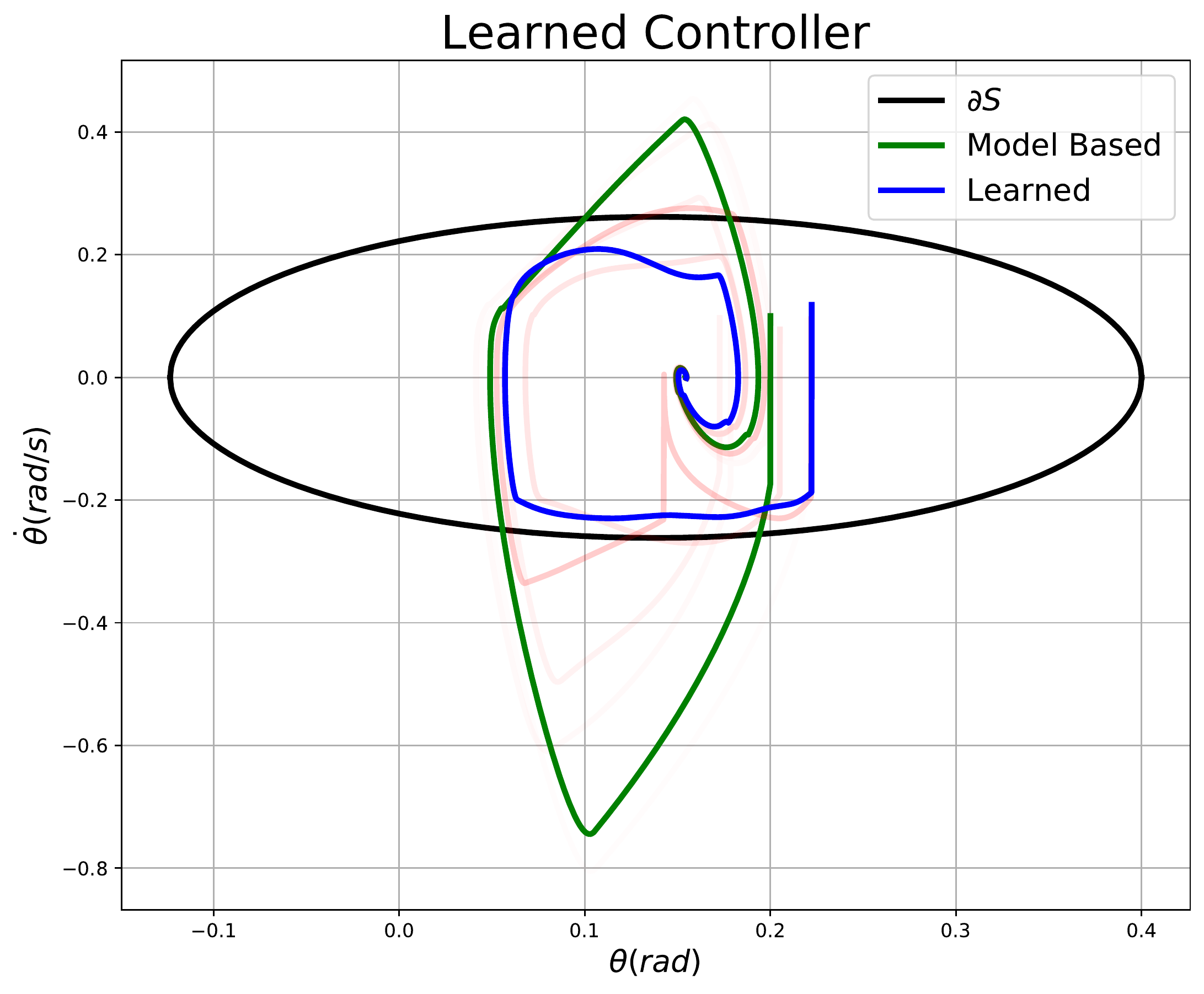}			
	\vspace{-5pt}
	\caption{Segway-ProBF with GP: one random training run, tested on 10 random initial points}\label{fig:NN_test_samples}
	\vspace{3pt}
	\vspace{-0.4cm}
	\label{fig:sgp_seed123}
\end{figure}

\begin{figure}[ht!]
	\centering
	\includegraphics[ width=0.32\columnwidth]{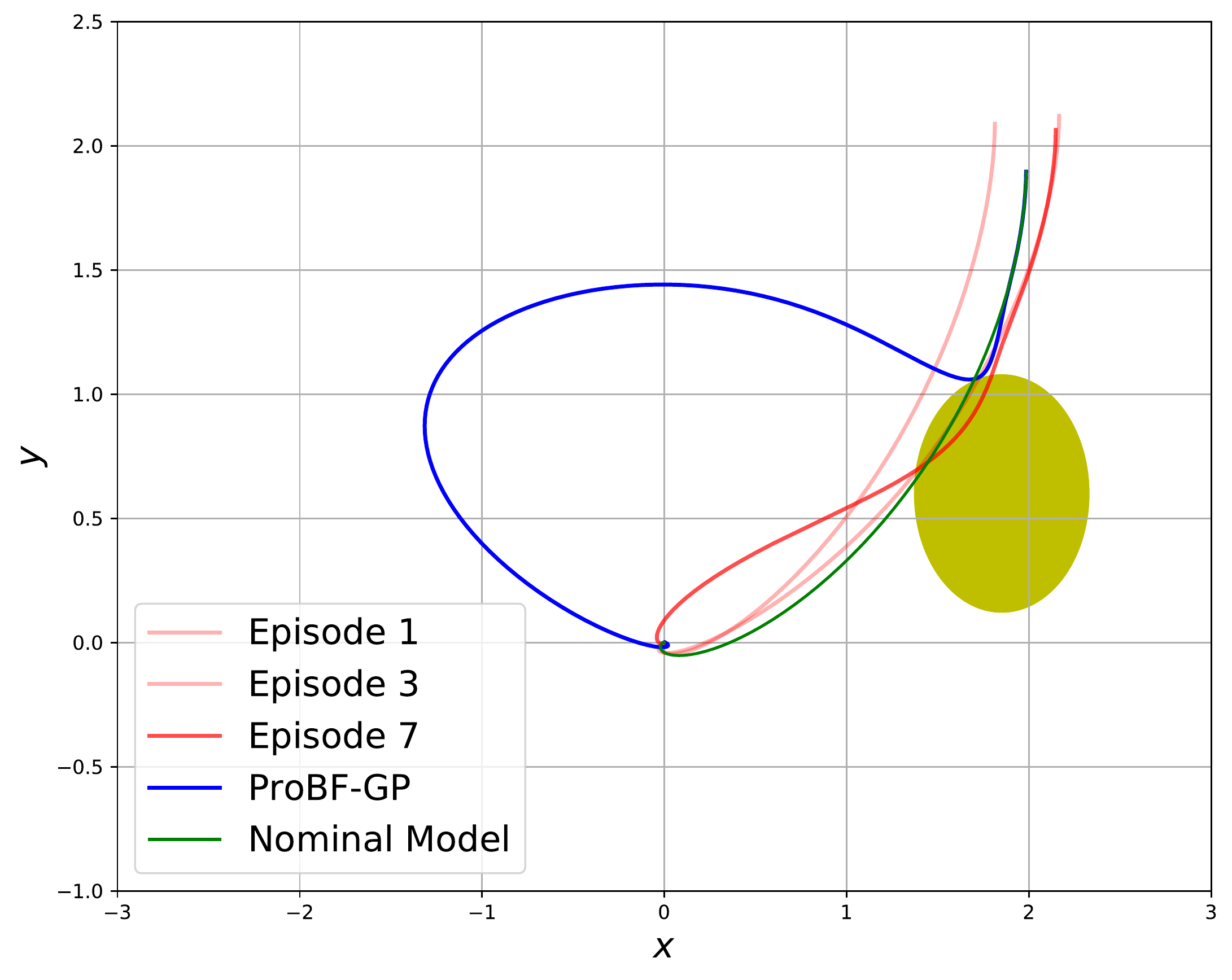}
	\includegraphics[ width=0.32\columnwidth]{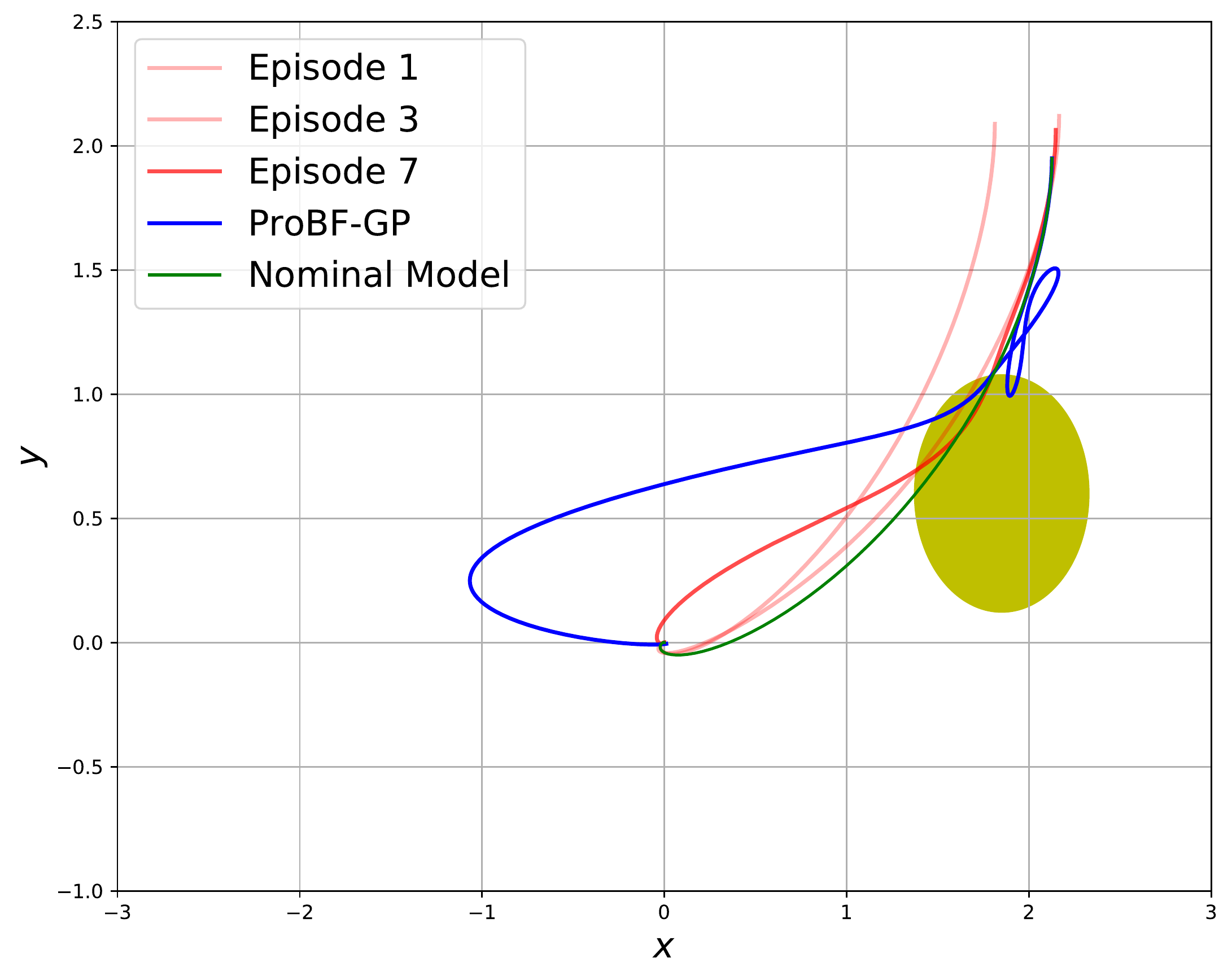}
	\includegraphics[ width=0.32\columnwidth]{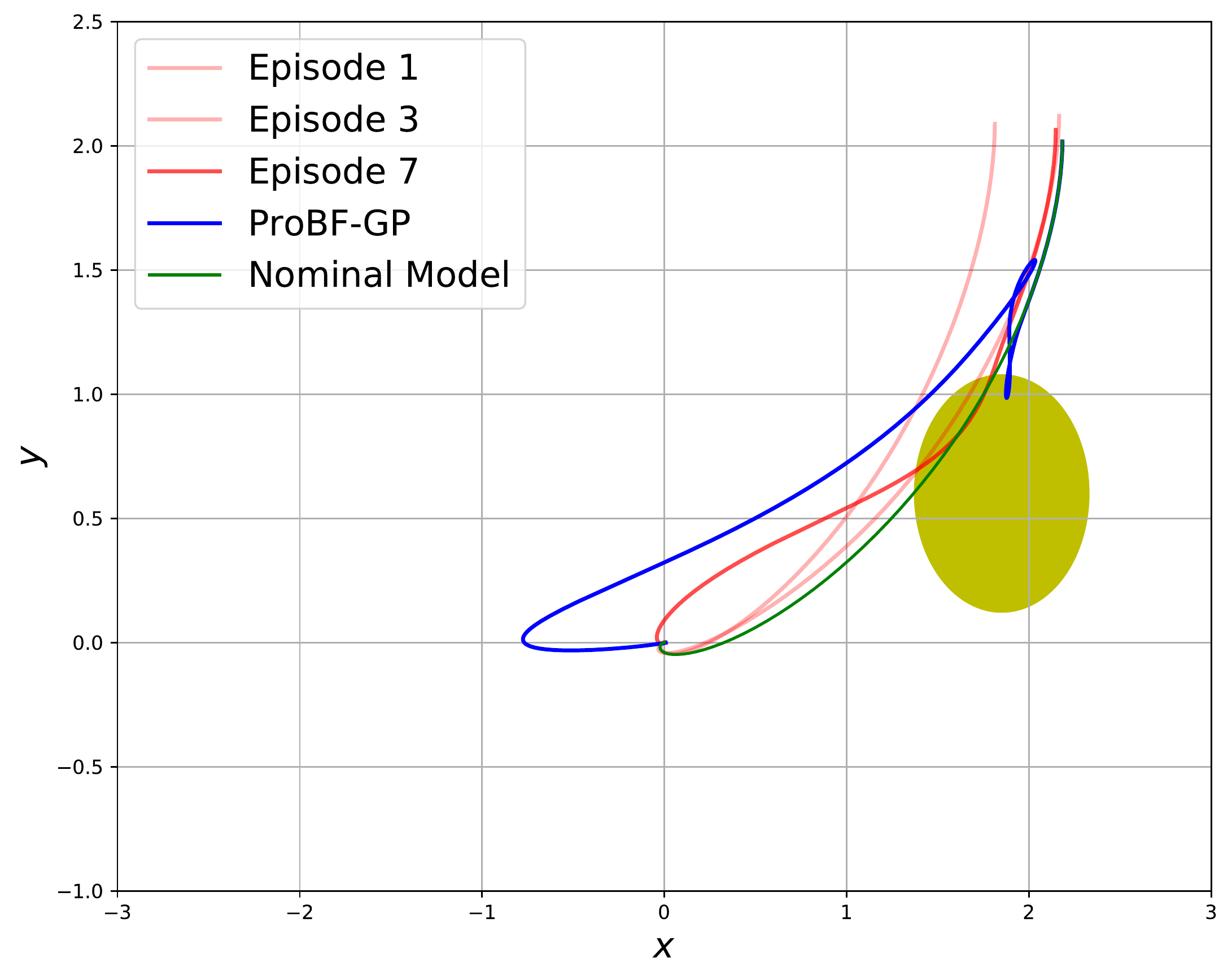}
	\includegraphics[ width=0.32\columnwidth]{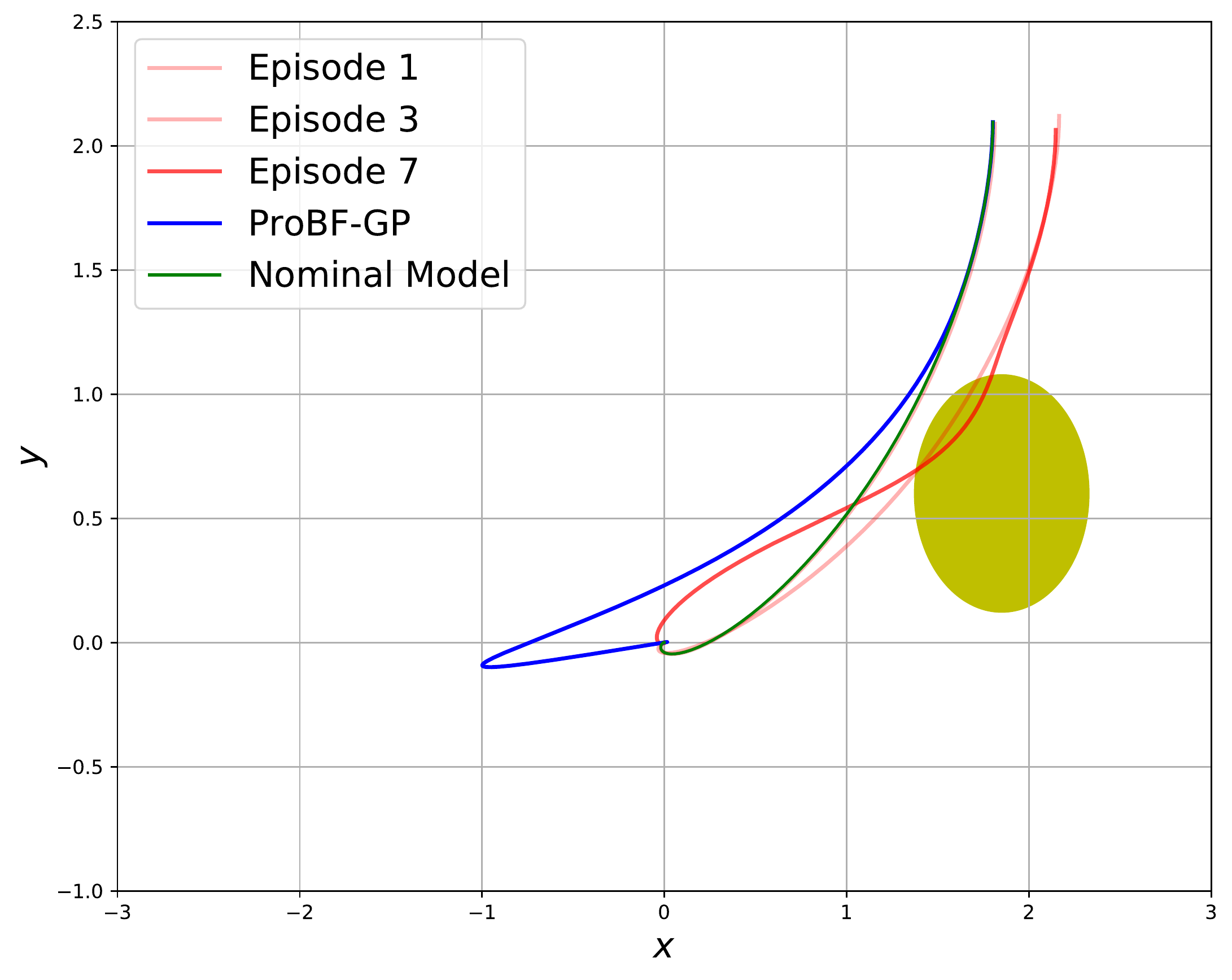}
	\includegraphics[ width=0.32\columnwidth]{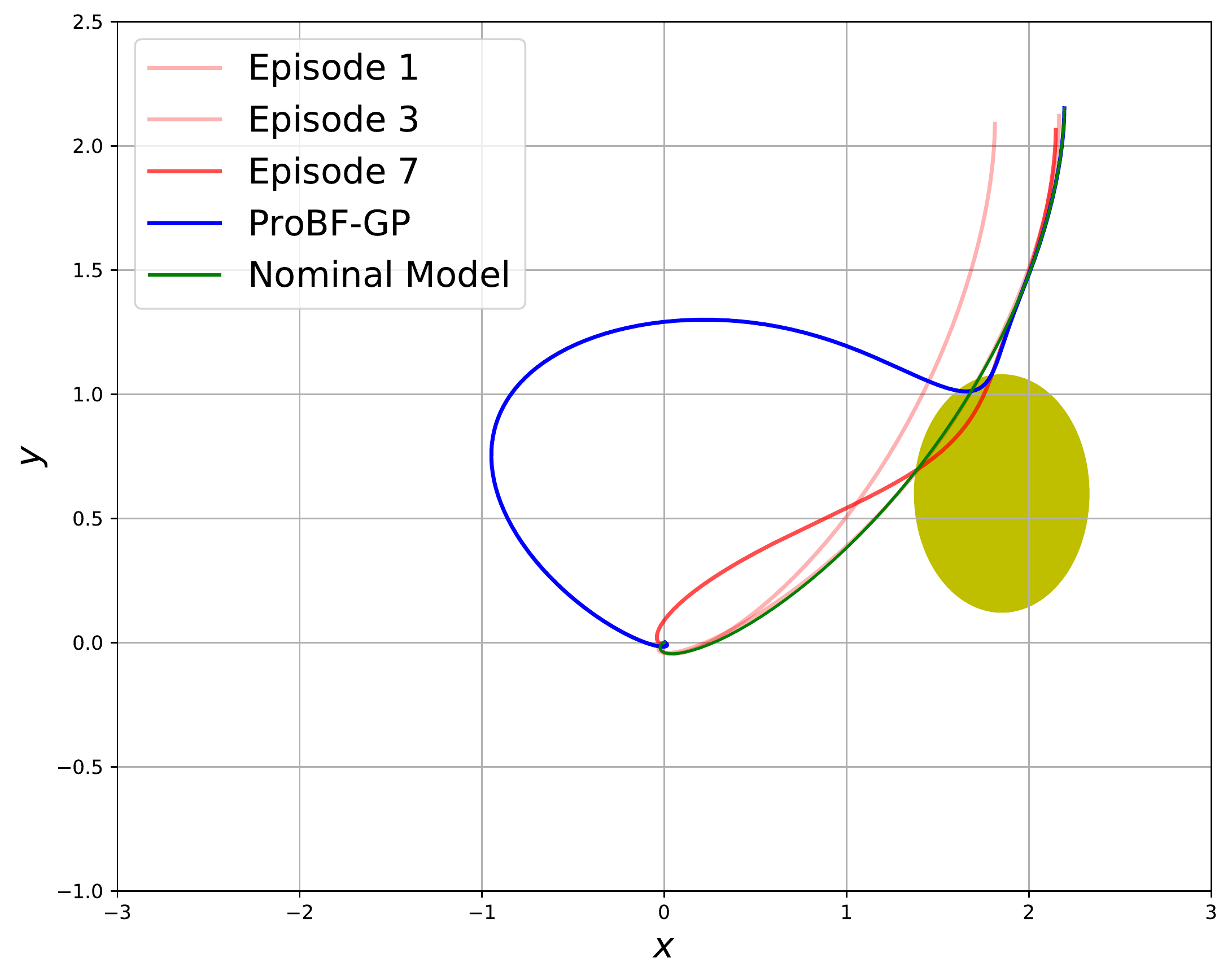}
	\includegraphics[ width=0.32\columnwidth]{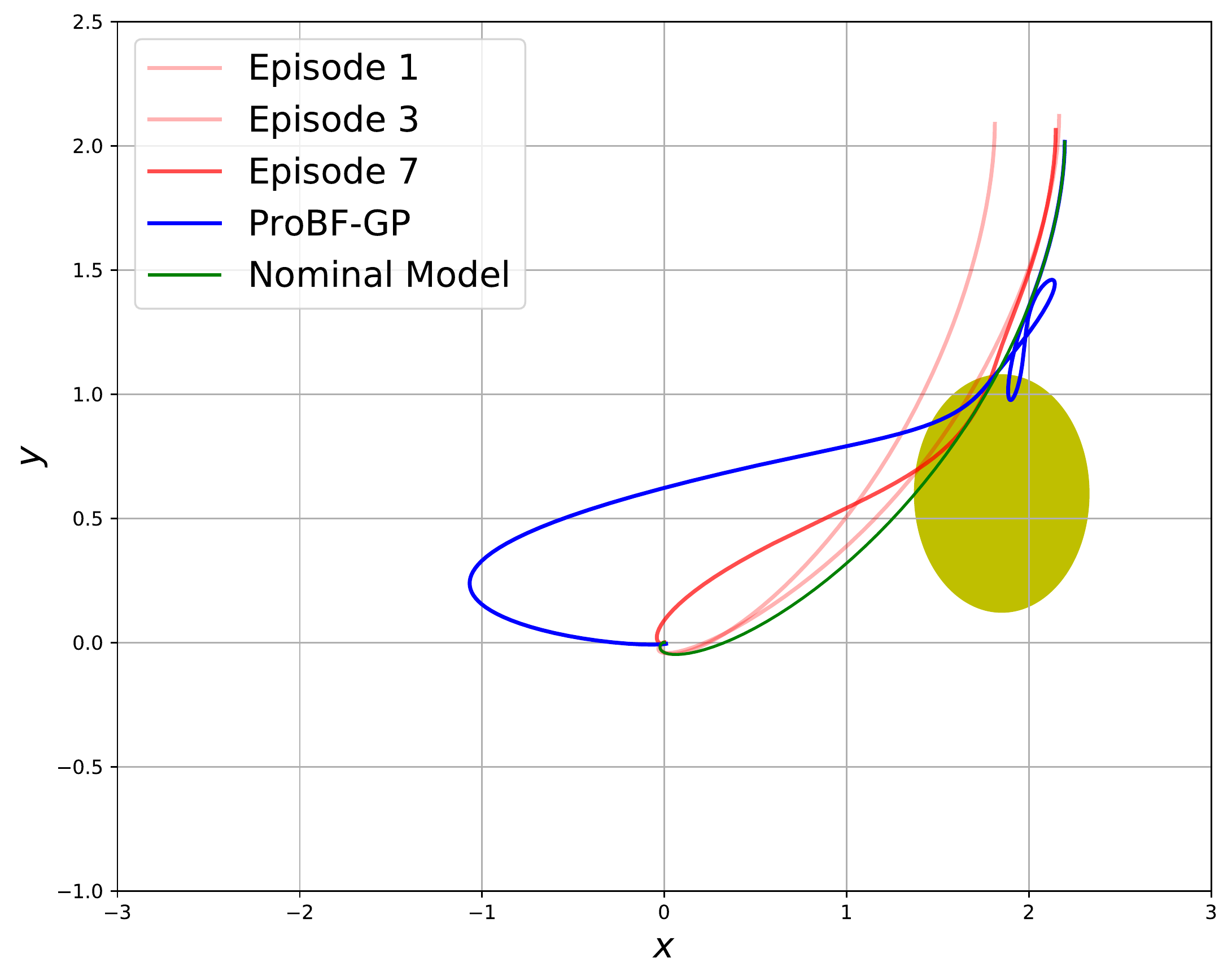}		
	\includegraphics[ width=0.32\columnwidth]{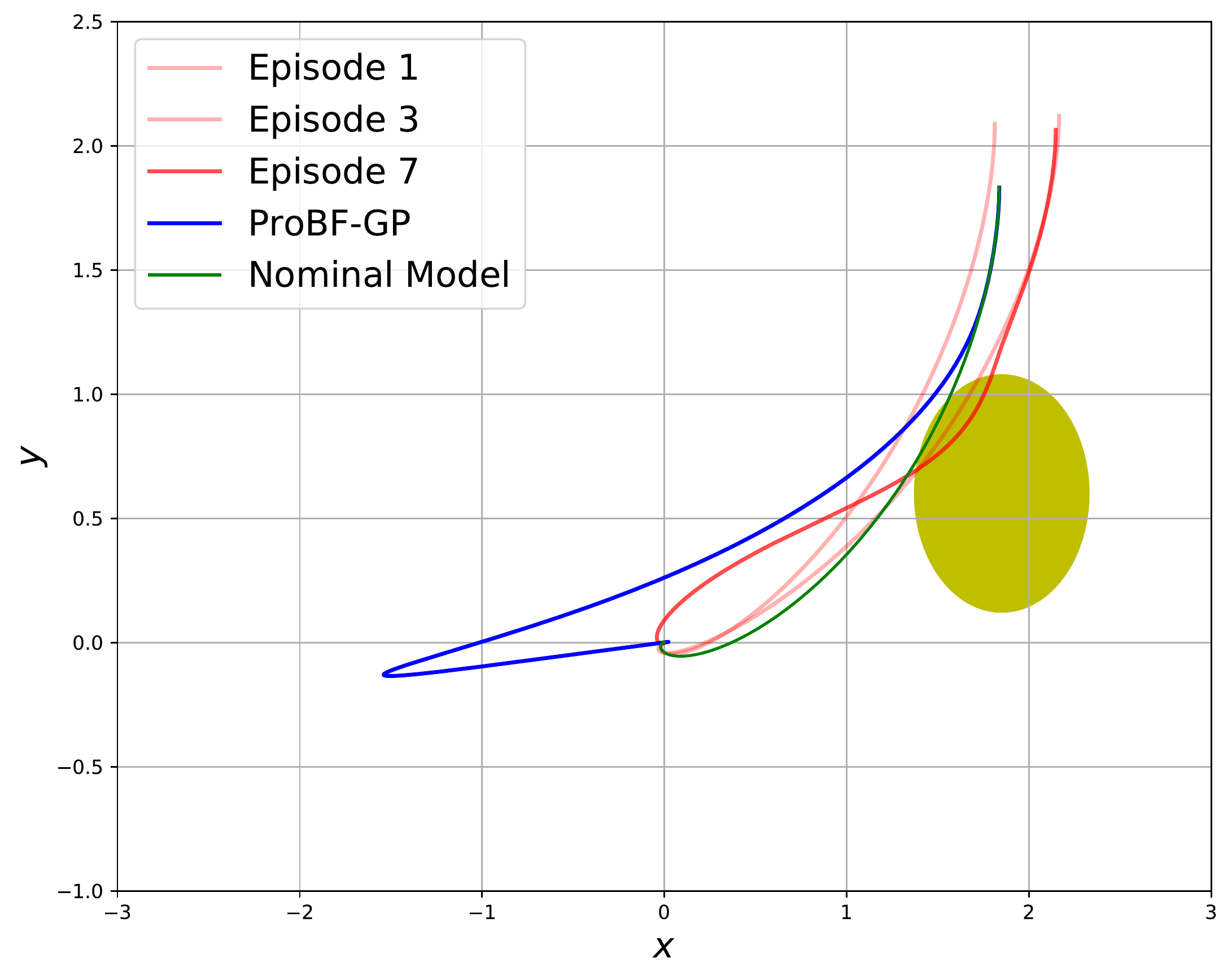}
	\includegraphics[ width=0.32\columnwidth]{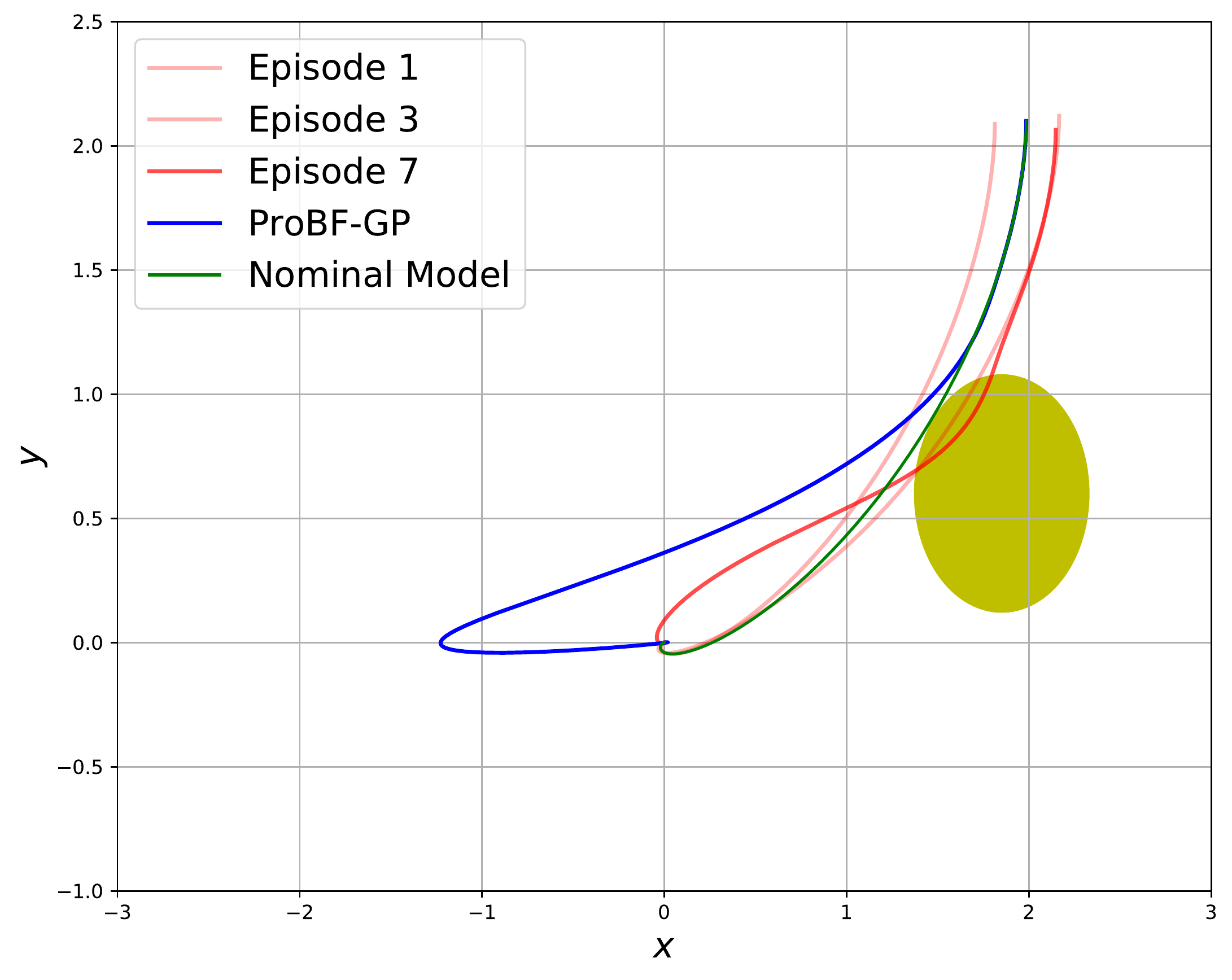}
	\includegraphics[ width=0.32\columnwidth]{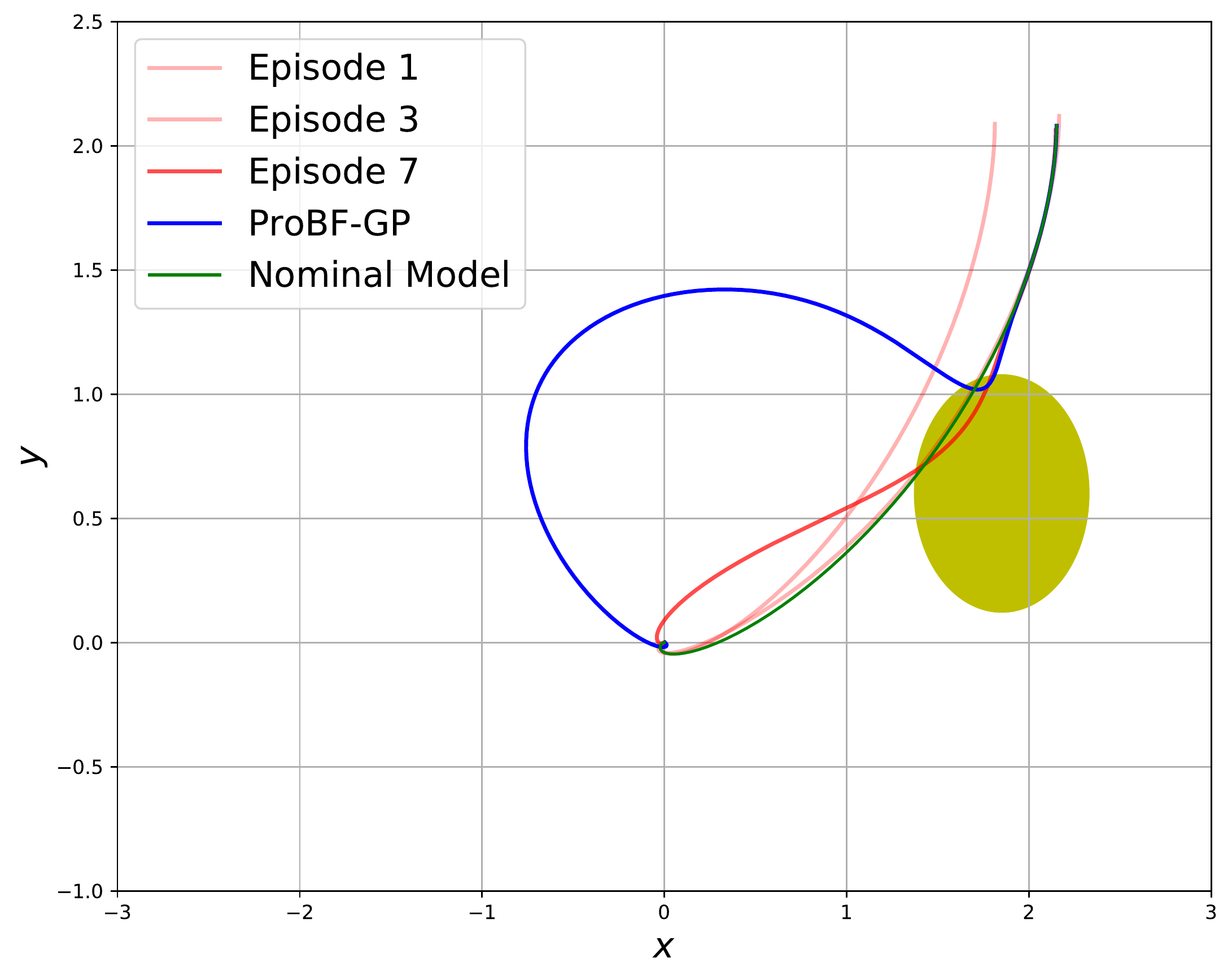}	
	\includegraphics[ width=0.32\columnwidth]{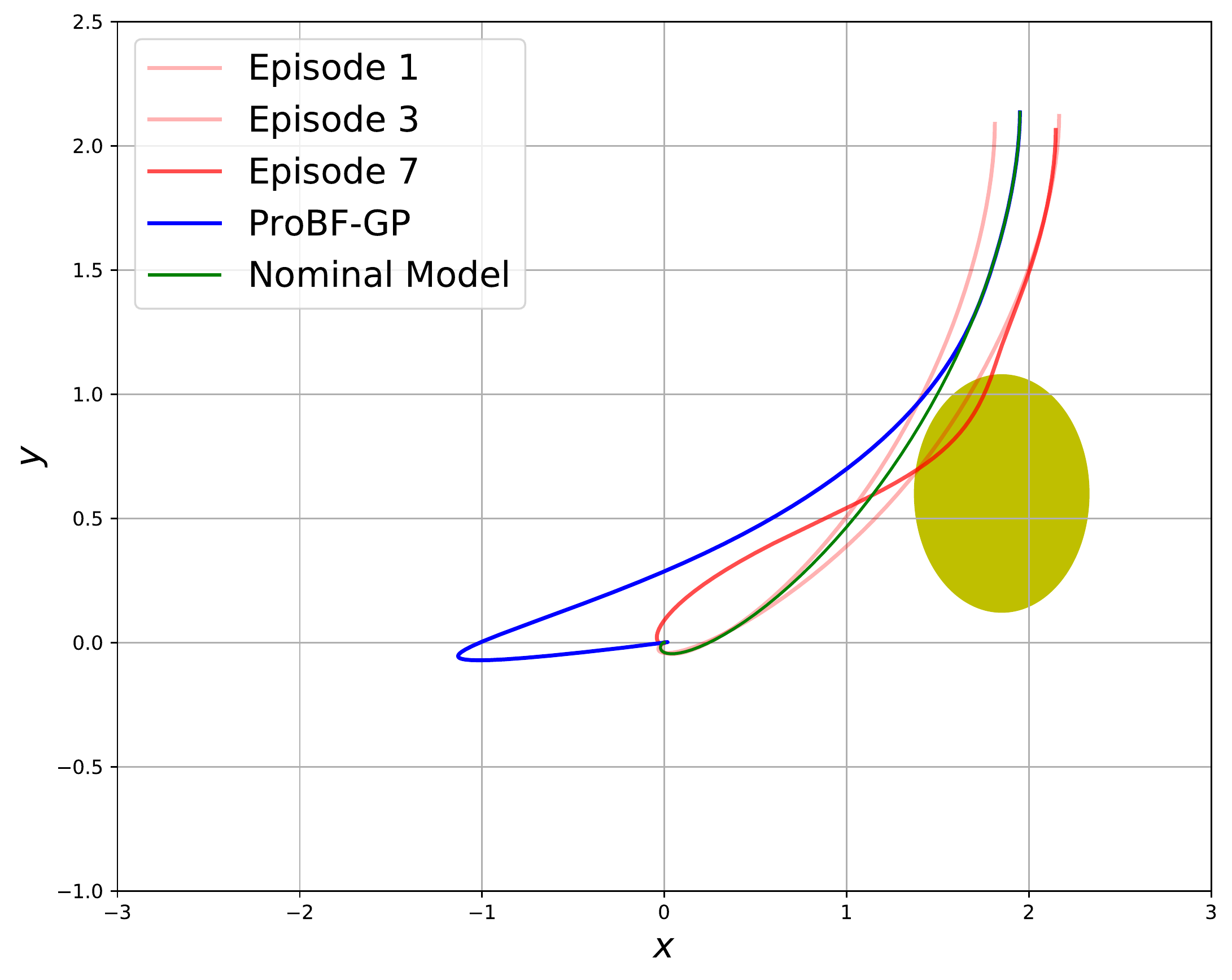}			
	\vspace{-5pt}
	\caption{Quadrotor-ProBF with GP: one random training run, tested on 10 random initial points}\label{fig:GP_quad_samples}
	\vspace{3pt}
	\vspace{-0.4cm}
	\label{fig:qgp_seed123}
\end{figure}

\begin{figure}[ht!]
	\centering
	\includegraphics[ width=0.32\columnwidth]{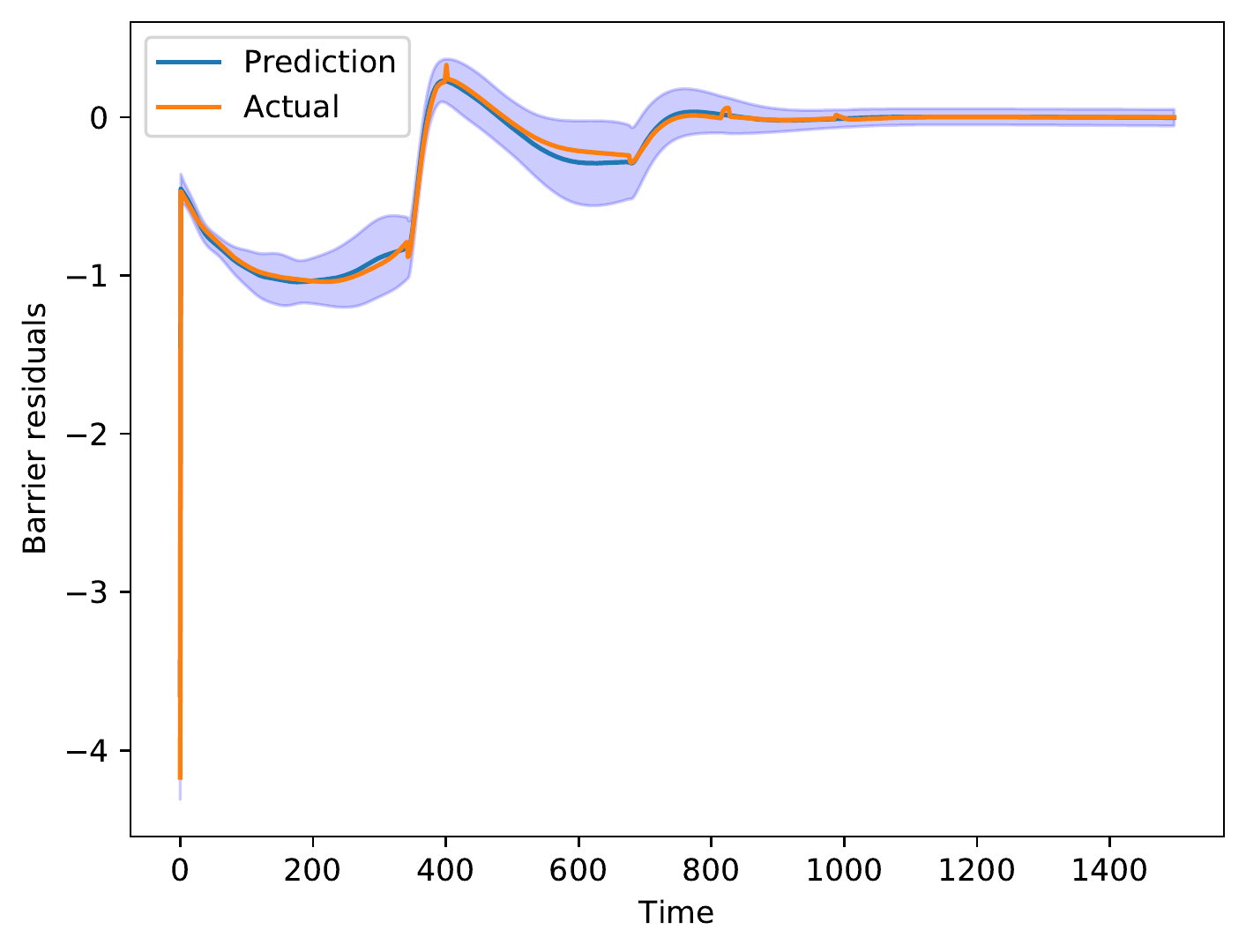}
	\includegraphics[ width=0.32\columnwidth]{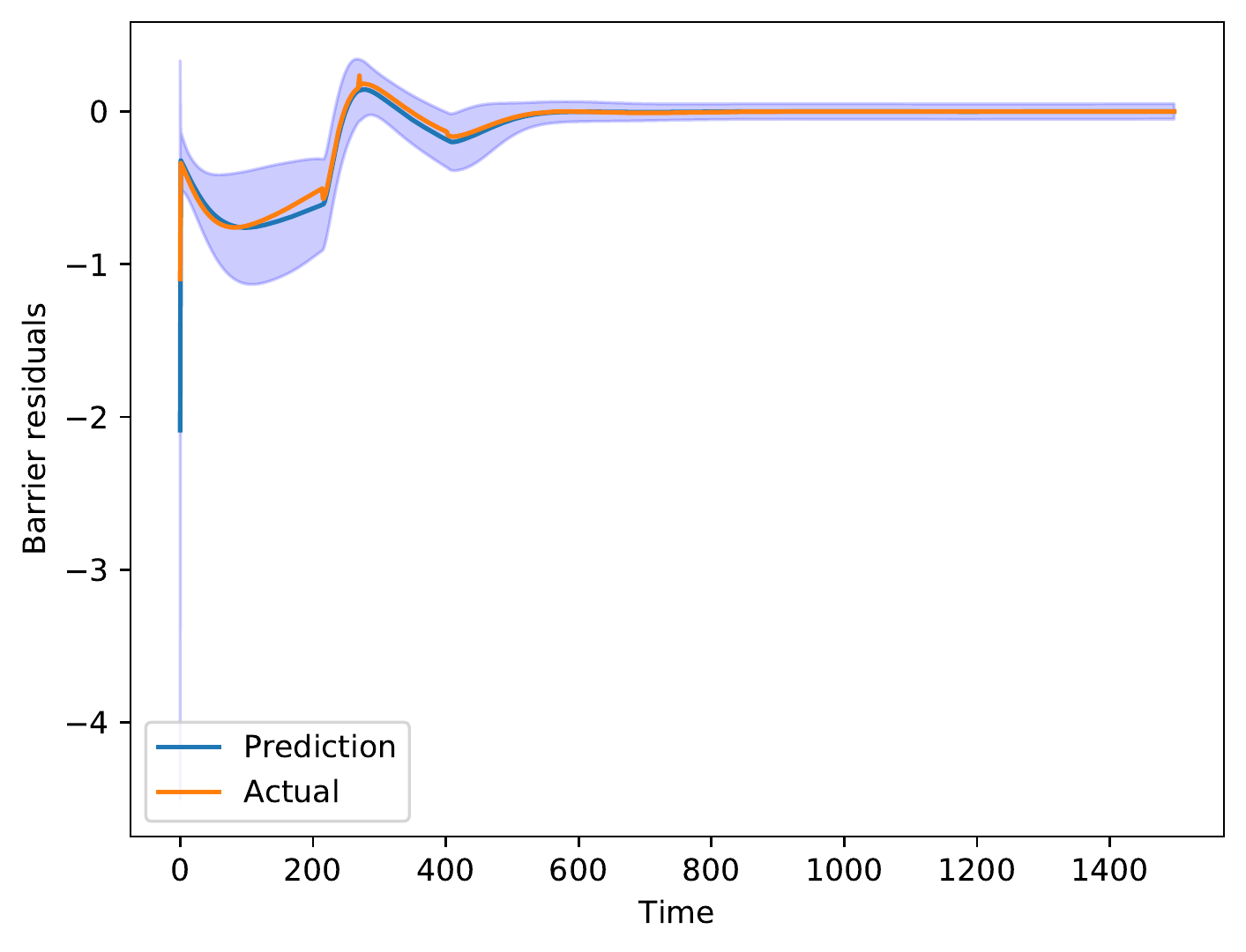}
	\includegraphics[
	width=0.32\columnwidth]{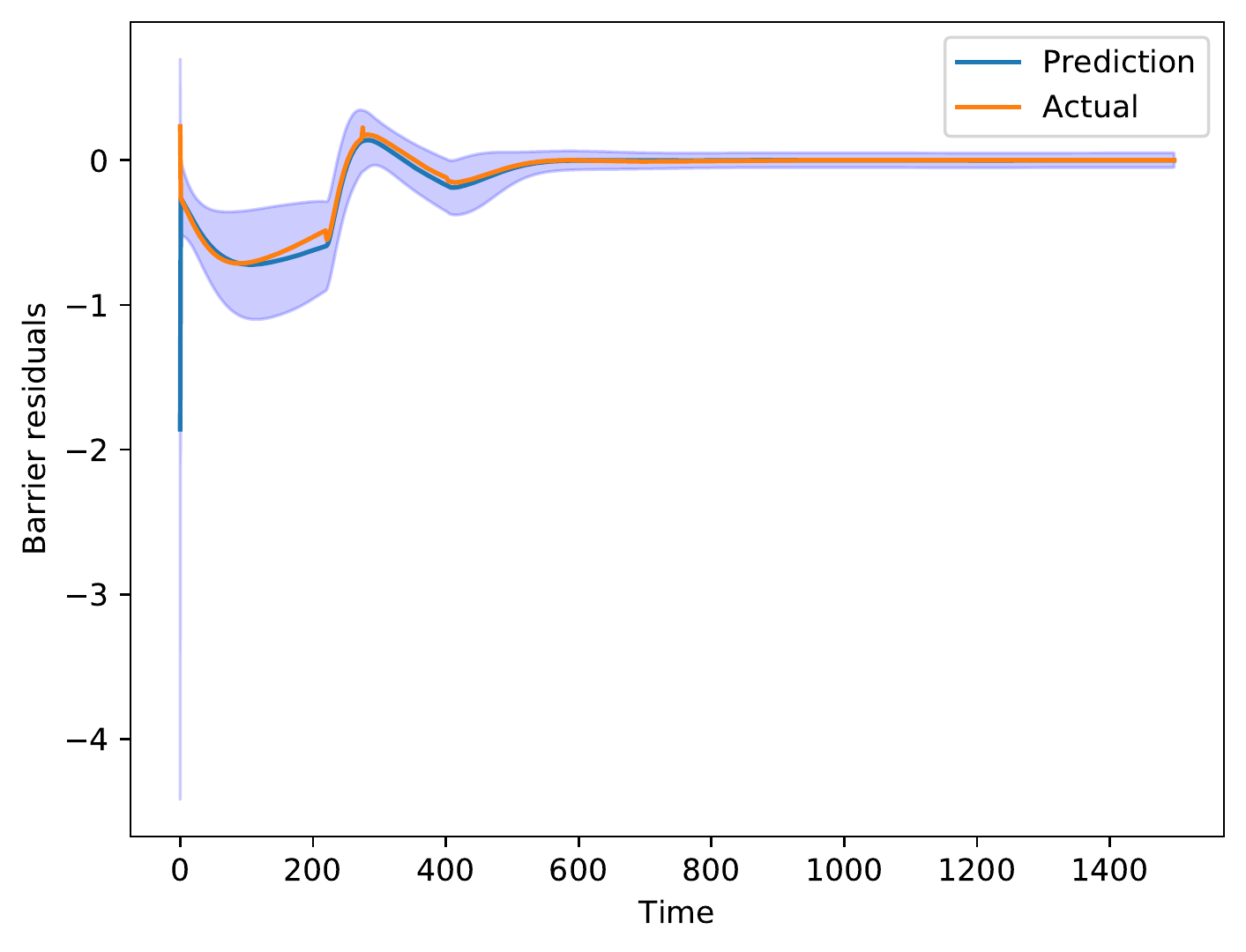}
	\includegraphics[ width=0.32\columnwidth]{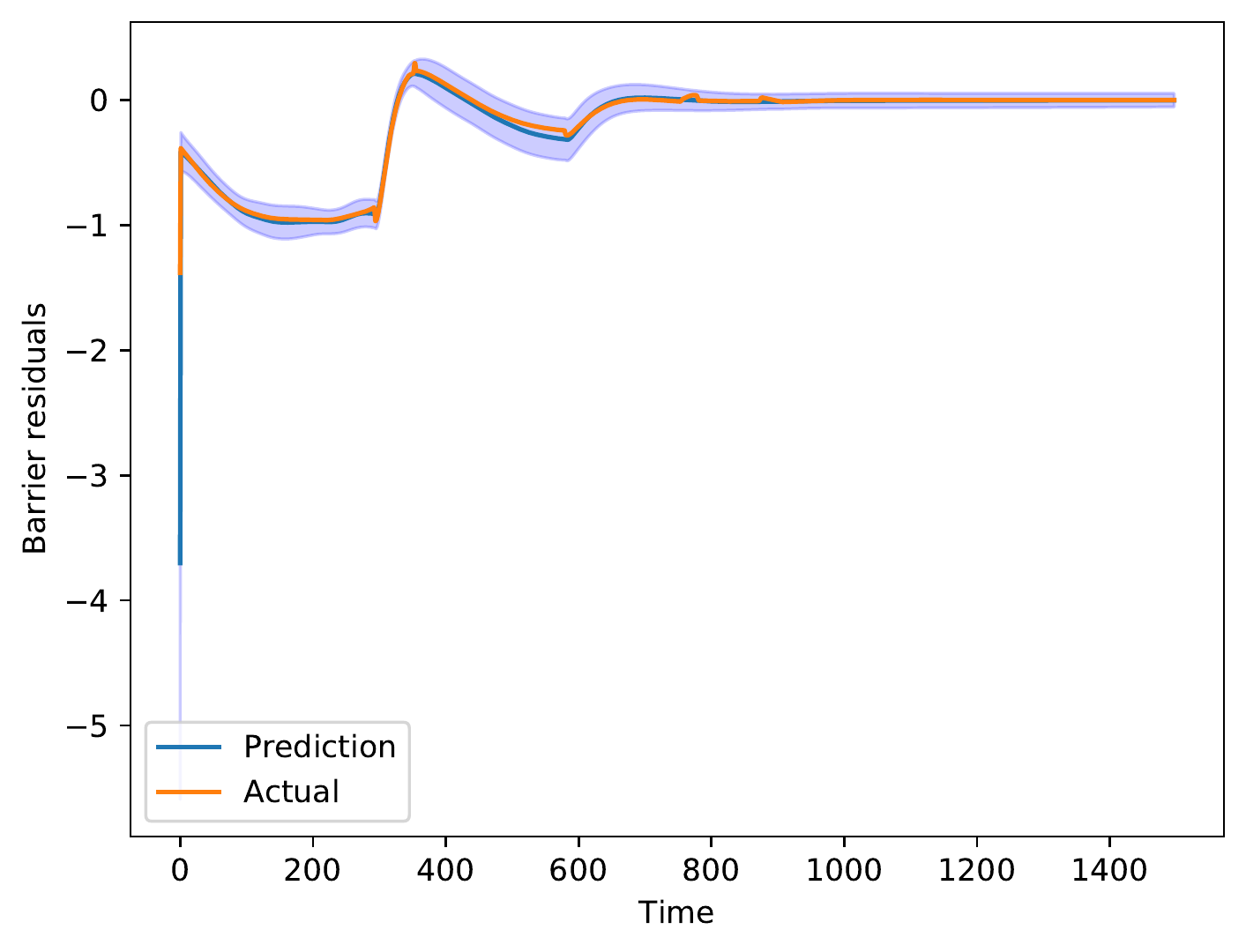}
	\includegraphics[ width=0.32\columnwidth]{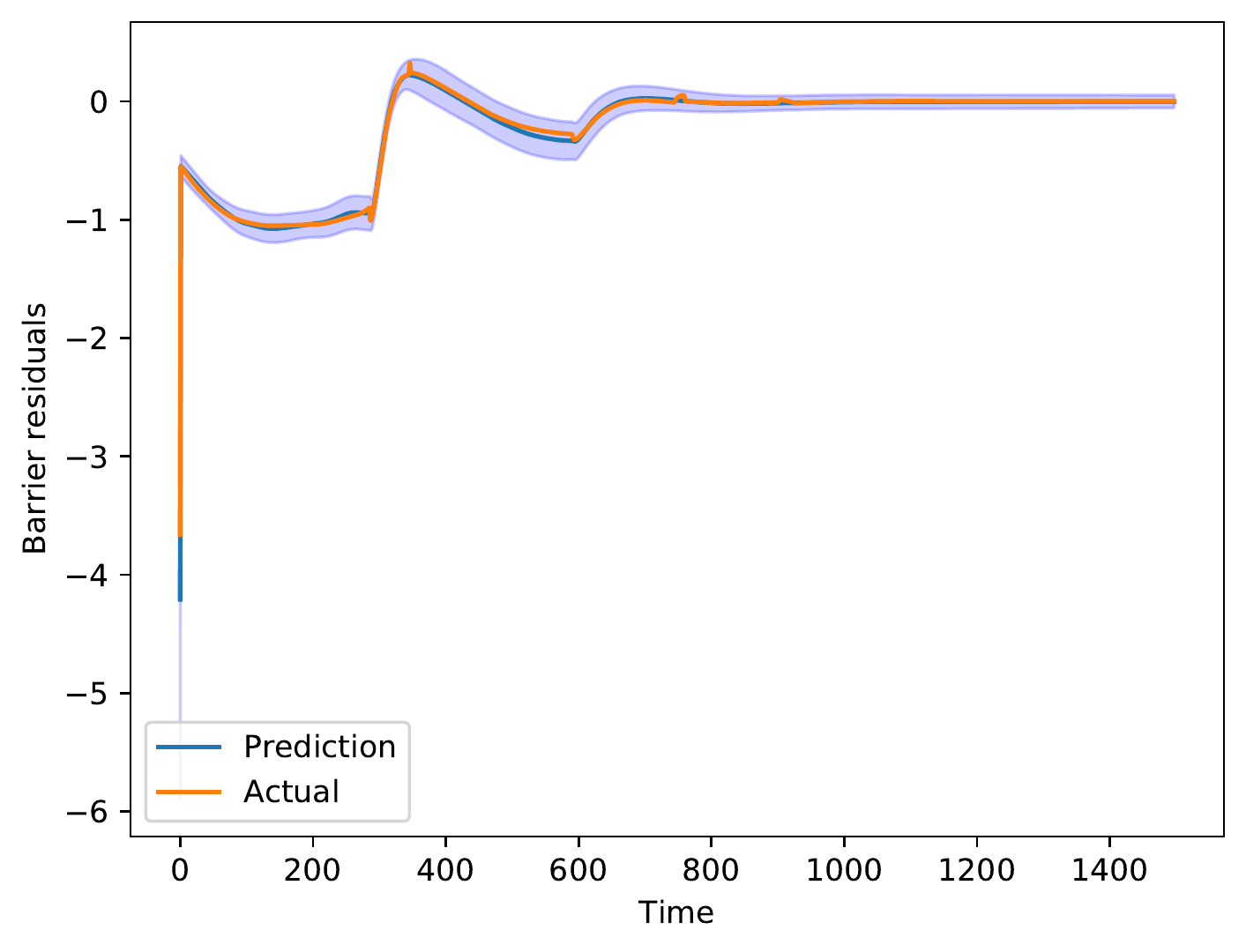}
	\includegraphics[ width=0.32\columnwidth]{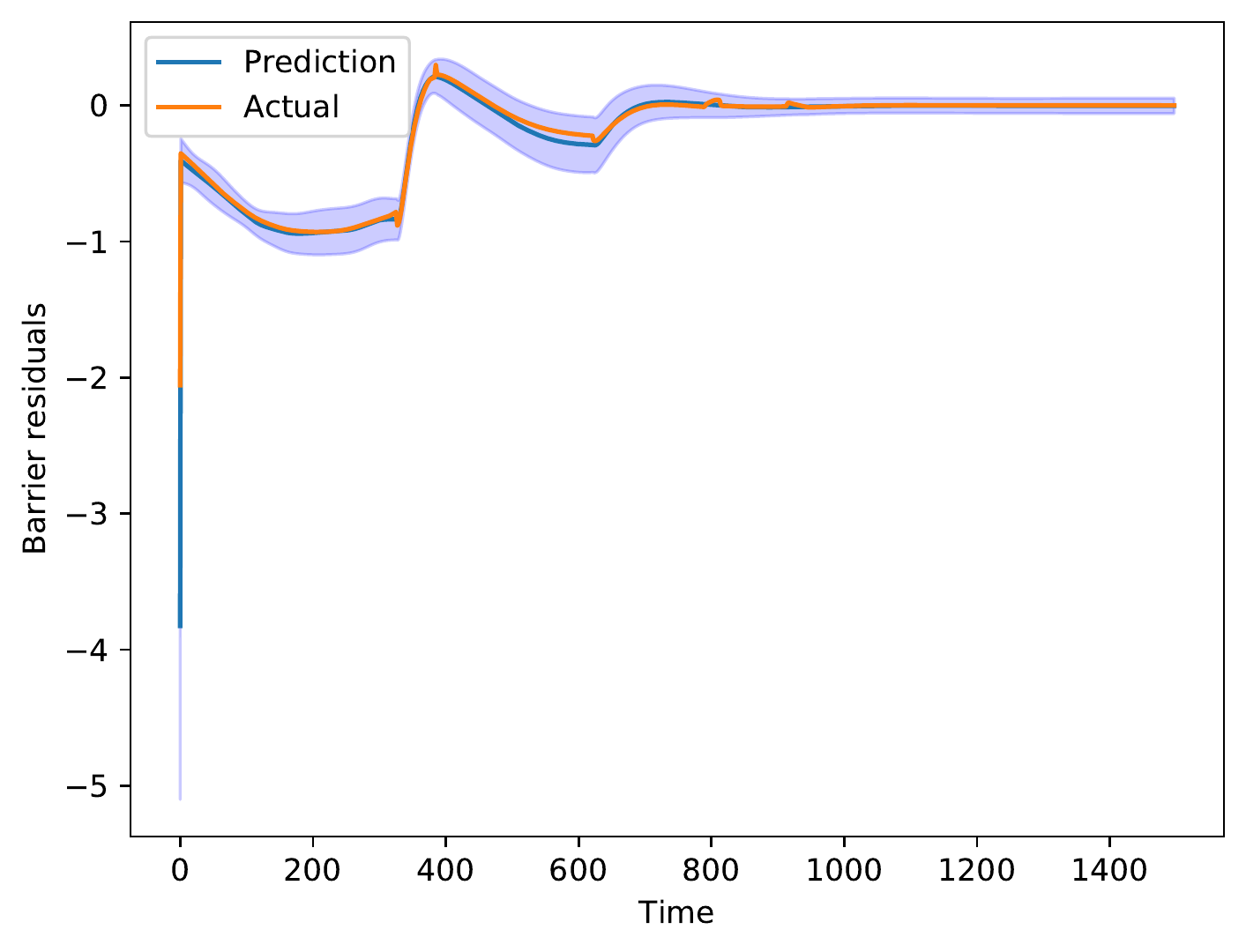}		
	\includegraphics[ width=0.32\columnwidth]{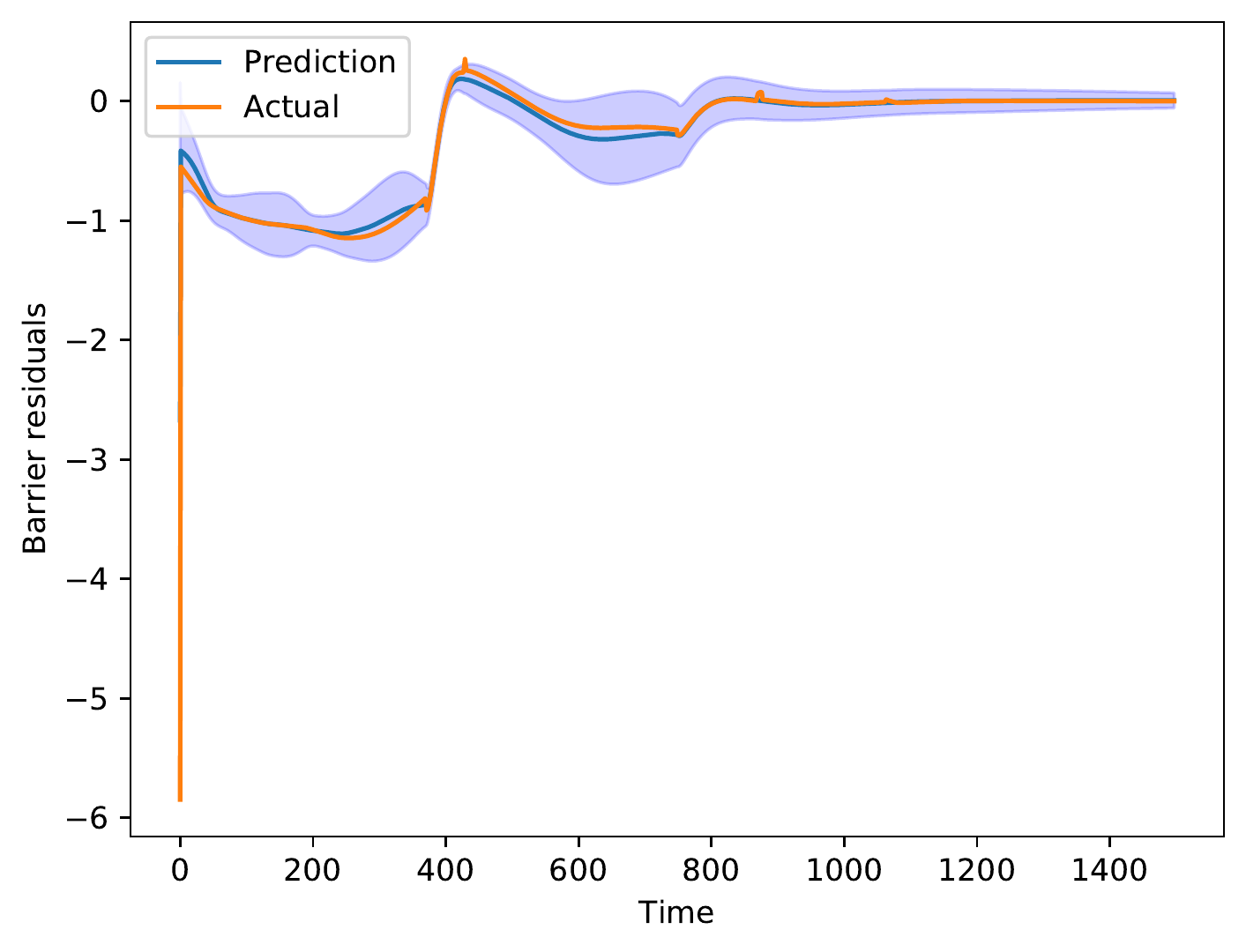}
	\includegraphics[ width=0.32\columnwidth]{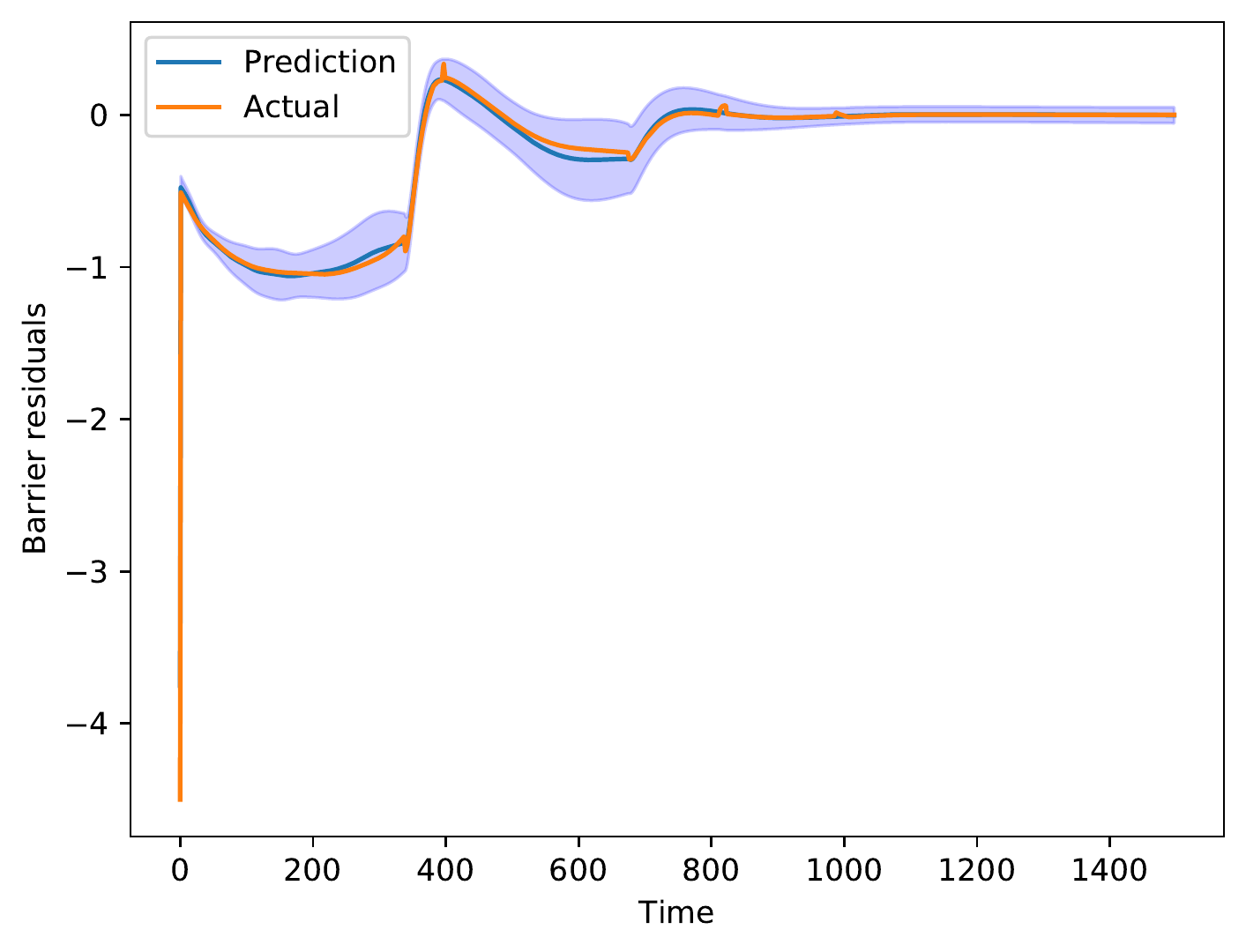}
	\includegraphics[ width=0.32\columnwidth]{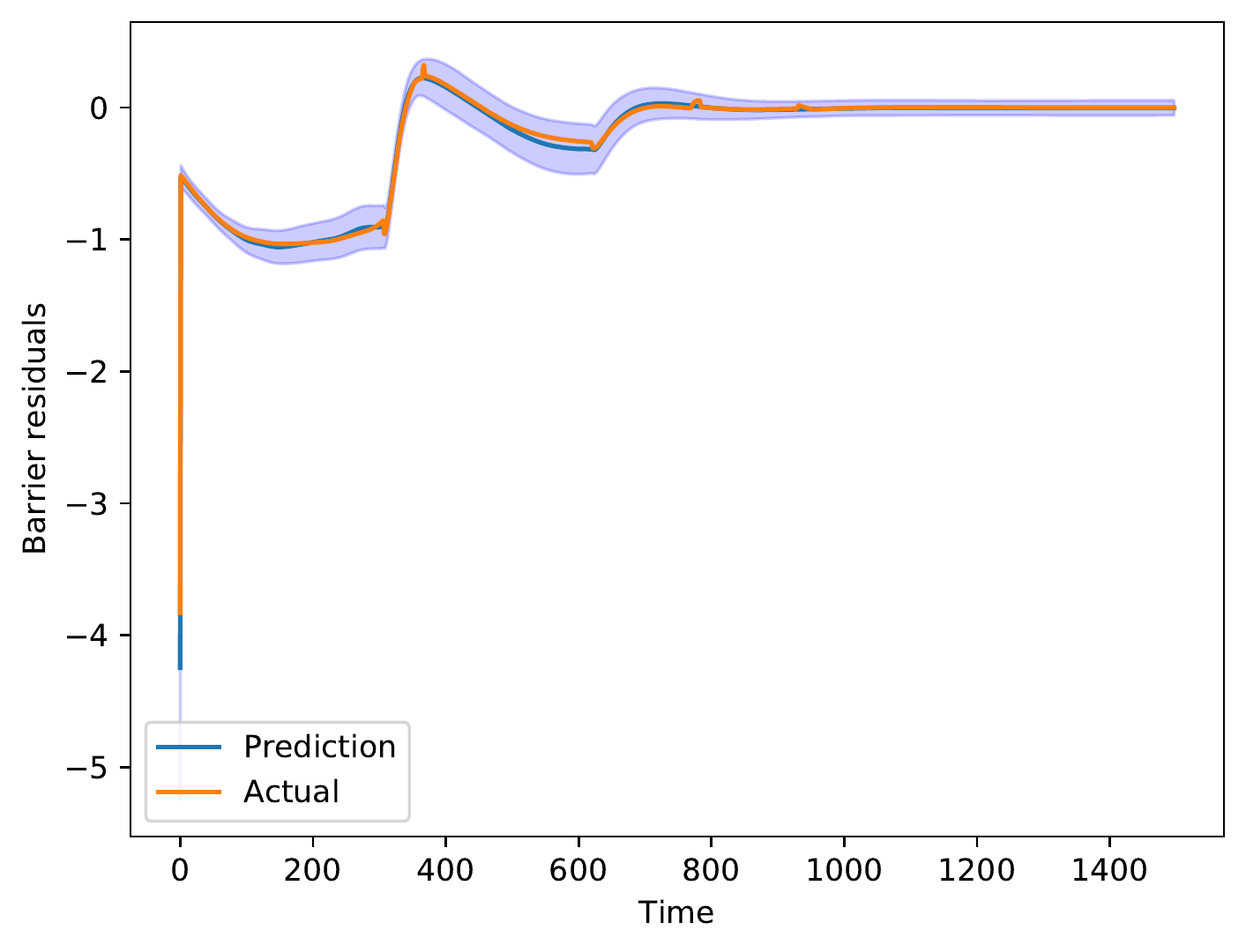}	
	\includegraphics[ width=0.32\columnwidth]{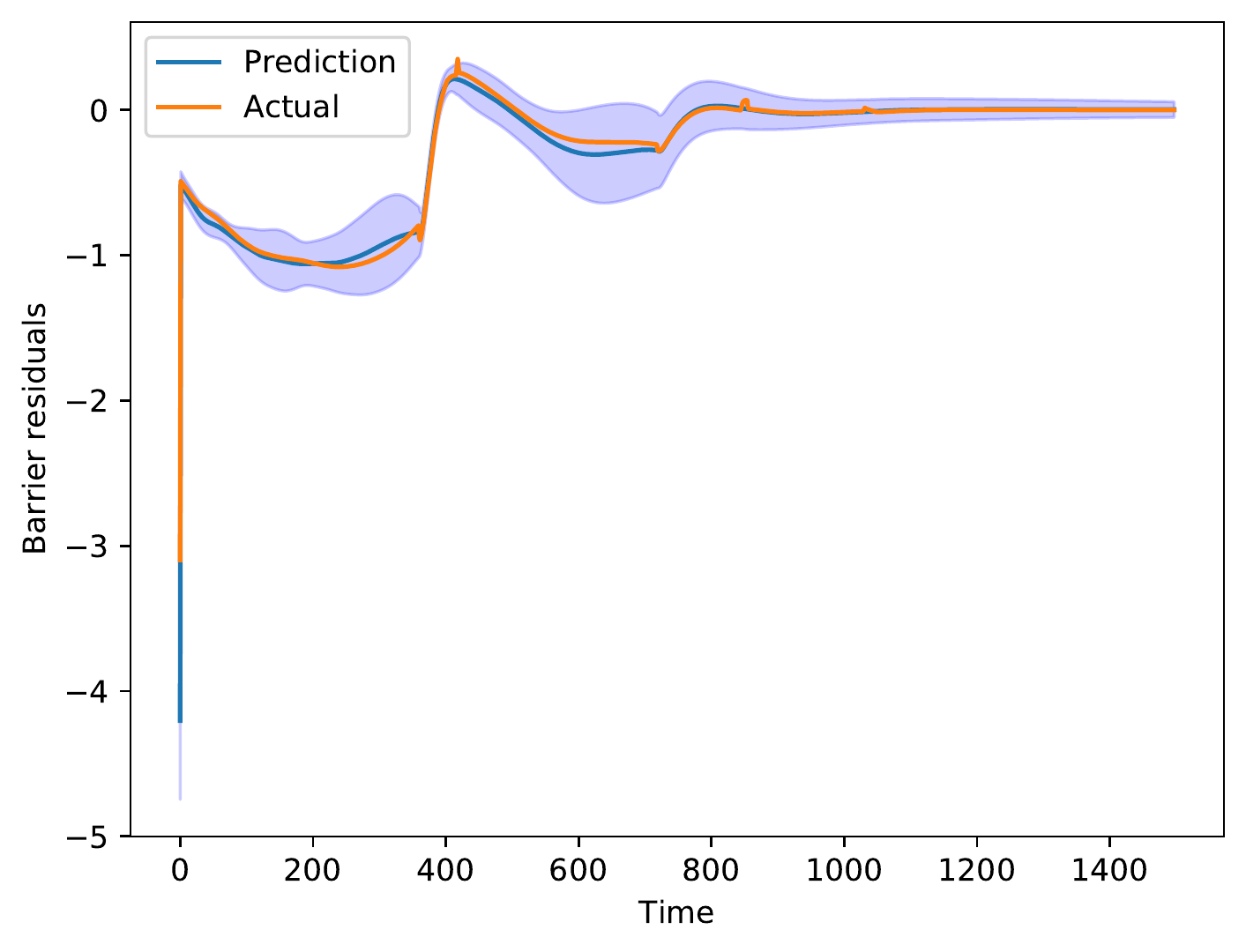}			
	\vspace{-5pt}
	\caption{ProBF with GP: residual predictions along the test trajectories}\label{fig:NN_test_samples}
	\vspace{3pt}
	\vspace{-0.4cm}
	\label{fig:gp_residual_seed123}
\end{figure}

In Figure~\ref{fig:snn_seed123} and~\ref{fig:sgp_seed123}, we plot the trajectories of the testing runs on the same set of 10 random initial points (shown as ``Learned'' in the legend) for one of the training runs for the segway experiment. LCBF-QP and ~\probf are both first trained for 5 episodes using the same initial points for training trajectories (shown as the transparent trajectories in the plots). Similar plots for Quadrotor are in Figures ~\ref{fig:qnn_seed123} and~\ref{fig:qgp_seed123}.
In Figure~\ref{fig:gp_residual_seed123}, we plot the predictions of the GP on the residuals of CBF time derivatives for the Segway experiment.
Code and plots of all training runs can be found at: \url{https://github.com/athindran/ProBF}.


\end{document}